\documentclass[%
 pof,
 aip,
amsmath,amssymb,
preprint,%
tightenlines,%
floatfix
]{revtex4-1}

\usepackage[pagebackref=false, colorlinks, linkcolor=blue, citecolor=blue, urlcolor=blue, pdfauthor=author]{hyperref}
\usepackage{graphicx,subcaption,tabularx}
\graphicspath{{Figures/}}
\usepackage{booktabs}
\usepackage{multirow}
\usepackage{bm}
\usepackage{dcolumn}
\usepackage[utf8]{inputenc}
\usepackage[T1]{fontenc}
\usepackage{mathptmx}
\usepackage{etoolbox}
\makeatletter
\def\@email#1#2{%
	\endgroup
	\patchcmd{\titleblock@produce}
	{\frontmatter@RRAPformat}
	{\frontmatter@RRAPformat{\produce@RRAP{*#1\href{mailto:#2}{#2}}}\frontmatter@RRAPformat}
	{}{}
}%
\makeatother

\begin{document}

\title[Transfer learning strategies for accelerating reinforcement-learning-based flow control]{Transfer learning strategies for accelerating reinforcement-learning-based flow control}

\author{Saeed Salehi}

\affiliation{Division of Applied Thermodynamics and Fluid Mechanics, Department of Management and Engineering, Linköping University, SE-581 83 Linköping, Sweden}%
\email{saeed.salehi@liu.se}

\date{\today}

\begin{abstract}
	
This work investigates transfer learning strategies to accelerate deep reinforcement learning (DRL) for multifidelity control of chaotic fluid flows. 
Progressive neural networks (PNNs), a modular architecture designed to preserve and reuse knowledge across tasks, are employed for the first time in the context of DRL-based flow control. In addition, a comprehensive benchmarking of conventional fine-tuning strategies is conducted, evaluating their performance, convergence behavior, and ability to retain transferred knowledge.
The Kuramoto--Sivashinsky (KS) system is employed as a benchmark to examine how knowledge encoded in control policies, trained in low-fidelity environments, can be effectively transferred to high-fidelity settings.
Systematic evaluations show that while fine-tuning can accelerate convergence, it is highly sensitive to pretraining duration and prone to catastrophic forgetting. In contrast, PNNs enable stable and efficient transfer by preserving prior knowledge and providing consistent performance gains, and are notably robust to overfitting during the pretraining phase. 
Layer-wise sensitivity analysis further reveals how PNNs dynamically reuse intermediate representations from the source policy while progressively adapting deeper layers to the target task.
Moreover, PNNs remain effective even when the source and target environments differ substantially, such as in cases with mismatched physical regimes or control objectives, where fine-tuning strategies often result in suboptimal adaptation or complete failure of knowledge transfer.
The results highlight the potential of novel transfer learning frameworks for robust, scalable, and computationally efficient flow control that can potentially be applied to more complex flow configurations.
	
\end{abstract}

\maketitle

\section{Introduction}
\label{sec:intro}

Reinforcement learning (RL) provides a general framework for solving sequential decision-making problems, where an agent learns optimal behavior through interaction with its environment. The integration of deep learning into RL has led to the development of Deep Reinforcement Learning (DRL), where neural networks are used to approximate policies or value functions in high-dimensional spaces. 

DRL has made it possible to address complex control problems that were previously intractable, due to the scalability and representational power of deep neural networks. Prominent achievements such as achieving super performance level at Atari video games~\citep{Mnih2015} and defeating world champions in the game of Go~\citep{Silver2017} have demonstrated the remarkable potential of DRL and contributed to its rapid rise in popularity and research interest.

In recent years, deep reinforcement learning has shown promising results in the domain of fluid dynamics, where complex, nonlinear interactions and high-dimensional state spaces pose significant challenges for traditional control and optimization methods. Neural networks trained via RL have been successfully applied to a range of fluid-related tasks, including the navigation of autonomous gliders~\citep{Reddy2016}, the coordination of swimming strategies among interacting fish~\citep{Novati2017, Verma2018}, and the control of rigid bodies in unsteady flows~\citep{Ma2018}. Beyond control, DRL has also been employed in shape optimization problems, where it learns to explore geometries that enhance flow performance~\citep{YeowLee, Viquerat2021}.

Deep reinforcement learning has also proven effective in discovering flow control strategies, including those targeting drag reduction and suppression of vortex shedding. Successful applications span a range of regimes, from laminar~\citep{Rabault2019a} and weakly turbulent~\citep{Ren2021} to fully turbulent flows~\citep{Fan2020}.

Despite these advances, extending DRL to high-Reynolds-number flows presents new challenges due to the increased complexity, chaotic behavior, and strong nonlinearity of turbulent dynamics. As noted by \citet{fluids7020062}, one of the key frontiers in the field is the extension of DRL-based control to fully turbulent conditions. The research community has already taken initial steps in this direction, demonstrating DRL's effectiveness in controlling chaotic dynamical systems that are significantly more intricate than periodic or quasi-periodic systems~\citep{Bucci2019, Peitz2023, Sonoda2023a, fontDeepReinforcementLearning2025}.

\citet{zhouReinforcementlearningbasedControlTurbulent2025a} demonstrated the effectiveness of such controllers for drag reduction in fully developed turbulent channel flows up to $\mathrm{Re}_\tau=\text{1000}$.
The DRL agent used a sophisticated non-linear control strategy that considers both streamwise and normal velocity fluctuations, unlike opposition control, which only considers normal fluctuations.

Despite progressive advancements in controlling complex flow configurations through DRL, \emph{computational efficiency} stands as one of the primary barriers to expanding the applicability of DRL to more realistic flows with higher Reynolds numbers due to the substantial increase in computational cost. DRL algorithms typically require a significant number of interactions between the agent and the environment. Given that over 99\% of computational costs are attributed to CFD computations~\citep{Rabault2019}, DRL-CFD frameworks tend to be computationally expensive. 

\citet{chatzimanolakisLearningTwoDimensions2024} argued that direct training in 3D at high Reynolds numbers was not feasible due to excessive computational demands, while \citet{suarezActiveFlowControl2025} reported requiring over 3 million CPU hours to train a controller for flow past a cylinder at $\mathrm{Re}=\text{3900}$. Notably, recent studies on turbulent drag reduction~\citep{Sonoda2023a, Guastoni2023} trained their models on an inexpensive ``minimal" channel to reduce computational costs, yet reported significant performance drops when applying the trained model to the full channel.

One promising approach to alleviate the computational burden of numerical simulations is the use of \emph{multifidelity} strategies\citep{MF_review2023}. In this context, high-fidelity models provide accurate but computationally expensive representations of fluid dynamics, while low-fidelity models offer simplified, cheaper approximations by means of coarser discretization, reduced physics, or lower-dimensional formulations. Although low-fidelity models are less accurate, they are considerably faster to simulate. Multifidelity learning aims to combine the strengths of both types of models, leveraging low-fidelity models to accelerate the training process without sacrificing the accuracy provided by high-fidelity models.

In order to facilitate the use of multifidelity strategies in DRL, the concept of \emph{transfer learning} (TL) in DRL~\citep{Taylor2009, Zhu2023} can be employed as a powerful tool for transferring knowledge gained from one task or environment to another, thereby accelerating learning and improving performance in the target task. 
A widely adopted transfer learning strategy in neural networks is \emph{fine-tuning}, where a model is first pretrained on a source domain and then adapted to the target domain by continuing training, potentially with some layers frozen and others updated. Since its introduction by \citet{Hinton2006Reducing}, fine-tuning has become the standard approach in TL across various domains. 
In the context of multifidelity DRL, fine-tuning usually involves pretraining a policy (and possibly value) networks in a low-fidelity environment and continuing training in a high-fidelity environment. This method reuses the learned weights as initialization, thereby accelerating convergence on the target task. 

An application of multifidelity reinforcement learning using fine-tuning in fluid mechanics can be noted in the aerodynamic shape optimization of an airfoil~\cite{bholaMultifidelityReinforcementLearning2023}, where a policy is first trained on coarse-resolution simulations and then fine-tuned on higher-resolution cases. This strategy was shown to reduce computational costs by more than 30\%.

The potential of transfer learning to accelerate DRL-based flow control across varying Reynolds numbers has been explored in recent studies \citep{wangAcceleratingImprovingDeep2022, wangDeepReinforcementTransfer2023}. A DRL agent was first trained at a lower Reynolds number and subsequently fine-tuned on higher Reynolds cases. This approach achieved a significant reduction in training episodes and improved drag reduction, highlighting its suitability for increasingly complex flow conditions.

\citet{PolicyTransferReinforcement2023} demonstrated effective use of transfer learning to migrate DRL-based flow control policies from two-dimensional to three-dimensional environments. A policy trained on 2D flow past a cylinder was successfully transferred to 3D configurations, showing generalization capabilities. \citet{yanDeepReinforcementCrossdomain2025} adopted a mutual information-based knowledge transfer (MIKT) strategy combined with the Soft Actor-Critic (SAC) algorithm to address cross-domain transfer learning problems in active flow control. Their framework successfully enabled policy transfer from two-dimensional to three-dimensional bluff body flows with mismatched state and action spaces, highlighting the potential of DRL transfer learning for realistic engineering applications.

Despite the demonstrated success of transfer learning in DRL applications to fluid flows, most studies rely solely on fine-tuning as the transfer mechanism. While simple and often effective, fine-tuning exhibits notable limitations in reinforcement learning contexts and frequently fails to enable robust knowledge transfer across complex domains~\citep{Campos2021}. On the one hand, the pre-trained model may overfit to specific dynamics and rewards of the source environment, which can lead to suboptimal performance in the target environment. On the other hand, extensive fine-tuning of the entire network often results in ``catastrophic forgetting", the inadvertent loss of previously learned information. This inherent destructiveness hinders the model's ability to generalize across multiple tasks or environments.

To overcome these limitations, more structured transfer learning frameworks are required. One such method is the Progressive Neural Network (PNN) architecture, which avoids overwriting prior knowledge by explicitly freezing previously learned models and augmenting them with new trainable columns~\citep{PNN2016}. In the standard pretrain-and-finetune approach, it is typically assumed that the source and target tasks share substantial similarity. This overlap allows fine-tuning to adapt the network with relatively minor modifications~\citep{NIPS2014_532a2f85}. In contrast, PNNs do not rely on such assumptions. They are designed to handle scenarios where the source and target tasks may differ significantly, or even be unrelated.

PNNs have been successfully employed in various deep learning domains such as Atari games~\citep{PNN2016}, emotion recognition~\citep{gideon2017PNN}, visual classiﬁcation~\citep{ergun2021sparse}, robotics~\citep{PNN_Robotics}, and acoustic models~\citep{PNN_Acoustic}. Despite these advancements, to the best of the author's knowledge, progressive neural networks (PNNs) have not yet been explored for transfer learning in DRL-based flow control. This work presents the first application of PNNs in this context, aiming to overcome the limitations of fine-tuning and enable more robust and efficient knowledge transfer across fidelity levels and physical regimes.

The remainder of this paper is organized as follows. Section~\ref{sec:NN_TL} provides a brief overview of neural networks and transfer learning, while Section~\ref{sec:DRL} discusses fundamentals of the deep reinforcement learning (DRL) framework. Section~\ref{sec:method_transfer} introduces the transfer learning strategies considered in this work. The case study, including the numerical setup of the RL environment and agent, is described in Section~\ref{sec:KS}. Section~\ref{sec:results} presents and discusses the results. Finally, Section~\ref{sec:conclusion} summarizes the main findings and outlines directions for future work.

All the codes and cases developed for the current study are open-source in the GitHub repository \texttt{TL\_DRL}: \url{https://github.com/salehisaeed/TL_DRL}.

\section{Neural networks and transfer learning}
\label{sec:NN_TL}
DRL algorithms rely on deep neural networks (DNNs) as universal function approximators to estimate components of the agent, such as policies, value functions, or $Q$-functions. A DNN defines a parameterized mapping $f_\theta: \mathbb{R}^n \rightarrow \mathbb{R}^m$, where $\theta$ denotes the trainable parameters. Feedforward neural networks, the most common type of DNNs, are constructed by stacking multiple layers, each consisting of several neurons. A neuron computes a weighted sum of its inputs with an added bias and applies a nonlinear activation function, enabling the network to represent highly nonlinear mappings.

\begin{figure}[!ht]
	\centering
	\includegraphics[width=0.45\textwidth]{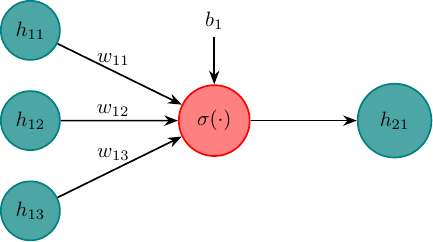}
	\caption{Schematic of a single neuron as the fundamental unit of a feedforward neural network. 
		Each input $h_{(ij)}$ is multiplied by a weight $w{ij}$, combined with a single bias $b_{i}$, and passed through a nonlinear activation function $\sigma(\cdot)$ to produce the output $h_{i+1}$. Here, the output of the neuron is therefore $h_{21} = \sigma \left( w_{11} h_{11} + w_{12} h_{12} +w_{13} h_{13} + b_1 \right)$.}
	\label{fig:neuron}
\end{figure}

This basic mechanism is illustrated in Fig.~\ref{fig:neuron}, which shows a single neuron receiving inputs from three neurons in the previous layer. Each input $h_{ij}$ is multiplied by a weight $w_{ij}$, and the weighted inputs are combined with a bias $b_i$. The result is passed through a nonlinear activation function $\sigma(\cdot)$ to produce the output $h_{(i+1)}$. Later in the paper, individual neurons are not shown for simplicity. Instead, each layer is represented as a box connected to the subsequent layer.

During the training process of a DNN, the parameters $\theta$ of the neural network correspond to the collection of all weights and biases. These parameters are optimized by minimizing a loss function.

Transfer learning~\cite{weiss2016survey} refers to the reuse of knowledge acquired in one domain (the source) to accelerate or improve learning in another domain (the target). In neural networks, knowledge is encoded in the learned parameters (weights and biases), and transfer can be achieved by initializing the target network with parameters trained in the source task, optionally fine-tuning parts or all of the network.

\section{Deep Reinforcement Learning}
\label{sec:DRL}

Reinforcement learning (RL) models sequential decision-making, based on the theory of Markov Decision Processes (MDPs), where the agent observes the state of the environment, selects actions, and receives feedback in the form of rewards. Over time, the agent aims to maximize the cumulative reward by learning a strategy, i.e., policy, that maps states to optimal actions. For a detailed theoretical foundation, readers are referred to \citet{sutton2018reinforcement}.

\begin{figure}[!ht]
	\centering
	\includegraphics[trim={0.45cm 0cm 0cm 0cm}, clip, width=0.5\textwidth]{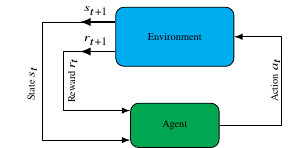}
	\caption{Simplified schematic of the DRL framework. A policy network observes the environment state, selects actions, and updates its parameters based on reward feedback.}
	\label{fig:RL}
\end{figure}

As illustrated in Fig.~\ref{fig:RL}, at each time step, the agent observes a state of the environment $s_t$ and selects an action $a_t$ according to its current policy $\pi(a_t | s_t)$. The environment responds by transitioning to a new state $s_{t+1}$ and providing a reward $r_t$. The goal of the agent is to maximize the expected return, typically defined as the discounted sum of rewards,
\begin{equation}
	G_t = \sum_{k=0}^{\infty} \gamma^k r_{k},
\end{equation}
where $\gamma \in [0, 1)$ is a discount factor that prioritizes immediate rewards over distant ones.

This study focuses on model-free RL, where the agent does not explicitly learn or use a model of the environment. This approach is particularly suitable for practical fluid dynamics applications, where the physics is chaotic and complex.

Model-free algorithms are further divided into on-policy and off-policy methods. On-policy methods improve the policy only using data generated by the current policy, while off-policy methods learn from data collected by different policies, allowing greater data efficiency and reuse of past experiences. In particular, off-policy methods typically incorporate experience replay, where past transitions are stored in a buffer and sampled during training. This is especially advantageous in computationally expensive environments like fluid simulations, where data collection is costly.

Actor–critic methods form a widely used class of algorithms in which the critic estimates the state–action value function
\begin{equation}
	Q^\pi(s,a) = \mathbb{E}\left[ r_t + \gamma \, Q^\pi(s_{t+1},a_{t+1}) \, \big| \, s_t = s, a_t = a \right],
\end{equation}
and the actor updates the policy based on these estimates. Deep reinforcement learning (DRL) extends this framework by parameterizing both actor and critic with deep neural networks, enabling RL to scale to high-dimensional, continuous state and action spaces.

Among DRL algorithms, the Soft Actor–Critic (SAC)~\citep{SAC} method is employed in this work. SAC is an off-policy, entropy-regularized actor–critic algorithm. The actor is trained to maximize a $Q$-augmented objective that balances reward and entropy, as
\begin{equation}
	J(\pi) = \mathbb{E}_{(s,a)\sim\pi}\left[ Q^\pi(s,a) - \alpha \log \pi(a|s) \right],
\end{equation}
where the second term corresponds to the policy entropy $\mathcal{H}(\pi(\cdot|s))$, weighted by the temperature parameter $\alpha$. A larger $\alpha$ encourages exploration, while a smaller $\alpha$ favors exploitation; in this work, $\alpha$ is automatically tuned during training. The SAC framework employs two $Q$-networks to approximate $Q^\pi(s,a)$, an actor network for the policy, and a target critic to stabilize learning.

\section{Multifidelity DRL using transfer learning}
\label{sec:method_transfer}

\begin{figure}[!hbt]
	\centering
	\begin{subfigure}{\textwidth}
		\centering
		\includegraphics[trim={0.55cm 0 0 0}, clip, width=1\textwidth]{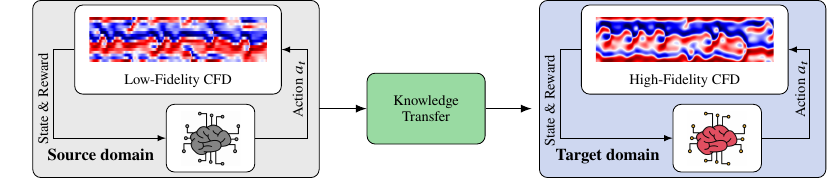}
	\end{subfigure}
	\par\bigskip
	\begin{subfigure}{0.6\textwidth}
		\includegraphics[trim={0.7cm 0.3cm 0.2cm 1.1cm}, clip, width=1\textwidth]{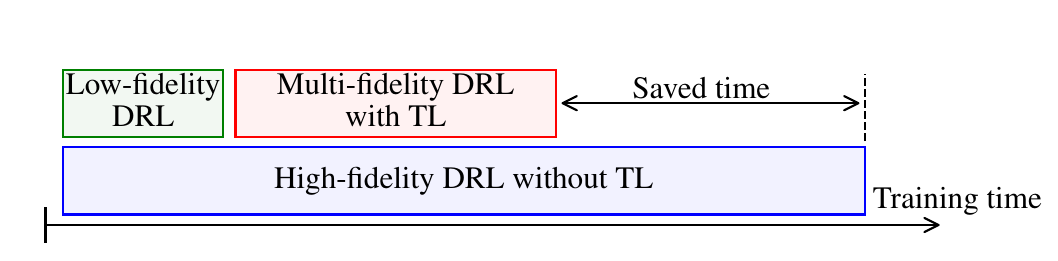}
	\end{subfigure}
	\caption{Conceptual illustration of the multifidelity DRL framework and its potential for computational acceleration. The top diagram shows knowledge transfer between low- and high-fidelity environments. The bottom panel depicts potential savings in training time through transfer learning. The CFD environments are represented by solutions of the Kuramoto–Sivashinsky (KS) equation, described in Section~\ref{sec:KS}, using different spatial discretizations.}
	\label{fig:MFRL}
\end{figure}

Fig.~\ref{fig:MFRL} provides a conceptual illustration of the multifidelity DRL framework considered in this study. The agent is first trained in a computationally inexpensive low-fidelity environment, where the mesh resolution and flow complexity are reduced. Once a preliminary policy is obtained, knowledge is transferred to a high-fidelity environment, where training is continued or adapted. The underlying assumption is that core control knowledge acquired in the source domain can accelerate learning in the more expensive target domain. The central challenge is how this knowledge, typically encoded in the parameters of deep neural networks, can be transferred and adapted across environments that differ in fidelity, physics, or control objectives.

The bottom panel of Fig.~\ref{fig:MFRL} highlights the motivation for this approach. By leveraging knowledge from the low-fidelity setting, the number of required training iterations in the high-fidelity domain can be substantially reduced. Low-fidelity simulations are typically inexpensive and can be run at negligible computational cost, making it feasible to pretrain a model extensively before transfer. When successful, this strategy yields significant savings compared to training a high-fidelity policy from scratch.

In the present work, multifidelity primarily refers to environments that differ in spatial discretization, i.e., the number of mesh points. Later sections also explore variations in physical regimes and control objectives, but mesh resolution serves as the main fidelity parameter in this study.

Two broad classes of transfer strategies are explored, namely, (i)~\emph{Fine-tuning}, which adapts a pretrained network by continuing training in the target environment, and (ii)~\emph{Progressive Neural Networks}, which explicitly separate source and target representations while enabling lateral transfer through learnable connections. 

The following subsections describe these strategies in detail and provide context for the systematic evaluation presented in Section~\ref{sec:results}.

\subsection{Fine-tuning strategies}
\label{sec:method_ft}

Fine-tuning is one of the most widely used strategies for transfer learning in deep reinforcement learning. In this approach, a policy trained in a source domain is reused in the target environment by continuing the training process from the same set of learned parameters. The rationale is that foundational features learned in the source domain may still be relevant in the target setting, allowing for faster convergence compared to training from scratch.

In the multifidelity DRL context, the source and target environments differ in resolution and potentially in physical behavior. While fine-tuning can be effective, it is also prone to catastrophic forgetting, where knowledge acquired in the source domain is overwritten during training in the target environment. This is particularly problematic when bidirectional transfer or knowledge retention is important.

Several variations of fine-tuning strategies are considered, based on which components of the agent are allowed to adapt.
\begin{itemize}
	\item \textbf{Full fine-tuning}: All parameters of the actor and critic networks are updated in the target environment.
	
	\item \textbf{Partial fine-tuning}: Only selected layers in the actor and critic networks (e.g., the last layer or final two layers) are updated, while earlier layers remain frozen. This encourages the reuse of low-level features while adapting higher-level representations to the target domain.
	
	\item \textbf{New-layer-only training}: The network is expanded with new layers. The existing layers from the source model are frozen, and only the newly added layers are trained in the target environment. This enforces strict feature reuse and isolates adaptation to the new capacity.
	
	\item \textbf{Full fine-tuning with new layers}: The network is expanded with new layers, and all layers, including both the original and new ones, are fine-tuned during training in the target environment. This provides more flexibility for adaptation while still initializing with source knowledge.
\end{itemize}

\begin{figure}[p]
	\centering
	\begin{subfigure}{0.24\textwidth}
		\centering
		\includegraphics[width=0.68\textwidth]{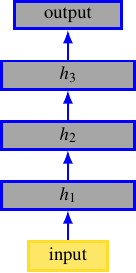}
		\caption{Scratch}
		\label{fig:MF_Strategies_scratch}
	\end{subfigure}
	\hfill
	\begin{subfigure}{0.24\textwidth}
		\centering
		\includegraphics[width=0.68\textwidth]{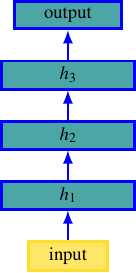}
		\caption{Fine-tune all}
		\label{fig:MF_Strategies_finetune_all}
	\end{subfigure}
	\hfill
	\begin{subfigure}{0.24\textwidth}
		\centering
		\includegraphics[width=0.68\textwidth]{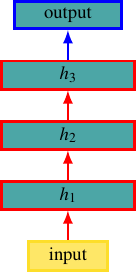}
		\caption{Fine-tune last}
		\label{fig:MF_Strategies_finetune_last}
	\end{subfigure}
	\hfill
	\begin{subfigure}{0.24\textwidth}
		\centering
		\includegraphics[width=0.68\textwidth]{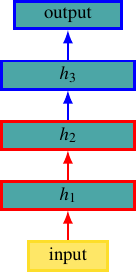}
		\caption{Fine-tune last two}
		\label{fig:MF_Strategies_finetune_lastTwo}
	\end{subfigure}
	\par\bigskip
	\begin{subfigure}{0.24\textwidth}
		\centering
		\includegraphics[width=0.68\textwidth]{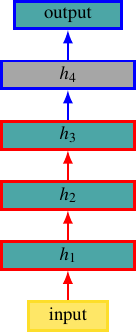}
		\caption{New layer (partial)}
		\label{fig:MF_Strategies_finetune_newLayer_1_freeze}
	\end{subfigure}
	\hfill    
	\begin{subfigure}{0.24\textwidth}
		\centering
		\includegraphics[width=0.68\textwidth]{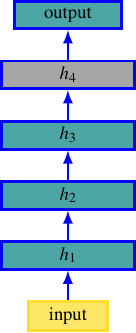}
		\caption{New layer (all)}
		\label{fig:MF_Strategies_finetune_newLayer_1}
	\end{subfigure}
	\hfill
	\begin{subfigure}{0.24\textwidth}
		\centering
		\includegraphics[width=0.68\textwidth]{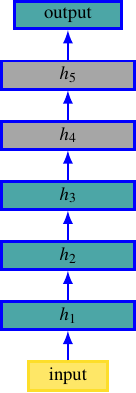}
		\caption{2 new layers (all)}
		\label{fig:MF_Strategies_finetune_newLayer_2}
	\end{subfigure}
	\hfill
	\begin{subfigure}{0.24\textwidth}
		\centering
		\includegraphics[width=0.68\textwidth]{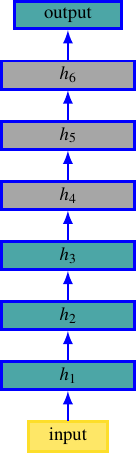}
		\caption{3 new layers (all)}
		\label{fig:MF_Strategies_finetune_newLayer_3}
	\end{subfigure}
	\par\bigskip
	\begin{subfigure}{\textwidth}
		\centering
		\includegraphics[width=0.85\textwidth]{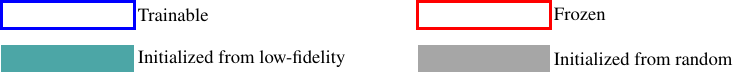}
	\end{subfigure}
	
	\caption{Multifidelity fine-tuning-based transfer learning strategies evaluated in this study. Each box represents a layer of the neural network architecture, corresponding to an input, hidden, or output layer with a specified number of neurons.
		(\subref{fig:MF_Strategies_scratch}) Training from scratch on the high-fidelity target environment.
		(\subref{fig:MF_Strategies_finetune_all}) Fine-tuning all layers of the pretrained low-fidelity model. 
		(\subref{fig:MF_Strategies_finetune_last}) Fine-tuning only the final output layer. 
		(\subref{fig:MF_Strategies_finetune_lastTwo}) Fine-tuning the final two layers. 
		(\subref{fig:MF_Strategies_finetune_newLayer_1_freeze}) Adding a new hidden layer and fine-tuning only the new and output layers, with the original layers frozen. 
		(\subref{fig:MF_Strategies_finetune_newLayer_1}) Adding a new hidden layer and fine-tuning all layers. 
		(\subref{fig:MF_Strategies_finetune_newLayer_2}) Adding two new layers and fine-tuning all layers. 
		(\subref{fig:MF_Strategies_finetune_newLayer_3}) Adding three new layers and fine-tuning all layers. 
		Trainable layers are outlined in blue and frozen layers in red. Layers are either initialized from the low-fidelity model or randomly.}
	\label{fig:MF_Strategies}
\end{figure}

Various fine-tuning-based transfer learning strategies used in this study are schematically illustrated in Fig.~\ref{fig:MF_Strategies}. Each rectangular box represents a layer in the neural network, with the color indicating whether the layer is trainable, frozen, initialized from the low-fidelity model, or randomly initialized.

These strategies are categorized based on which parts of the network are retained from the low-fidelity model and which parts are allowed to adapt during training in the high-fidelity environment. The configurations are designed to assess the trade-offs between knowledge reuse and model flexibility. The baseline strategy (single-fidelity),  shown in Fig.~\ref{fig:MF_Strategies_scratch}, is trained from scratch on the target high-fidelity environment, without access to prior low-fidelity knowledge. The baseline network architecture consists of three hidden layers, with the actor using $256$ units per layer and the critic using $128$ units per layer, both with Tanh activations. This configuration provides the reference against which all transfer learning strategies are evaluated. See Section~\ref{sec:KS}, for more information about the baseline architecture. As illustrated in Fig.~\ref{fig:MF_Strategies}, the remaining strategies either retain this baseline architecture or extend it by adding new hidden layers.

Fine-tuning strategies (Figs.~\ref{fig:MF_Strategies_finetune_all}--\ref{fig:MF_Strategies_finetune_lastTwo}) initialize the target policy using weights trained on a low-fidelity source environment and update all or a subset of those layers in the target setting. These approaches aim to leverage prior knowledge while enabling adaptation to higher resolution.
Alternatively, transfer via network expansion (Figs.~\ref{fig:MF_Strategies_finetune_newLayer_1_freeze}--\ref{fig:MF_Strategies_finetune_newLayer_3}) adds one or more new hidden layers to the existing architecture. In some variants, only the new and output layers are trained, enforcing strict feature reuse. In others, all layers are fine-tuned to allow full model adaptation. These expanded architectures increase representational capacity and allow for progressive refinement of features learned in the low-fidelity environment.

Note that in all numerical experiments, the same strategy is applied to both the actor and critic networks, unless otherwise specified.

\subsection{Progressive neural networks}
\label{sec:method_pnn}

This section introduces progressive neural networks, a transfer learning framework designed to preserve previously acquired knowledge when adapting to new tasks. The following subsections detail the core structure of PNNs, the use of adapter layers to improve lateral knowledge transfer, the intuitive method for analyzing the transfer dynamics, and different PNN strategies adopted for the numerical experiments.

\subsubsection{Network structure}

A PNN begins with a standard deep neural network, i.e., typically called a \emph{column} in this context, containing $L$ layers with hidden activations $h_i^{(1)} \in \mathbb{R}^{n_i}$, with $n_i$ being the number of neurons at layer $i$, and trainable parameters $\Theta^{(1)}$. The network is trained on a source task (Task 1). When transferring the trained network to a new target task (Task 2), a new column with the same network architecture is added and connected to the first column, where the parameters of the previous column $\Theta^{(1)}$ are frozen, while the new column parameters $\Theta^{(2)}$ are randomly initialized and remain trainable.

\begin{figure}[!htb]
	\centering
	\includegraphics[width=0.3\textwidth]{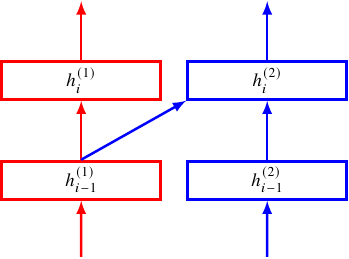}
	\caption{Simplified illustration of the progressive neural network mechanism. 
		Features from a frozen previous column (red) are passed through lateral connections into the trainable new column (blue), enabling transfer of previously learned representations without modifying the original parameters.}
	
	\label{fig:pnn_simple_schematic}
\end{figure}

As shown schematically in Fig.~\ref{fig:pnn_simple_schematic}, layer $h_i^{(2)}$ in the new column receives inputs from both the preceding layer of the same column $h_{i-1}^{(2)}$ and from the corresponding layer of the previously trained column $h_{i-1}^{(1)}$, i.e., lateral connections. These inter-column links are trainable and designed to enable the new network to reuse useful features extracted from the source domain.

PNNs are not limited to two-task settings and can be generalized to $K$ tasks. The general formulation for computing the activations at layer \( i \) in column \( k \) is given by
\begin{equation}
	h_i^{(k)} = f\left( W_i^{(k)} h_{i-1}^{(k)} + \sum_{j<k} U_i^{(k:j)} h_{i-1}^{(j)} \right),
	\label{eq:pnn_basic}
\end{equation}
where $W_i^{(k)} \in \mathbb{R}^{n_i^{(k)} \times n_{i-1}^{(k)}}$ denotes the weight matrix of layer $i$ within column \( k \) and
$U_i^{(k:j)} \in \mathbb{R}^{n_i^{(k)} \times n_{i-1}^{(j)}}$
represents the lateral connection weight matrix that connects the layer $i-1$ of column \( j \) to layer \( i \) of column \( k \). The function \( f \) is an activation function applied element-wise (e.g., $\mathrm{Tanh}$).

\subsubsection{Adapters}

In practice, the lateral connections in progressive networks are enhanced using non-linear adapter modules. Rather than using direct linear projections between columns, each lateral path is enhanced with a subnetwork to perform dimensionality reduction and adapt the transferred features.

Let \( h_{i-1}^{(<k)} = [h_{i-1}^{(1)}, \dots, h_{i-1}^{(j)}, \dots, h_{i-1}^{(k-1)}] \) represent the concatenated inputs from the previous layer across all preceding columns, i.e., anterior features. Each lateral connection is replaced with a single-layer feedforward adapter, consisting of a projection followed by a non-linear activation. To normalize the scale of different inputs, a learnable scalar gain \( \alpha_i^{(j)} \) is applied elementwise before the transformation.

The resulting hidden activation at layer \( i \) in column \( k \) is then computed as
\begin{equation}
	h_i^{(k)} = f\left[ W_i^{(k)} h_{i-1}^{(k)} + \sum_{j<k} U_i^{(k:j)} \, f\left( V_i^{(k:j)} \alpha_i^{(j)} h_{i-1}^{(j)} \right) \right]
	\label{eq:pnn_adapter}
\end{equation}
where \( V_i^{(k:j)} \) is a projection matrix connecting column $j$'s features to the adapter space and \( U_i^{(k)} \) maps the adapter output to the current layer in column $k$.

\begin{figure}[!tb]
	\centering
	\includegraphics[width=0.8\textwidth]{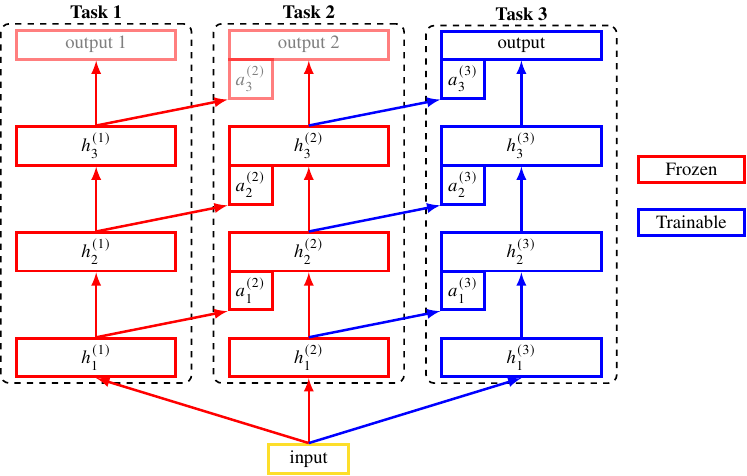}
	\caption{Schematic of a progressive neural network with three sequential tasks. Each task is represented by a distinct column comprising several hidden layers. Only the column corresponding to the latest task is trainable (blue), and the previous columns are kept frozen (red).
		Once a task is trained, its corresponding column (shown in red) is frozen. Lateral connections include adapter modules \( a_i^{(k)} \), which project and transform features from previous columns to support transfer without overwriting prior knowledge.}
	\label{fig:pnn_schematic}
\end{figure}

Fig.~\ref{fig:pnn_schematic} illustrates a schematic of a PNN with three columns ($K=3$) corresponding to three sequential tasks. The third column, responsible for the final task, is connected laterally to both earlier columns and can utilize their frozen features through trainable adapter paths.

\subsubsection{Transfer analysis of PNN}
\label{sec:transfer_aps}

To systematically assess the contribution of columns in a progressive network, the intuitive Average Perturbation Sensitivity (APS) method, proposed by \citet{PNN2016}, can be adopted. APS provides a quantitative measure of the impact of each layer of each column on the final performance by introducing controlled Gaussian noise into individual layers and evaluating the resulting performance degradation. 

In this context, $\Lambda_i^{(k)}$ represents the precision of the Gaussian noise injected at layer $i$ of column $k$, defined as the inverse of the noise variance, that results in a 50\% drop in task performance, i.e.,
\begin{equation}
	\Lambda_i^{(k)} = \frac{1}{\sigma_i^2{(k)}},
	\label{eqn:lambda}
\end{equation}
where $\sigma_i^2{(k)}$ is the variance of the injected Gaussian noise. The APS score for a given layer $i$ in column $k$ is defined as:
\begin{equation}
	\text{APS}(i, k) = \frac{\Lambda_i^{(k)}}{\sum_j \Lambda_i^{(j)}}.
	\label{eqn:aps}
\end{equation}
Thus, APS provides a normalized measure of the sensitivity of a given layer across all columns, reflecting the relative importance of each column in a given layer to the overall task performance.

\subsubsection{PNN strategies}
\label{sec:MF_pnn_strategies}

\begin{figure}[!tb]
	\centering
	\begin{subfigure}{0.45\textwidth}
		\centering
		\includegraphics[width=0.7\textwidth]{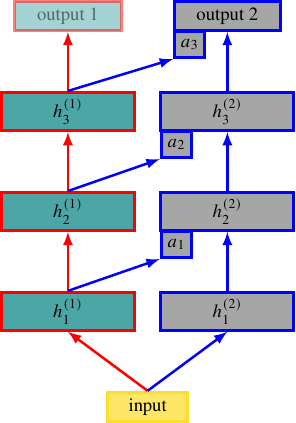}
		\caption{Standard PNN}
		\label{fig:MF_Strategies_pnn}
	\end{subfigure}
	\hfill
	\begin{subfigure}{0.45\textwidth}
		\centering
		\includegraphics[width=0.7\textwidth]{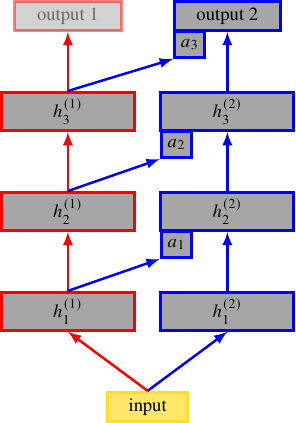}
		\caption{PNN with random source column}
		\label{fig:MF_Strategies_pnn_random}
	\end{subfigure}
	\par\bigskip
	\begin{subfigure}{\textwidth}
		\centering
		\includegraphics[width=0.85\textwidth]{Figures/MF/Strategies/legend_hor}
	\end{subfigure}    
	\caption{PNN-based transfer learning strategies. (\subref{fig:MF_Strategies_pnn}) Standard PNN with a frozen source column trained in a low-fidelity environment, and a newly initialized target column. (\subref{fig:MF_Strategies_pnn_random}) PNN with a randomly initialized frozen source column, used as a reference to evaluate the importance of meaningful source features.}
	\label{fig:MF_Strategies_PNN}    
\end{figure}

To evaluate the effectiveness and robustness of PNN-based transfer learning, multiple strategies are considered. These differ in how the source and target columns are initialized and whether any components beyond the actor are reused. The following PNN scenarios are considered.
\begin{itemize}
	
	\item \textbf{Standard PNN}: The baseline configuration is the standard PNN strategy, illustrated in Fig.~\ref{fig:MF_Strategies_pnn}. In this setup, the actor network trained on the low-fidelity environment forms the first (source) column and is frozen. A new target column is added and trained on the high-fidelity environment, with lateral adapter connections enabling knowledge transfer. Only the new column and adapters are updated during training. The critic networks are trained from scratch and are not connected to the source model.

	\item \textbf{PNN with random source column}: To isolate the contribution of meaningful source features, a control setup is tested in which the first column is randomly initialized and frozen (Fig.~\ref{fig:MF_Strategies_pnn_random}). This configuration evaluates whether transfer performance truly stems from previously acquired low-fidelity knowledge or is simply an artifact of architectural expansion.
	
	\item \textbf{PNN with fine-tuned critic}: A third configuration (not shown) allows fine-tuning of the critic networks while keeping the actor in standard PNN form. This hybrid strategy assesses whether reusing the critic structure from the low-fidelity model can benefit convergence or if it introduces limitations due to fidelity mismatch.
	
\end{itemize}

\section{Control of chaotic Kuramoto--Sivashinsky system}
\label{sec:KS}

This section presents the reinforcement learning setup of the test case used to evaluate the proposed transfer learning strategies. The Kuramoto--Sivashinsky (KS) system is introduced as the benchmark environment, followed by the control formulation and learning framework.

\subsection{Kuramoto--Sivashinsky system}
\label{sec:KS_env}

The KS equation is a canonical model that was originally derived in studies of flame front stability~\citep{sivashinskyNonlinearAnalysisHydrodynamic1977} and reaction-diffusion processes~\citep{kuramotoPersistentPropagationConcentration1976}. The KS system is among the simplest nonlinear Partial Differential Equations (PDEs) known to exhibit spatiotemporal chaos that resembles turbulence~\citep{cvitanovicStateSpaceGeometry2010a}. Thus, the system has emerged as a benchmark for evaluating control algorithms, due to its rich non-linear chaotic dynamics and complex flow structures, while having an affordable computational cost~\citep{Bucci2019, Zeng2021, zengDatadrivenControlSpatiotemporal2022, Paris2023, wernerNumericalEvidenceSample2024, peitzDistributedControlPartial2024}.

In the formulation of the one-dimensional KS PDE, the time evolution of the flow velocity, $u(x,t)$, on a periodic spatial domain of $x \in [0, L]$, is described by 
\begin{equation}
	\frac{\partial u}{\partial t} + u \frac{\partial u}{\partial x} + \frac{\partial^2 u}{\partial x^2} + \lambda \frac{\partial^4 u}{\partial x^4} = \phi(x,t), \qquad x(0,t) = x(L,t).
	\label{eqn:KS_1D}
\end{equation}
The equation consists of a non-linear convective term ($uu_x$), a second-order viscous term ($u_{xx}$), a fourth-order hyperviscous term ($u_{xxxx}$) with a hyperviscosity $\lambda$, and a source (forcing) term on the right-hand side $\phi(x,t)$. In the original formulation of the one-dimensional KS, the source term and hyperviscosity are $\phi(x,t)=0$ and $\lambda=1$.

The system exhibits chaotic behavior for sufficiently large domain lengths $L$, characterized by the development of unstable modes and complex dynamics. In this study, the domain length is set to $L=22$, which is large enough to capture turbulent dynamics~\citep{cvitanovicStateSpaceGeometry2010a} yet small enough to remain computationally efficient for benchmarking studies.

The spatial periodicity of the one-dimensional KS problem makes it well-suited for Fourier spectral methods, as the periodic boundary conditions allow for a natural decomposition of the solution into Fourier modes. The numerical simulations are conducted using a range of spatial resolutions, with the number of Fourier modes \( N \) varied to represent different fidelity levels (e.g., \( N = 16, 32, 64, 128 \)). A third-order semi-implicit Runge--Kutta method is used for time integration, where the linear terms are treated implicitly and the nonlinear term explicitly. The simulation time step is fixed at \( \Delta t_\mathrm{solution} = 0.05 \). This numerical setup follows the approach outlined by \citet{Bucci2019} and utilizes code from the \texttt{pyKS} package \citep{JswhitPyksData}.

\begin{figure}[!t]
	\centering
	\includegraphics[trim={0 0.35cm 0 0.2cm}, clip, width=.8\linewidth]{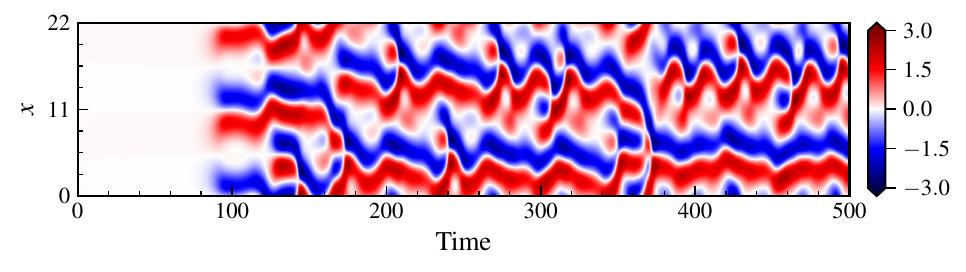}
	\caption{Numerical solution of the 1D KS system initialized with an infinitesimal perturbation given by \( u(x,0) = 10^{-8} \cdot \mathcal{N}(0, 1) \), where \(\mathcal{N}(0,1)\) denotes standard Gaussian white noise. The simulation is performed on a periodic domain of length \( L = 22 \) using \( N = 128 \) Fourier modes. It is observed that an infinitesimal perturbation quickly grows due to the chaotic nature of the system.}
	\label{fig:KS_Baseline}
\end{figure}

The KS equation has the trivial solution of $u_0(x,t)=0$. However, when the system is initialized with a significantly small white noise perturbation, even a small disturbance can quickly grow. Fig.~\ref{fig:KS_Baseline} shows the numerical solution of the KS system initialized with a random Gaussian noise of magnitude $10^{-8}$, i.e., 
\begin{equation}
	u(x,0) = 10^{-8} \cdot \mathcal{N}(0, 1).
	\label{eqn:KS_initial_condition}
\end{equation}
The infinitesimal perturbation grows exponentially, and after only $t=100$ time units, the system reaches a saturated state, indicating the chaotic nature of the KS equation.

\begin{figure}[!ht]
	\centering
	\includegraphics[trim={0 0.35cm 0 0.2cm}, clip, width=0.7\linewidth]{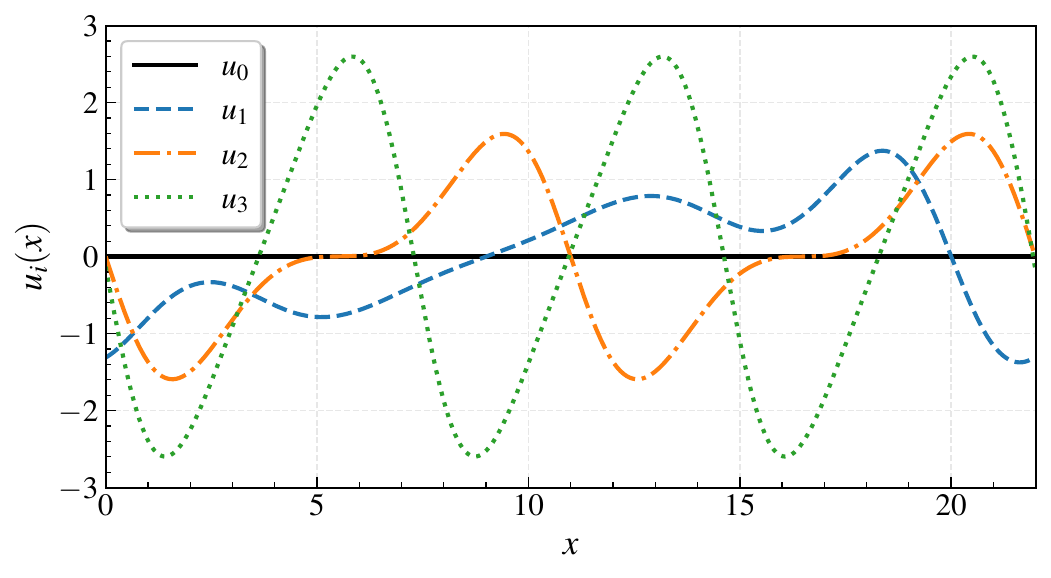}
	\caption{Trivial solution (\( u_0 \)) and three steady-state solutions ($\partial u / \partial t = 0$), namely, \( u_1 \), \( u_2 \), and \( u_3 \), of the KS system for \( L = 22 \).}
	\label{fig:KS_steady_solutions}
\end{figure}

For the spatial domain of $L=22$, the KS equation shows three steady-state solutions, $u(x,t)=u_q(x)$, namely, $u_1$, $u_2$, and $u_3$ (shown in Fig.~\ref{fig:KS_steady_solutions}), and two traveling waves due to its translational symmetry~\citep{cvitanovicStateSpaceGeometry2010a}.

\subsection{Control mechanism and RL framework}
\label{sec:KS_control}

The problem of controlling the KS system is addressed by introducing the source term, $\phi(x,t)$, to the right-hand side of Eq.~\eqref{eqn:KS_1D}. This source term represents the control actuators, which impose Gaussian-shaped forcing functions onto the system. The source term is defined as
\begin{equation}
	\phi(x,t) = \sum_{i = 0}^{n-1} a_i(t) \frac{1}{\sqrt{2 \pi \sigma}} \exp{\left( -\frac{(x - x_i^{\mathrm{act}})^2}{2 \sigma^2} \right)},
\end{equation}
where $n$ is the number of actuators, $x_i^{\mathrm{act}}$ is their corresponding location (the center of the Gaussian kernel), $a_i(t)$ is the time-dependent amplitude of the imposed forcing source term, and $\sigma$ controls the width of the Gaussian forcing. Here, four actuators are considered at the locations of $x_i^{\mathrm{act}} \in \{ 0, 1, 2, 3 \} L/4 $ with  $\sigma=0.4$. The amplitude values are decided by the controller at each control time step in $a_i(t) \in [-0.5, 0.5]$.

In practical flow-control applications, the full flow field is rarely available. Instead, feedback is obtained from a limited number of local measurements. Motivated by this, the control mechanism in the present study operates on partial observations of the environment. Eight sensors, positioned at $x_i^{\mathrm{act}} \in \{ 1, 3, 5, \cdots, 15 \} L/16 $ observe the flow field and provide feedback to the controller at each control time step. Based on this feedback, the controller computes new actions to adjust the system. The control time step is set to $\Delta t_\mathrm{control} = 5 \Delta t_\mathrm{solution} = 0.25$, as a compromise between the timescale of the flow dynamics and the system's response time~\citep{Paris2023}. Each episode consists of $1024$ control time steps.

The reward function is designed to encourage the agent to steer the system toward one of the steady-state solutions shown in Fig.~\ref{fig:KS_steady_solutions}. At each control step, the deviation of the current flow field \( u(x,t) \) from a fixed reference state \( u_\mathrm{ref}(x) \) is measured using the \( L^2 \)-norm. The reward is defined as
\begin{equation}
	r_t = 1 - \frac{\| u(t) - u_\mathrm{ref} \|_2}{\bar{d}_0},
	\label{eqn:KS_reward}
\end{equation}
where \( \bar{d}_0 \) denotes the time-averaged \( L^2 \)-distance between the uncontrolled KS solution and the same reference state. This normalization ensures the highest reward of \( r_t = 1 \) when the system exactly matches the reference profile, and \( r_t \approx 0 \) when the controlled behavior resembles the uncontrolled baseline. The reward may become negative if the agent drives the system further away from the target than the baseline dynamics. In this work, the reference state is chosen as the first steady-state solution \( u_1(x) \), unless otherwise specified.

As outlined in Section~\ref{sec:DRL}, flow control is performed using the Soft Actor–Critic (SAC) algorithm, implemented through the Stable-Baselines3 (SB3) library~\citep{stable-baselines3}. The agent is trained with a replay buffer of size $10^5$ samples. During training, the agent collects experience every $100$ environment steps and performs $200$ gradient updates per collection step, with a training batch size of $256$. The discount factor is set to \( \gamma = 0.97 \), and the soft target update coefficient is \( \tau = 0.005 \). The entropy coefficient is automatically tuned during training.

The actor and critic networks are independently parameterized, with a baseline architecture that is used unless otherwise specified. The actor (policy) network consists of three hidden layers with 256 units each, and the critic (Q-function) networks use three hidden layers with 128 units. All layers employ the hyperbolic tangent (Tanh) activation function. This baseline configuration provides a reference point for the subsequent transfer learning strategies, where variations such as adding or freezing layers are introduced. The choice of hyperparameters was guided by standard benchmarks and found to offer a good balance between stability and learning speed for the chaotic KS control task. The details of the hyperparameter tuning process are omitted here for brevity.

\section{Results and discussion}
\label{sec:results}

This section presents and analyzes the key results of the study. The performance of the reinforcement learning agent is first evaluated in single-fidelity environments to establish baseline behavior. The effect of different fine-tuning strategies is then examined in a multifidelity setting. Subsequently, the performance and transfer characteristics of PNNs are assessed and compared with conventional approaches.

All results presented in this section are based on multiple independent trials to ensure statistical robustness and reproducibility. Each experiment was repeated at least four times (in some cases, six times) with different random seeds. Learning curves are reported as the mean performance across trials, with shaded regions representing \( \mu \pm \sigma \), where \( \mu \) is the mean and \( \sigma \) is the standard deviation.

\subsection{Single fidelity learning}
\label{sec:SF}

The learning performance of the SAC agent is first evaluated in environments of varying spatial resolution. In this single-fidelity setting, the agent interacts with only one environment configuration during training, without any transfer or reuse of knowledge from other fidelities.

\begin{figure}[!b]
	\centering
	\includegraphics[trim={0 0.25cm 0 0.2cm}, clip, width=0.7\linewidth]{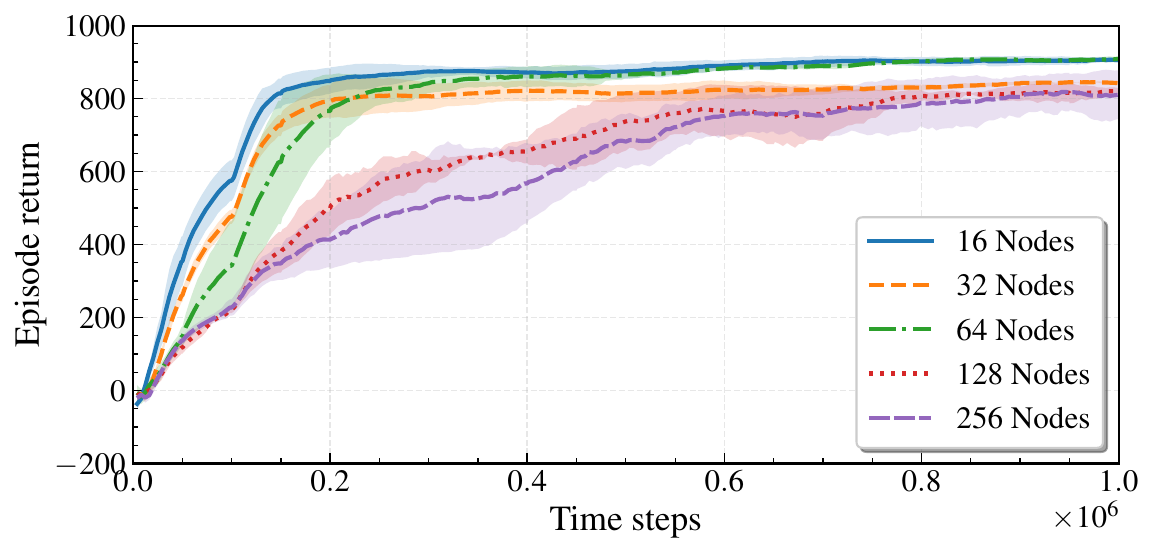}
	\caption{Learning curves for agents trained on single-fidelity environments with different numbers of discretization nodes, $N \in \{16, 32, 64, 128\}$. The curves show mean episode returns averaged over multiple trials, with shaded regions representing a standard-deviation region ($\pm\sigma$).}
	\label{fig:sf_convergence}
\end{figure}

Fig.~\ref{fig:sf_convergence} shows the learning curves for agents trained on environments characterized by different numbers of discretization nodes, $N \in \{16, 32, 64, 128\}$. %
The curves show the mean episode returns averaged over multiple trials, with shaded regions representing a standard deviation region ($\pm\sigma$).

Across most fidelities ($N < 128$), a two-phase convergence characteristic is observed. Initially, the agent exhibits a sharp increase in episode return, followed by a much slower rate of improvement, referred to as a \emph{knee} point. The sharp early rise likely corresponds to the agent learning the fundamental principles of control, such as suppressing large-scale instabilities and establishing a basic input-output relationship. The slower phase reflects more nuanced policy refinement, which is inherently more difficult and slower to converge.

As expected, convergence is significantly faster in lower-fidelity environments. The agent trained with $N=16$ reaches stable performance within the first $2 \times 10^5$ time steps. In contrast, learning with $N=128$ is considerably slower and does not exhibit a clear knee point, likely due to the increased dimensionality and complexity of the underlying dynamics.

\begin{figure}[!tb]
	\centering
	\begin{subfigure}{0.48\textwidth}
		\centering
		\includegraphics[trim={0 0.35cm 0 0.2cm}, clip, width=\textwidth]{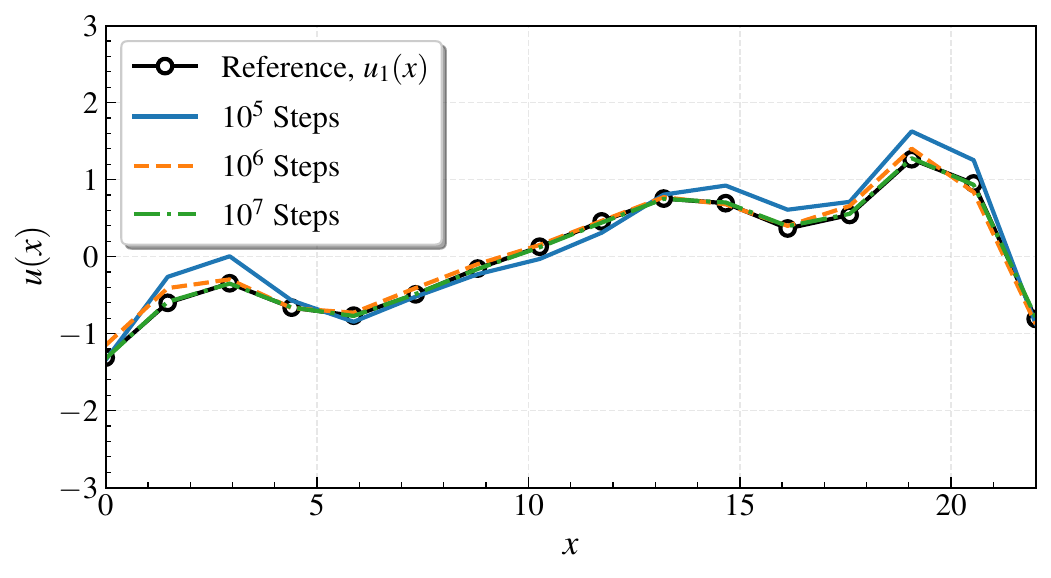}
		\caption{$N=16$}
		\label{fig:sf_u_final_16}
	\end{subfigure}
	\hfill
	\begin{subfigure}{0.48\textwidth}
		\centering
		\includegraphics[trim={0 0.35cm 0 0.2cm}, clip, width=\textwidth]{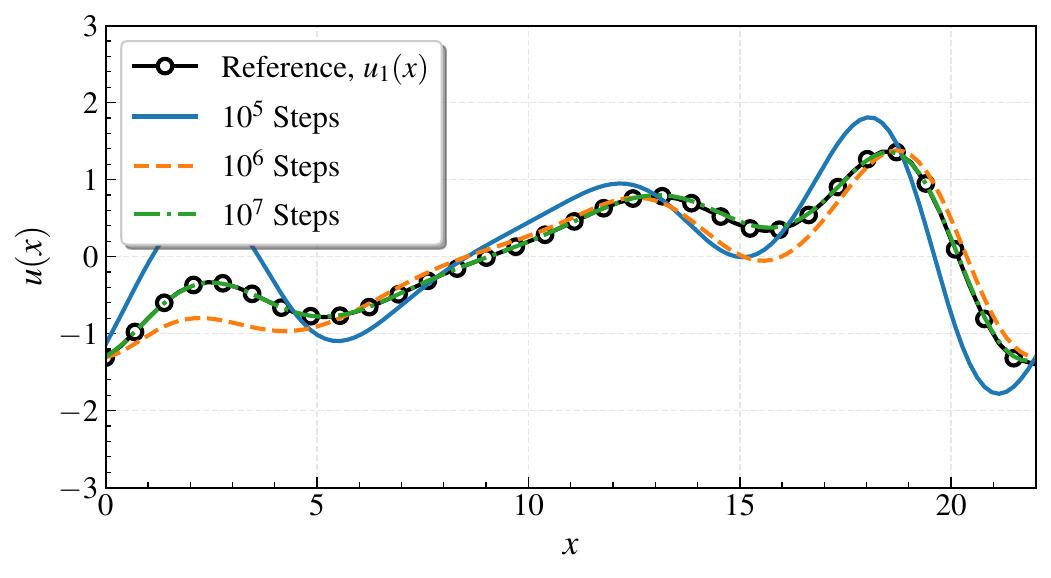}
		\caption{$N=128$}
		\label{fig:sf_u_final_128}
	\end{subfigure}
	\caption{Final velocity profiles \( u(x) \) of the controlled flow using single-fidelity agents trained for different durations. (\subref{fig:sf_u_final_16}) \( N = 16 \); (\subref{fig:sf_u_final_128}) \( N = 128 \).}
	\label{fig:sf_u_final}    
\end{figure}

\begin{figure}[!tb]
	\begin{subfigure}{0.495\textwidth}
		\centering
		\includegraphics[trim={0 0.35cm 0 0.2cm}, clip, width=\textwidth]{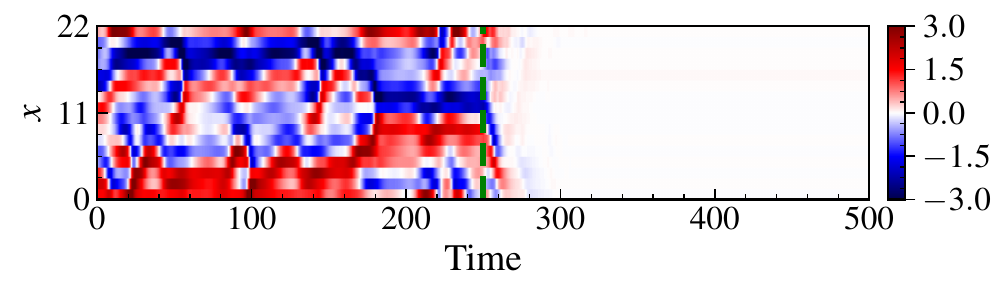}
		\caption{$N=16$}
		\label{fig:sf_contours_16}
	\end{subfigure}
	\hfill
	\begin{subfigure}{0.495\textwidth}
		\centering
		\includegraphics[trim={0 0.35cm 0 0.2cm}, clip, width=\textwidth]{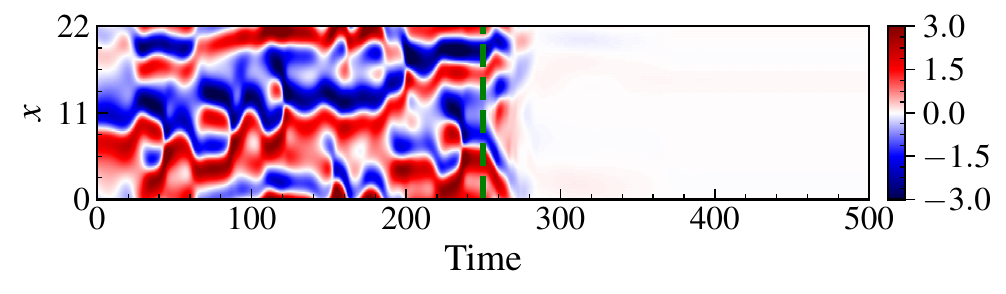}
		\caption{$N=128$}
		\label{fig:sf_contours_128}
	\end{subfigure}    
	\caption{Spatiotemporal evolution of the deviation field $u(x,t) - u_1(x)$ for environments with (\subref{fig:sf_contours_16}) $N=16$ and (\subref{fig:sf_contours_128}) $N=128$, suing a fully trained agent. The controller is activated at $t = 250$ (green dashed line).}
	\label{fig:sf_contours}
\end{figure}

The spatial structure of the controlled flow at various stages of training is examined in Fig.~\ref{fig:sf_u_final}, which presents instantaneous velocity profiles $u(x)$ compared to the target solution $u_{1}(x)$ for agents trained for different durations in environments with $N=16$ and $N=128$. At $N=16$, the agent closely matches the reference profile after a relatively small number of time steps (e.g., $10^5$), with only marginal improvements observed from additional training. In contrast, the agent trained at $N=128$ requires significantly more training to suppress finer-scale deviations, and acceptable agreement with $u_{1}(x)$ is only achieved after $10^6$ time steps.

Spatiotemporal contour plots of the deviation field $u(x,t) - u_{1}(x)$ are shown in Fig.~\ref{fig:sf_contours}, where white regions indicate minimal deviation and thus accurate control. The controller becomes active at time $t = 250$. In both cases, the agents rapidly achieve broad suppression of the deviation field following activation. However, the qualitative features and level of detail differ considerably. For \( N = 16 \), the low spatial resolution fails to capture the finer-scale structures present in the solution. However, the agent succeeds in driving the overall flow toward the desired profile.

\subsection{Transfer learning with fine-tuning strategies}
\label{sec:transfer}

Transfer scenarios from $N=16$ and $N=32$ to $N=128$ are considered, comparing the learning performance of transferred agents with those trained from scratch at the higher fidelity. Transfer is performed by loading the parameters of the low-fidelity agent after a given number of time steps and continuing training in the high-fidelity environment.

The transfer is initialized by loading all network weights from the pretrained SAC agent, including the policy's actor and both Q-networks (critic). Depending on the transfer strategy, the architecture used for the high-fidelity environment may either match the low-fidelity agent or include modifications such as additional layers. All strategies considered here are described in detail in Section~\ref{sec:method_ft}.

\subsubsection{Effect of Low-fidelity training and overfitting}
\label{sec:transfer_low_fidelity_training}

In this section, the influence of the duration of pretraining in low-fidelity environments on the performance of transfer learning to a high-fidelity target is investigated. Both the actor and critics of the SAC agent are fine-tuned in the target environment without freezing or modifying any layers, corresponding to the \emph{fine-tune-all} approach illustrated in Fig.~\ref{fig:MF_Strategies_finetune_all}.

\begin{figure}[!tb]
	\centering
	\begin{subfigure}{0.495\textwidth}
		\centering
		\includegraphics[trim={0 0.25cm 0 0.2cm}, clip, width=\textwidth]{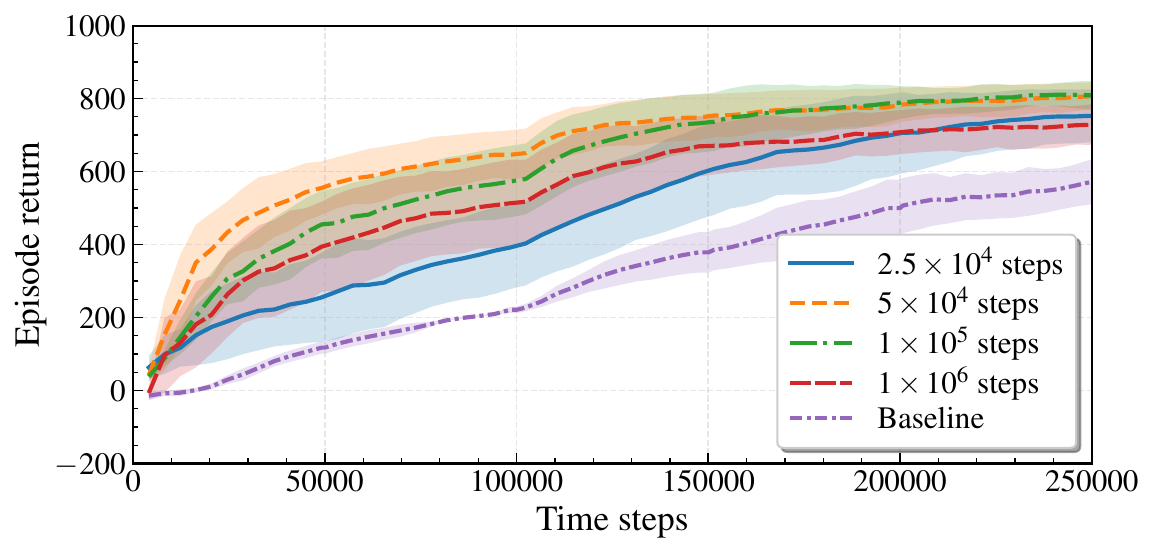}
		\caption{Low-fidelity $N=16$}
		\label{fig:convergene_startstep_16}
	\end{subfigure}
	\hfill
	\begin{subfigure}{0.495\textwidth}
		\centering
		\includegraphics[trim={0 0.25cm 0 0.2cm}, clip, width=\textwidth]{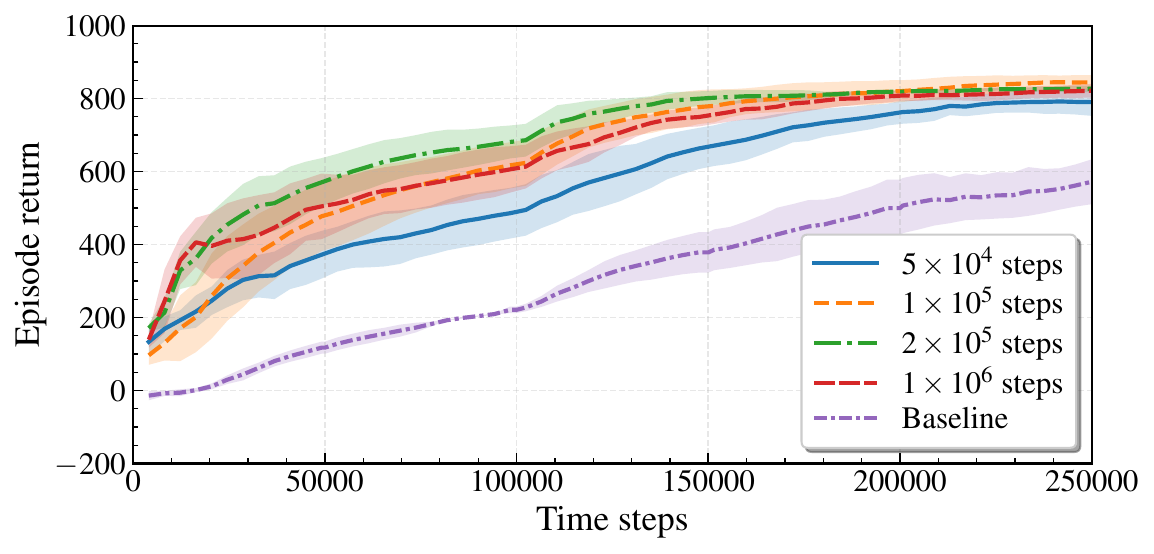}
		\caption{Low-fidelity $N=32$}
		\label{fig:convergene_startstep_32}
	\end{subfigure}
	\caption{Learning curves of the multifidelity model after transferring from low-fidelity trained for different durations (time steps). Entire actor and critic networks are fine-tuned. Each curve corresponds to a different pretraining step on the low-fidelity environment before transfer. (\subref{fig:convergene_startstep_16}) Transfer from $N=16$ to $N=128$ and (\subref{fig:convergene_startstep_32}) transfer from $N=32$ to $N=128$.}
	\label{fig:convergene_startstep}
\end{figure}

Fig.~\ref{fig:convergene_startstep} presents the multifidelity learning curves, illustrating the learning performance of the high-fidelity environment after transferring from models trained on low-fidelity setups for varying numbers of time steps. It is observed that transfer from pretrained models accelerates convergence compared to training from scratch (baseline). However, the learning seems to be very sensitive to the amount of low-fidelity pretraining. Increasing the amount of low-fidelity training does not always improve early learning on the target task. There appears to be an optimal amount of low-fidelity training that maximizes transfer performance. For instance, in Fig.~\ref{fig:convergene_startstep_16}, the model pretrained for $5 \times 10^4$ steps on $N=16$ yields the steepest initial learning, while in Fig.~\ref{fig:convergene_startstep_32}, the model pretrained for $2 \times 10^5$ steps on $N=32$ outperforms the others. This suggests that overfitting may occur during extended low-fidelity training, leading to suboptimal performance after transfer.

Although the initial learning rates are higher for these optimally pretrained models, the differences in final return after $2.5 \times 10^5$ time steps are less significant across all multifidelity models. Therefore, it is useful to assess the transfer performance more quantitatively.

\begin{figure}[!tb]
	\centering
	\begin{subfigure}{0.495\textwidth}
		\centering
		\includegraphics[trim={0 0.25cm 0 0.2cm}, clip, width=\textwidth]{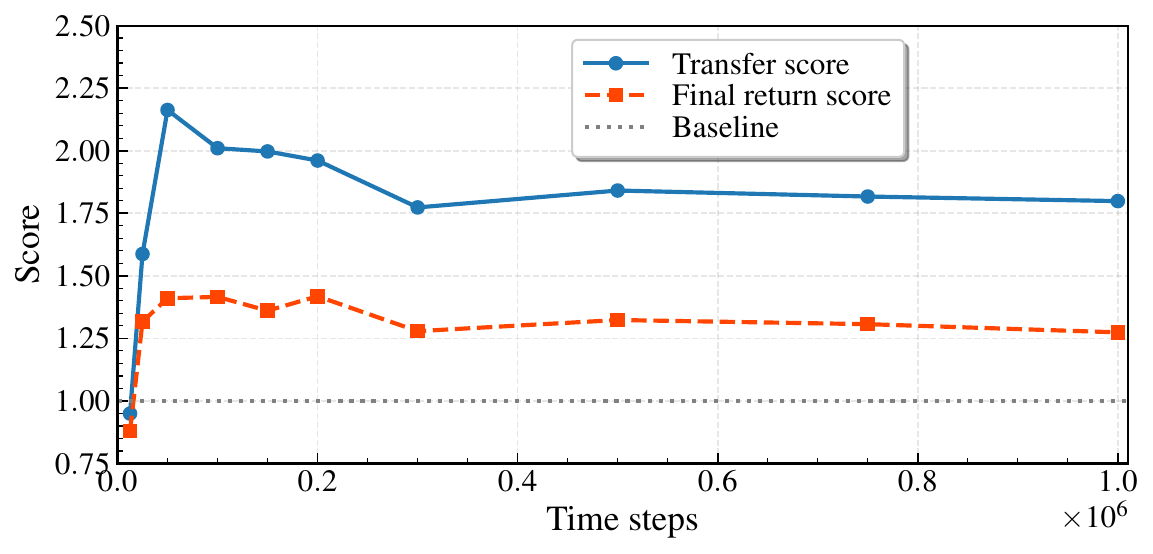}
		\caption{Low-fidelity $N=16$}
		\label{fig:score_startstep_16}
	\end{subfigure}
	\hfill
	\begin{subfigure}{0.495\textwidth}
		\centering
		\includegraphics[trim={0 0.25cm 0 0.2cm}, clip, width=\textwidth]{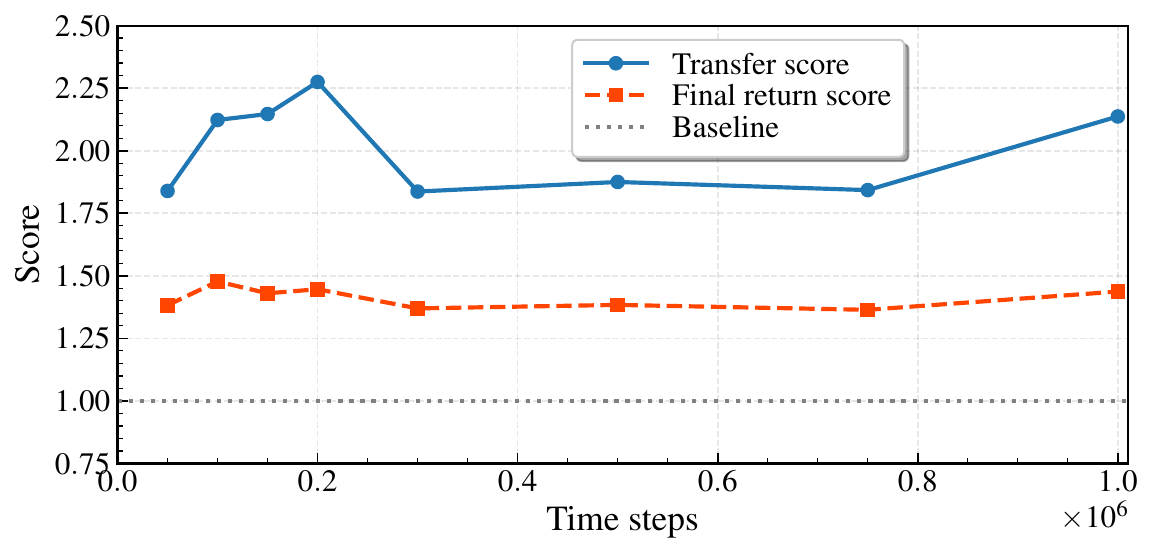}
		\caption{Low-fidelity $N=32$}
		\label{fig:score_startstep_32}
	\end{subfigure}
	\caption{Effect of low-fidelity pretraining duration on the transfer learning scores. (\subref{fig:convergene_startstep_16}) Transfer from $N=16$ to $N=128$ and (\subref{fig:convergene_startstep_32}) transfer from $N=32$ to $N=128$.}
	\label{fig:score_startstep}
\end{figure}

To that end, Fig.~\ref{fig:score_startstep} presents two performance scores as functions of the pretraining duration: the \emph{transfer score} and the \emph{final return score}. The transfer score is defined as the area under the learning curve, normalized by the area under the baseline curve. This score quantifies the effectiveness of transfer learning in accelerating convergence. The final return score is simply the average return achieved after $2.5 \times 10^5$ time steps, providing a measure of the ultimate performance of the transferred agent.

The transfer score indicates the existence of an optimal pretraining duration, with peak performance achieved at $5 \times 10^4$ steps for $N=16$ and $2 \times 10^5$ steps for $N=32$, at $2.15$ and $2.25$, providing $115\%$ and $125\%$ performance enhancements, respectively. Beyond these points, the transfer score declines and saturates, showing no improvement with further pretraining. In contrast, the final return score exhibits a more stable trend and is not significantly influenced by the amount of low-fidelity training. This suggests that while transfer learning can substantially speed up convergence, the final performance is relatively insensitive to the pretraining duration.

In the remainder of the paper, the observed optimal pretraining durations of $5 \times 10^4$ and $2 \times 10^5$ time steps for $N=16$ and $N=32$, are used, respectively, for all transfer learning experiments, unless otherwise specified.

\subsubsection{Effect of transfer learning strategies}
\label{sec:transfer_strategies_effect}

The impact of different fine-tuning strategies on the performance of the transferred SAC agent is investigated in this section. These strategies, illustrated in Fig.~\ref{fig:MF_Strategies}, differ in how the actor and critic networks are updated during transfer, including approaches with partial freezing of layers or the addition of new layers.

\begin{figure}[!htbp]
	\centering
	\includegraphics[trim={0 0.25cm 0 0.2cm}, clip, width=0.7\textwidth]{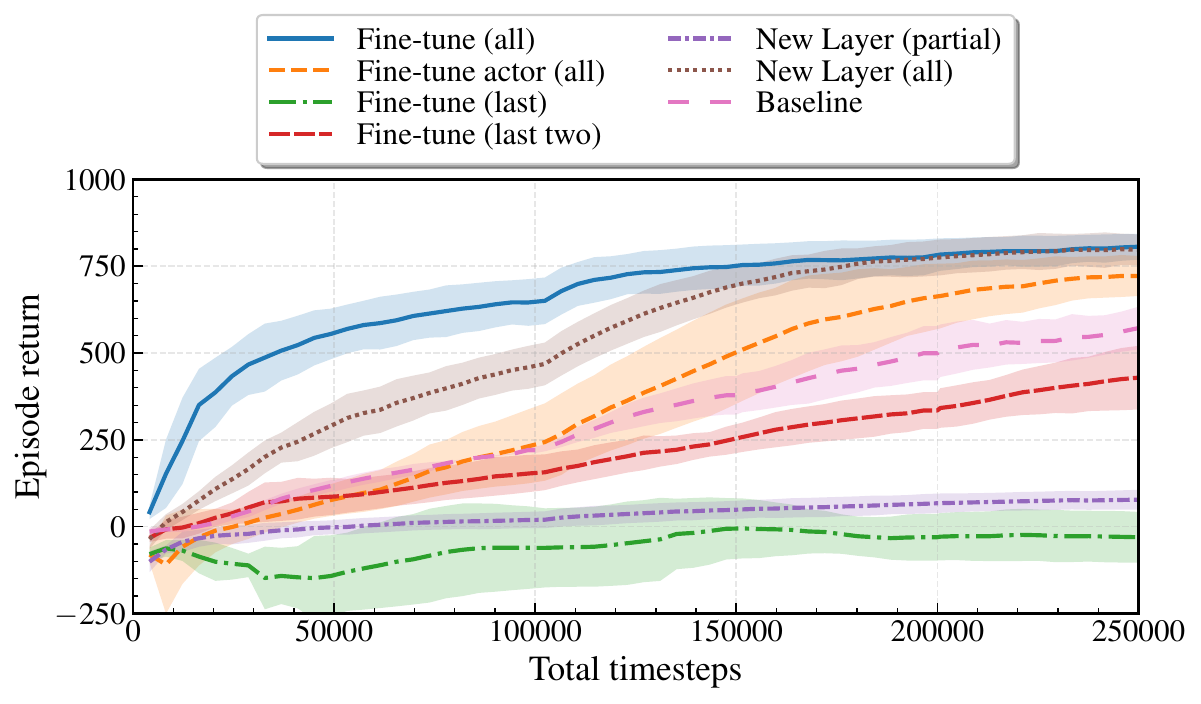}
	\caption{Learning curves for different multifidelity transfer learning strategies from $N=16$ to $N=128$.}
	\label{fig:strategy_comparison_learning_curves}
\end{figure}

Fig.~\ref{fig:strategy_comparison_learning_curves} presents the learning curves for various mutifildiety strategies transferred from $N=16$ to $N=128$, including fine-tuning all layers, fine-tuning only the actor network, partially fine-tuning the networks, and strategies that introduce new layers. It should be noted that in this figure, only the strategy involving a single new layer is displayed, as the two strategies incorporating two or three new layers produced similar learning performances and thus are omitted for clarity. It is observed that fully fine-tuning all layers results in the fastest initial convergence, while approaches that add new layers tend to have a slower initial learning phase.

Notably, the strategies that involve freezing certain layers, such as fine-tuning the last layer, the last two layers, or adding a new layer while freezing the others, show a significant performance drop compared to the baseline model. This suggests that restricting the trainable parameters in the later layers may limit the agent's ability to adapt to the complex dynamics of the high-fidelity environment, potentially due to insufficient capacity for learning new features or adapting to task-specific variations.

These results emphasize the importance of selecting an appropriate transfer strategy based on the specific requirements of the target task, as different strategies can lead to significantly different learning dynamics.

\begin{figure}[!htbp]
	\centering
	\includegraphics[trim={0 0.35cm 0 0.2cm}, clip, width=0.75\textwidth]{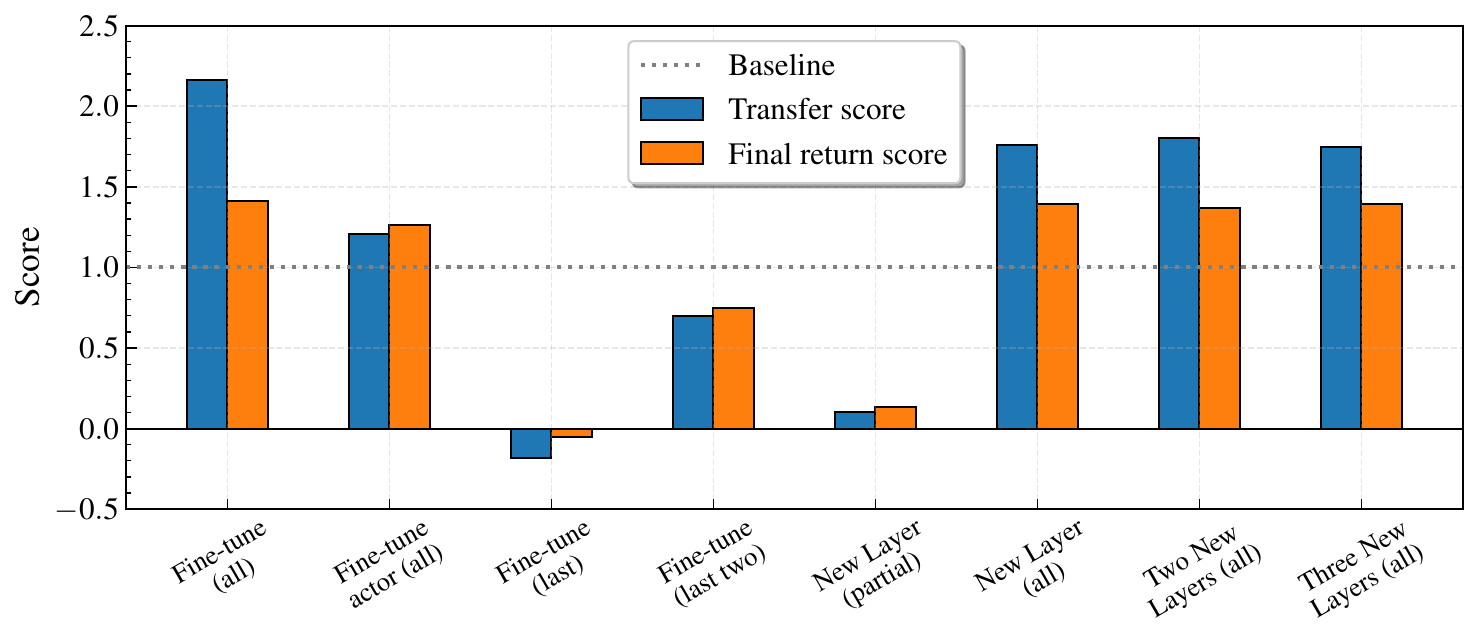}
	\caption{Transfer and final return scores for different transfer strategies from $N=16$ to $N=128$.}
	\label{fig:strategy_comparison_scores}
\end{figure}

To quantitatively compare the overall effectiveness of multifidelity strategies, Fig.~\ref{fig:strategy_comparison_scores} presents the corresponding transfer scores and final return scores. Fine-tuning all layers yields the highest transfer score of $2.15$, indicating the most effective knowledge transfer. In contrast, fine-tuning only the actor network is less effective, achieving a transfer score of $1.19$. The three strategies involving the addition of a new layer with all layers fine-tuned exhibit a consistent transfer score of approximately $1.75$, highlighting their potential for accelerating initial learning without fully unfreezing the network.

The final return scores across most strategies remain relatively similar, suggesting that despite differences in initial learning rates, the long-term performance is less affected by the choice of strategy.

\subsubsection{Transfer of knowledge across fidelities}
\label{sec:transfer_knowledge}

A key aspect of effective multifidelity RL is the nature of the information transferred between fidelities. In the context of fluid flow control, this information is closely tied to the physical structures present in the flow, which are strongly influenced by the spatial resolution of the environment. Given that low-fidelity environments, such as those with $N=16$ nodes, inherently lack the ability to capture fine-scale structures due to their coarse discretization (see Fig.~\ref{fig:sf_contours_16}), one can hypothesize that agents trained in such settings primarily learn to control and dampen large-scale flow structures. These large structures are typically the most energetic and therefore dominate the overall flow dynamics at coarse resolutions. As the fidelity increases, the transferred control strategy is expected to refine and extend to smaller structures that are only resolved in the high-fidelity environment.

To test this hypothesis, first, the ability of an agent, trained in a low-fidelity environment ($N=16$), to control a high-fidelity environment ($N=128$) is evaluated. The goal is to assess whether the agent can manage to dampen the large-scale structures in the high-resolution setting, even though it was not explicitly trained to control the fine-scale features present at this resolution. For this, the low-fidelity agent was deterministically evaluated in the high-fidelity environment, and the resulting flow fields were decomposed into their large-scale and small-scale components using a spectral filtering approach.

The decomposition is based on the Fourier transform, which naturally separates flow structures based on their characteristic wavelengths. Specifically, a cutoff wavenumber $k_c$ is defined as
\begin{equation}
	k_c = \frac{\pi N_{\text{low}}}{2L},
	\label{eq:cutoff}
\end{equation}
where $N_{\text{low}}$ is the number of nodes in the low-fidelity environment and $L$ is the domain length. This cutoff is chosen to reflect the maximum wavenumber that can be reasonably resolved by the low-fidelity environment. Using this approach, the flow field $u-u_\mathrm{ref}$ is decomposed into its large-scale ($u_\text{large}$) and small-scale ($u_\text{small}$) components.

\begin{figure}[!htbp]
	\centering
	\includegraphics[trim={0 0.35cm 0 0.2cm}, clip, width=\textwidth]{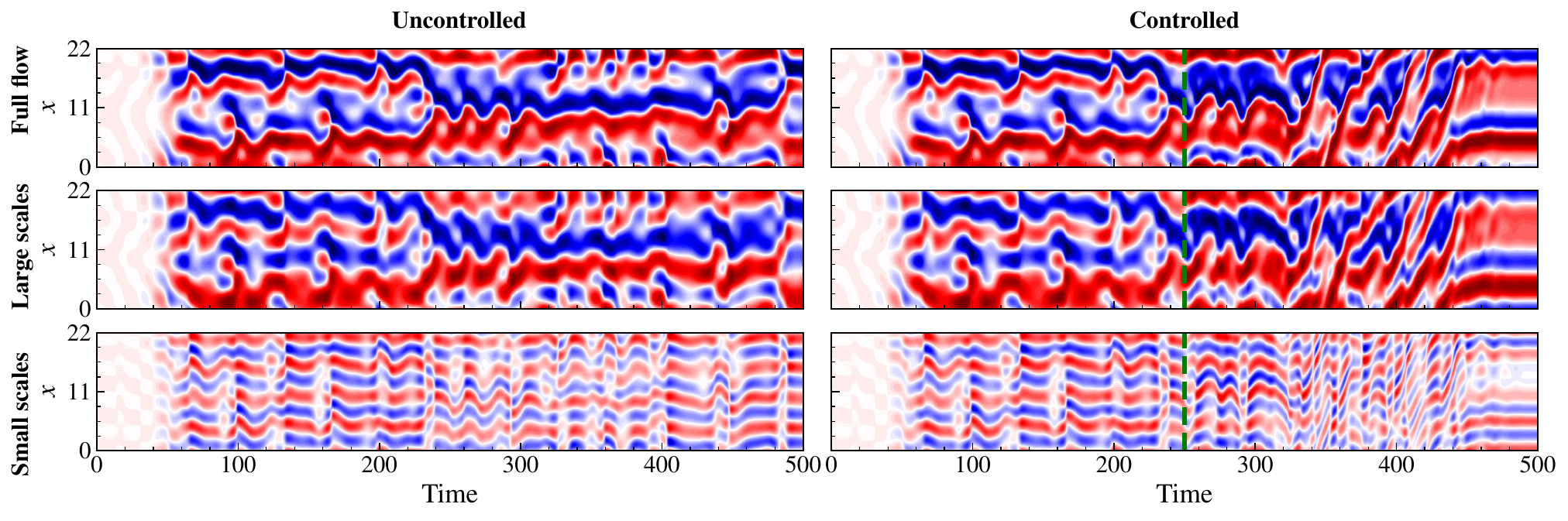}
	\caption{Decomposition of the flow into large-scale and small-scale components for both uncontrolled and controlled cases. The agent was trained on a low-fidelity environment ($N=16$) and tested on a high-fidelity environment ($N=128$).}
	\label{fig:high_128_low_16_decomposed_all}
\end{figure}

The spectral filtering separates the large-scale structures, which are represented by the low-frequency components, from the smaller-scale structures that exist beyond the resolution of the low-fidelity environment. 
Fig.~\ref{fig:high_128_low_16_decomposed_all} presents the resulting full flow, large-scale, and small-scale components of both uncontrolled and controlled cases. While it appears that the small-scale structures in the controlled case exhibit more intense fluctuations after around $t=300$, the effect of control on the large-scale structures is less obvious from this decomposition alone. This highlights the need for a more refined analysis to confirm whether the agent indeed focuses primarily on large-scale structures.

To further assess this hypothesis, a Proper Orthogonal Decomposition (POD) analysis of the flow field is performed. POD extracts the most energetic spatial structures, or modes, from a complex flow field by projecting the data onto a set of orthogonal basis functions. This is achieved through the singular value decomposition (SVD) of the snapshot matrix, which is constructed by collecting instantaneous flow snapshots over time
\begin{equation}
	\mathbf{X} = \mathbf{U} {\Sigma} \mathbf{V}^T,
	\label{eq:svd}
\end{equation}
where $\mathbf{X}$ is the snapshot matrix, $\mathbf{U}$ contains the temporal coefficients, ${\Sigma}$ is a diagonal matrix of singular values, and $\mathbf{V}^T$ contains the spatial modes. The squared singular values in ${\Sigma}$ are proportional to the energy captured by each mode, and the ratio of these squared singular values to their total sum provides a measure of the relative energy contribution of each mode. This decomposition is particularly well-suited for identifying the dominant structures in turbulent flows, where a small number of energetic modes can often capture a significant portion of the overall dynamics.

\begin{figure}[!htbp]
	\centering
	\includegraphics[trim={0 0.35cm 0 0.2cm}, clip, width=0.65\textwidth]{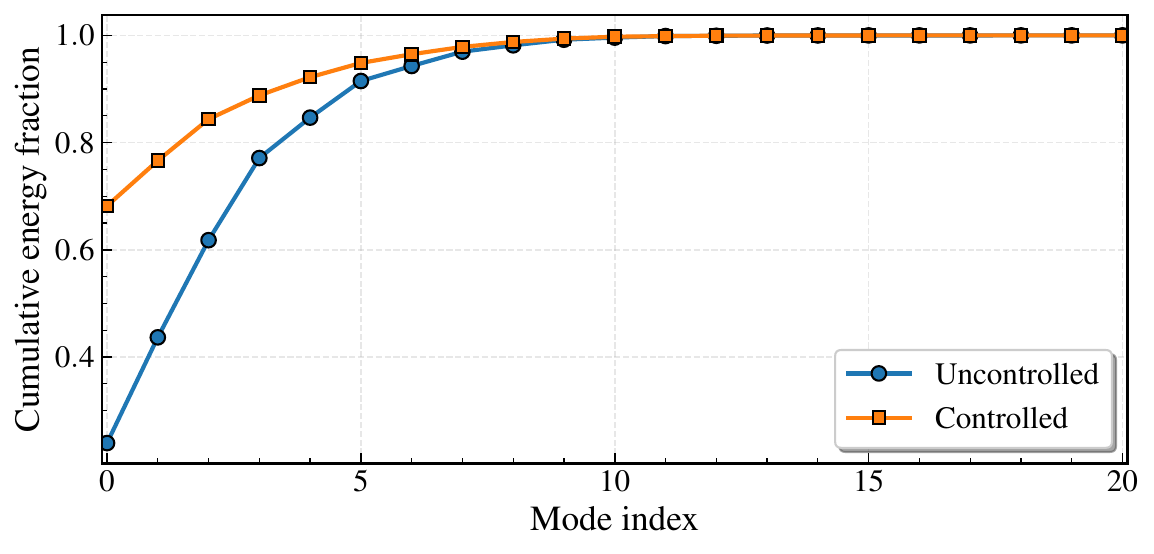}
	\caption{Cumulative energy fraction of the uncontrolled and controlled flow fields.}
	\label{fig:energy_fraction}
\end{figure}

To quantify the distribution of energy across the extracted modes, the cumulative energy fraction of the leading $k$ modes, defined as
\begin{equation}
	\text{Cumulative energy fraction} = \frac{\sum_{i=1}^k \sigma_i^2}{\sum_{j=1}^r \sigma_j^2}, 
\end{equation}
is examined, where $\sigma_i$ are the singular values from the SVD, and $r$ is the total number of modes. The cumulative energy fraction measures the proportion of the total flow energy captured by the leading modes. Fig.~\ref{fig:energy_fraction} shows this fraction for the uncontrolled and controlled cases. %
The controller appears to redistribute the energy across the spectrum, concentrating a larger portion of the total energy into the first few modes. This effectively reduces the complexity of the flow, allowing it to be represented with fewer spectral modes. This reduction in modal complexity could potentially indicate a more structured and less chaotic flow, aligning with the hypothesis that low-fidelity agents primarily target large-scale structures.

\begin{figure}[!htbp]
	\centering
	\includegraphics[trim={0 0.35cm 0 0.2cm}, clip, width=0.65\textwidth]{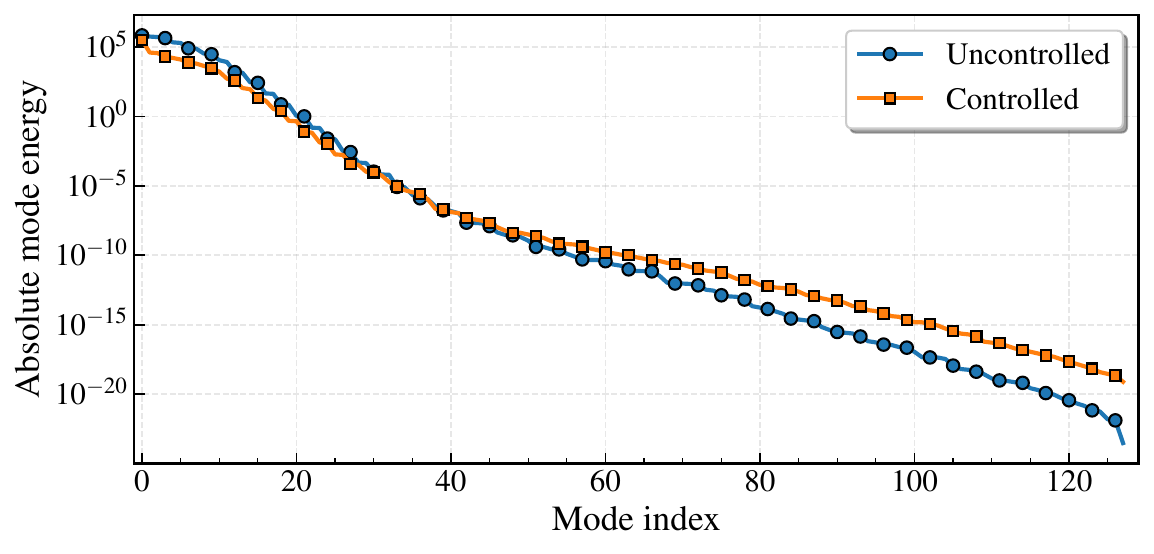}
	\caption{Absolute mode energies for the uncontrolled and controlled flow fields.}
	\label{fig:absolute_energy}
\end{figure}

Fig.~\ref{fig:absolute_energy} presents the absolute mode energies for the uncontrolled and controlled cases. It can be observed that the controlled case has significantly reduced the absolute energy in the leading modes (primarily in the first $30$ modes), consistent with our hypothesis that the agent trained in the low-fidelity environment primarily targets large-scale structures. However, the energy in the higher modes is increased, which is also in line with our observation in Fig.~\ref{fig:high_128_low_16_decomposed_all}.

\begin{figure}[!htbp]
	\centering
	\includegraphics[trim={0 0.35cm 0 0.2cm}, clip, width=1\textwidth]{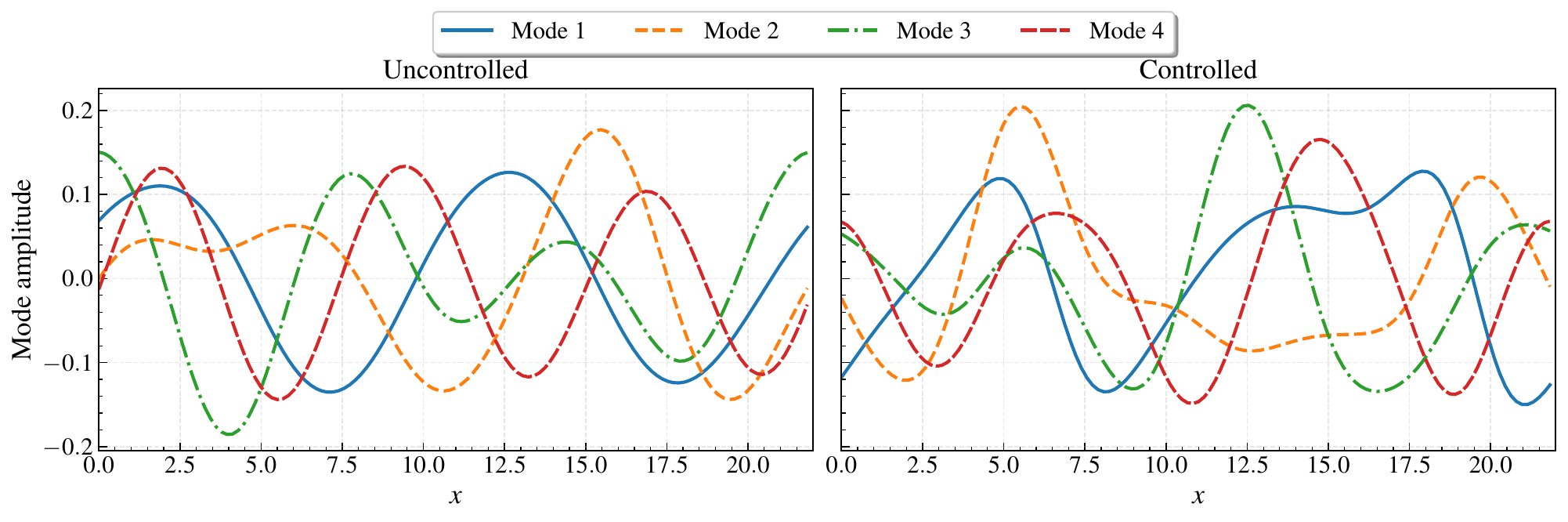}
	\caption{Spatial structure of the leading modes for the uncontrolled and controlled cases.}
	\label{fig:leading_modes}
\end{figure}

Fig.~\ref{fig:leading_modes} shows the first four spatial modes for both the uncontrolled and controlled cases. The uncontrolled modes appear to distribute energy more chaotically across the entire domain, while the controlled modes exhibit a more localized energy distribution, suggesting a possible stabilization effect.

\subsubsection{Knowledge retention and catastrophic forgetting}
\label{sec:transfer_catastrophic_forgetting}

While fine-tuning strategies often aim to accelerate learning on the high-fidelity target environment, they can inadvertently lead to catastrophic forgetting of information learned in the low-fidelity source environment. This is not necessarily an issue if the primary goal is rapid convergence on the target task. However, for two-way transfer learning, where knowledge is expected to flow in both directions, it is critical that the fine-tuned model retains its ability to control the original low-fidelity environment.

To quantify this, a \emph{knowledge retention score} is computed by evaluating the fine-tuned model back on its original, low-fidelity environment. The score is defined based on the ratio of the final return of the fine-tuned model to that of the original low-fidelity model. A score of 1 (or 100\%) indicates that the model has perfectly preserved the original information, while lower scores reflect varying degrees of catastrophic forgetting. 

\begin{figure}[!htbp]
	\centering
	\includegraphics[trim={0 0.25cm 0 0.2cm}, clip, width=0.65\textwidth]{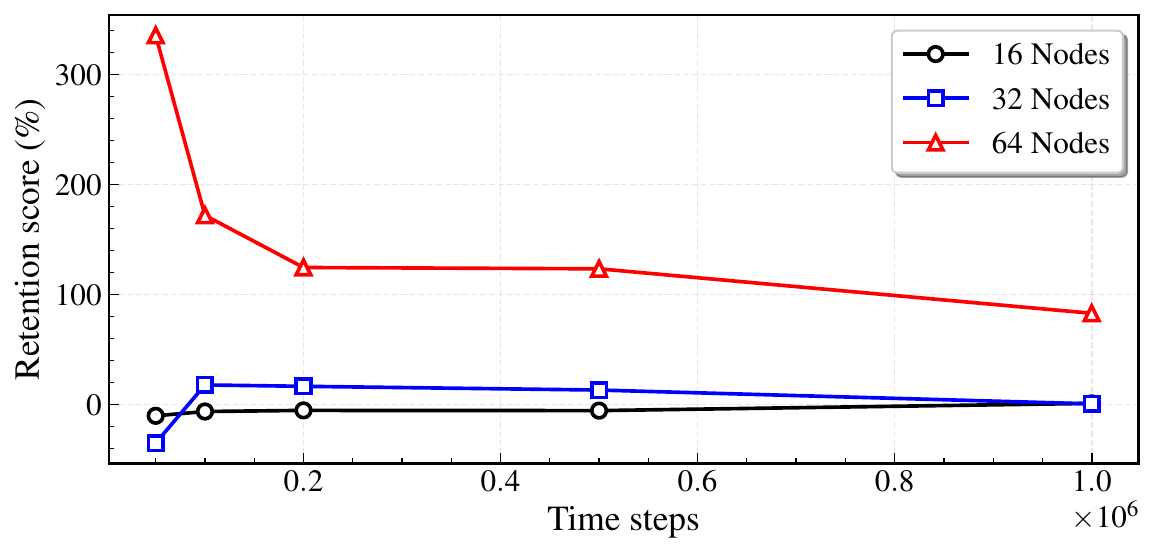}
	\caption{Knowledge retention scores for different low-fidelity training durations and source fidelities.}
	\label{fig:forgetting_steps}
\end{figure}

Fig.~\ref{fig:forgetting_steps} presents the retention scores for models fine-tuned on the high-fidelity environment of $N=128$ after initial training on low-fidelity setups with $N=16$, $32$, and $64$ nodes. It can be seen that the retention score is only high for the model pretrained on $N=64$, which has a stronger correlation and similarity to the target environment. This model retains and even improves its original knowledge, even after extensive fine-tuning, reflecting a more stable transfer. Increasing the low-fidelity training duration reduces the retention score, suggesting that more comprehensive pretraining can degrade knowledge retention.

In contrast, the retention scores for models pretrained on $N=16$ and $N=32$ are close to zero, indicating complete loss of the original low-fidelity knowledge, i.e., a clear case of catastrophic forgetting.

\begin{figure}[!htb]
	\centering
	\includegraphics[trim={0 0.35cm 0 0.2cm}, clip, width=0.65\textwidth]{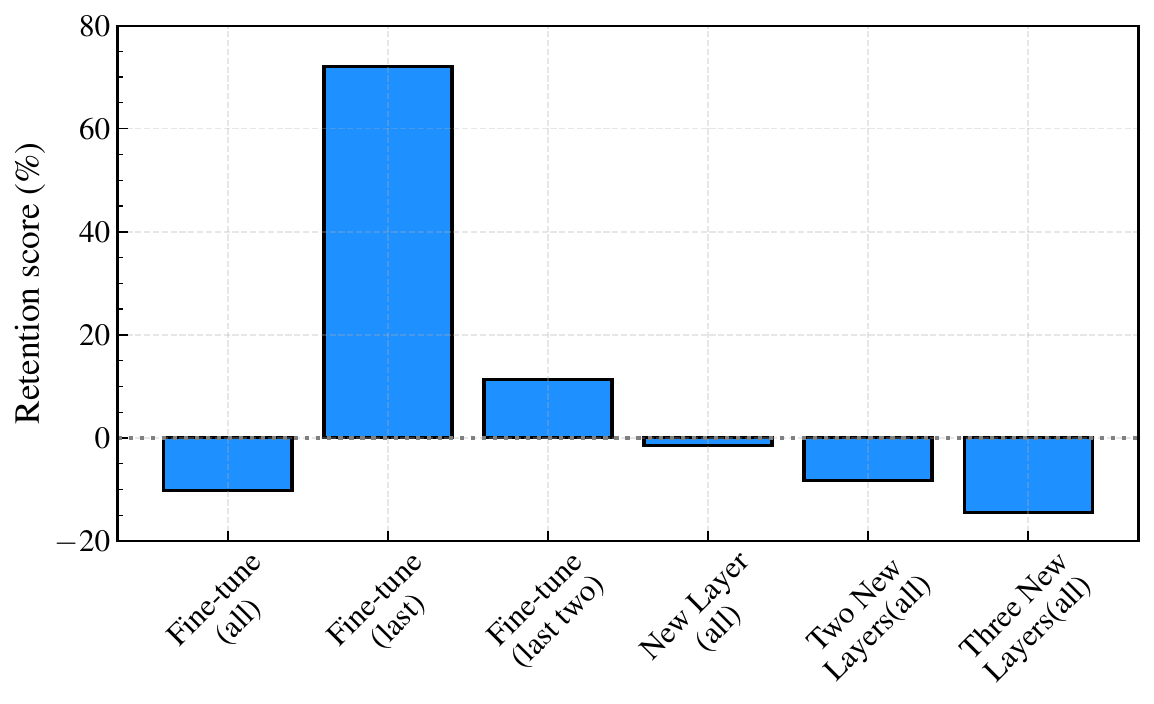}
	\caption{Comparison of forgetting scores across different fine-tuning strategies applied to models pretrained on $N=16$ for 50000 steps} 
	\label{fig:forgetting_strategies}
\end{figure}

Fig.~\ref{fig:forgetting_strategies} presents the retention scores for different fine-tuning strategies applied to models pretrained on $N=16$ for 50000 steps. It can be seen that the strategies involving frozen layers (e.g., fine-tuning only the last layer or the last two layers) have positive retention scores compared to the fully fine-tuned models. However, it is important to note that these strategies also struggled to converge and effectively learn in the high-fidelity environment, as previously discussed (see Fig.~\ref{fig:strategy_comparison_scores}).

\subsection{Progressive neural networks}
\label{sec:transfer_progressive}

This section evaluates the effectiveness of PNNs as an alternative to conventional fine-tuning strategies for multifidelity control of chaotic fluid flows.

\subsubsection{PNN strategies}
\label{sec:pnn_strategies}

The PNN strategies involve the first column being fully trained (with $10^7$ time steps) on the source low-fidelity environment of $N=16$, except for the random first column strategy, which uses a randomly initialized and frozen first column. The second column is trained on the target environment of $N=128$.

\begin{figure}[!htbp]
	\centering
	\includegraphics[trim={0 0.25cm 0 0.2cm}, clip, width=0.75\textwidth]{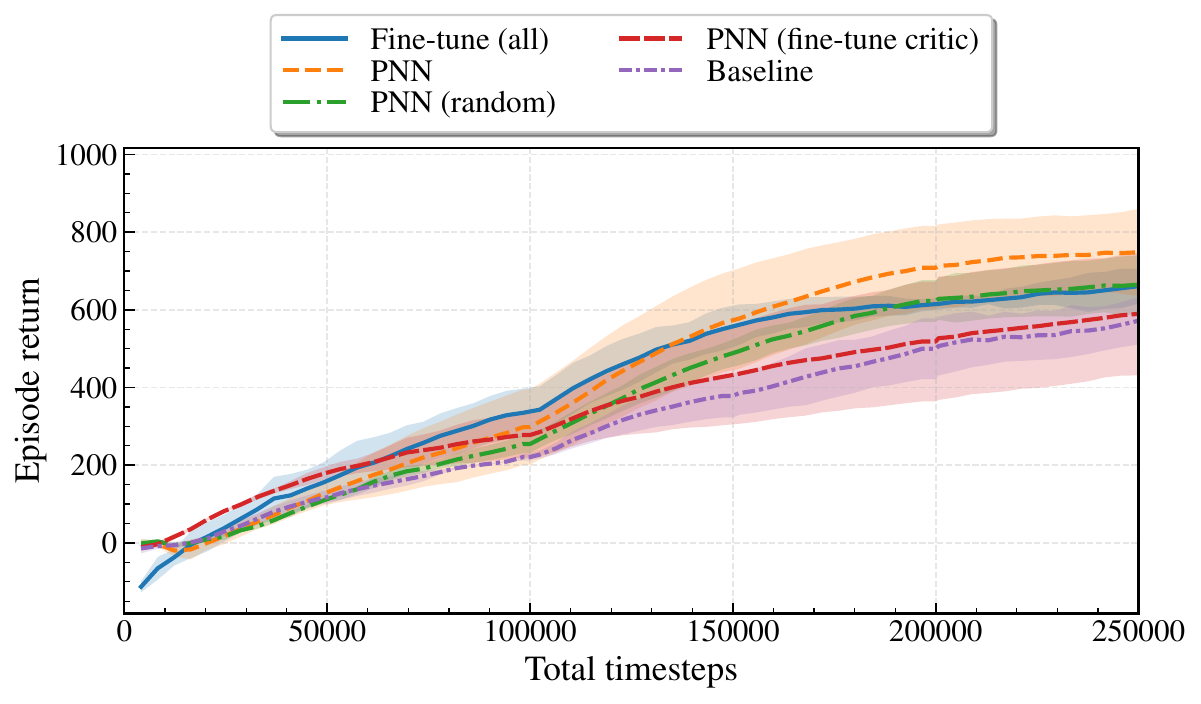}
	\caption{Learning curves for different PNN strategies with the source model being fully converged on the low-fidelity environment of $N=16$ after $10^7$ time steps, and the target model being trained on the high-fidelity environment of $N=128$.}
	\label{fig:convergence_pnn}
\end{figure}

The convergence behavior for the different strategies is presented in Fig.~\ref{fig:convergence_pnn} and compared to the fine-tuning and baseline approaches. It can be seen that the standard PNN and fine-tuning approaches have initially relatively similar convergence rates. However, the standard PNN approach outperforms the fine-tuning approach after about $1.5\times10^5$ time steps, indicating that the PNN approach is more effective in leveraging the knowledge from the source environment. 

The PNN with a random first column performs notably worse, confirming that a meaningful prior-column representation is crucial for effective transfer in the context of PNNs. The PNN with fine-tuned critics shows a slower initial learning that only marginally outperforms the baseline. This suggests that reusing critic features from the source environment may introduce constraints that limit the overall learning speed.

\begin{figure}[!htbp]
	\centering
	\includegraphics[trim={0 0.35cm 0 0.2cm}, clip, width=0.75\textwidth]{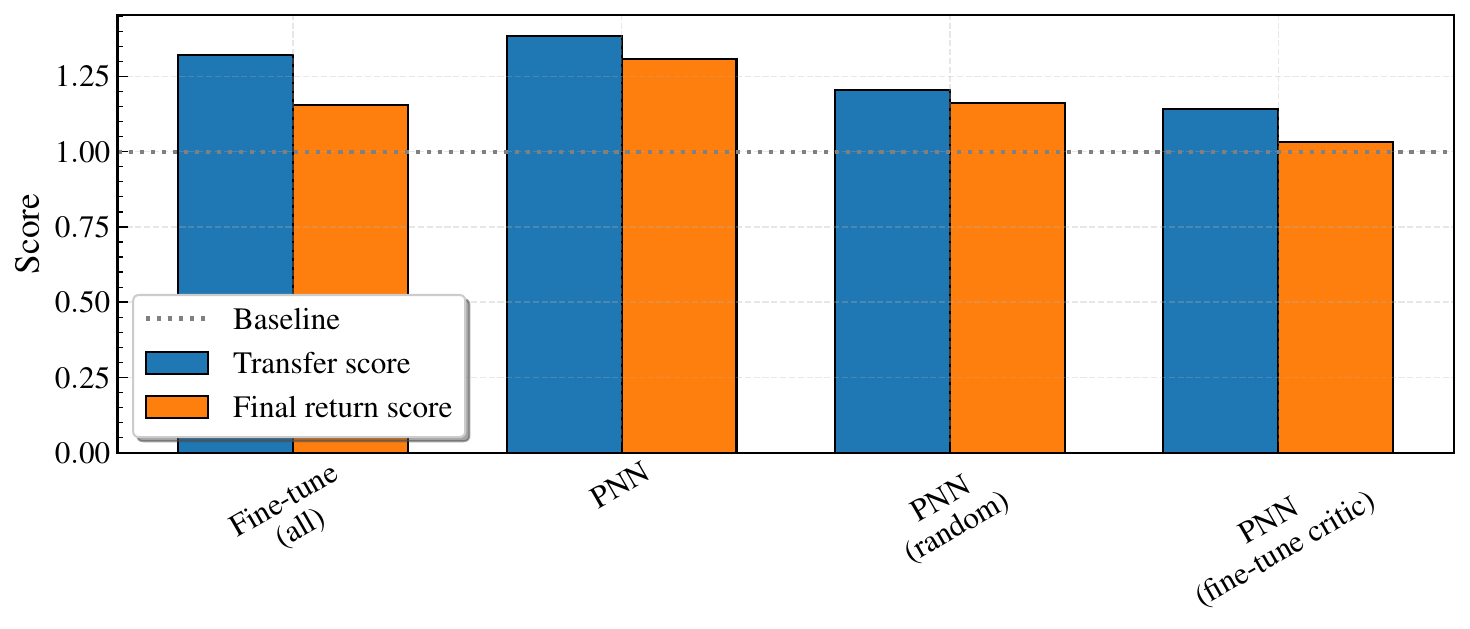}
	\caption{Transfer and final return scores for different PNN strategies, with the source model being trained on the low-fidelity environment ($N=16$) and the target model trained on the high-fidelity environment ($N=128$).}
	\label{fig:scores_pnn}
\end{figure}

The bar chart in Fig.~\ref{fig:scores_pnn} presents the transfer and final return scores for the different PNN strategies. It is clear that the standard PNN approach achieves the highest transfer scores, indicating effective reuse of the initial low-fidelity knowledge.

\begin{figure}[!tb]
	\centering
	\begin{subfigure}{0.495\textwidth}
		\centering
		\includegraphics[trim={0 0.25cm 0 0.2cm}, clip, width=\textwidth]{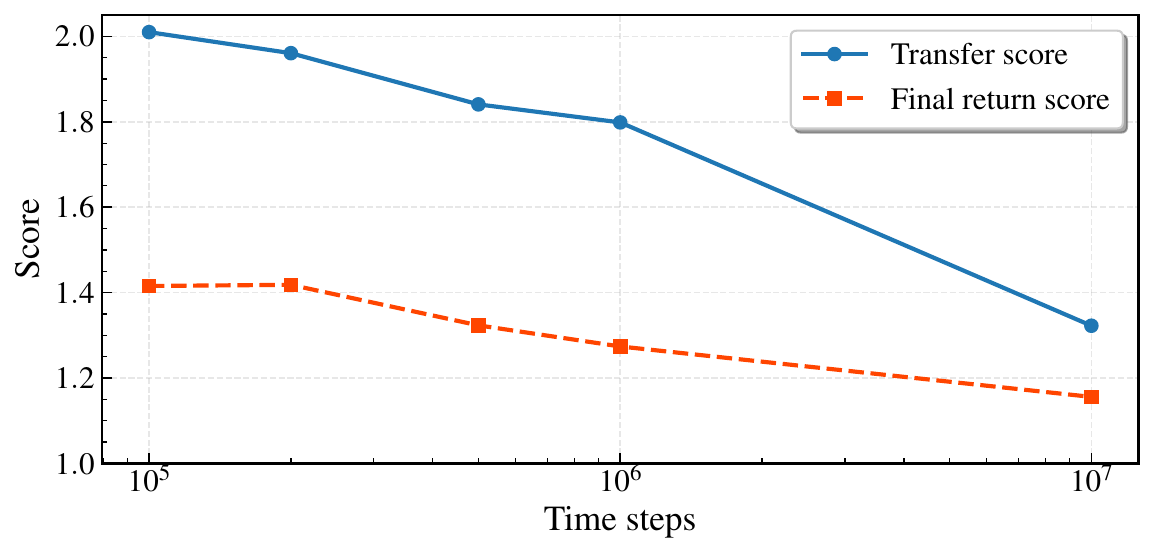}
		\caption{Fine-tunning}
		\label{fig:score_startstep_pnn_ft}
	\end{subfigure}
	\hfill
	\begin{subfigure}{0.495\textwidth}
		\centering
		\includegraphics[trim={0 0.25cm 0 0.2cm}, clip, width=\textwidth]{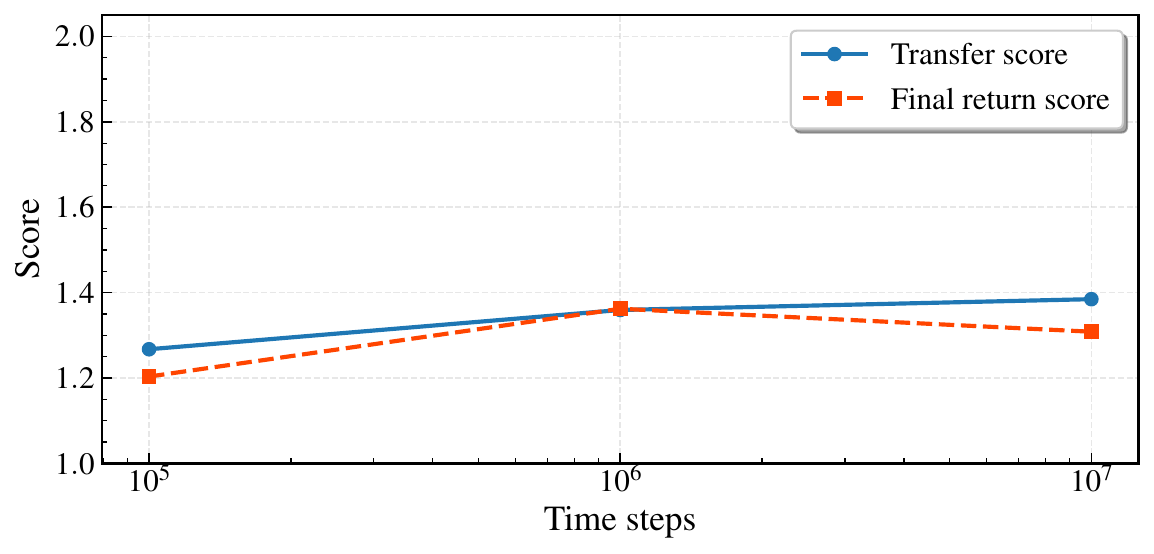}
		\caption{PNN}
		\label{fig:score_startstep_pnn_pnn}
	\end{subfigure}
	\caption{Effect of low-fidelity pretraining duration on the transfer learning scores from $N=16$ to $N=128$ . (\subref{fig:score_startstep_pnn_ft}) Fine tunning and (\subref{fig:score_startstep_pnn_pnn}) PNN.}
	\label{fig:score_startstep_pnn}
\end{figure}

The effect of the low-fidelity pretraining duration on the transfer learning scores of PNN is studied. Fig.~\ref{fig:score_startstep_pnn} compares the transfer and final return scores for fine-tuning and PNN strategies as a function of the low-fidelity pretraining duration from $N=16$ to $N=128$. 

It can be seen that the fine-tuning strategy (Fig.~\ref{fig:score_startstep_pnn_ft}) initially achieves higher transfer scores for short pretraining durations, indicating efficient transfer when the source model is not heavily overfitted. However, the performance sharply declines as the pretraining duration increases, reflecting a high sensitivity to the amount of source training. 

In contrast, the PNN approach (Fig.~\ref{fig:score_startstep_pnn_pnn}) exhibits a remarkable degree of robustness to the pretraining duration. The transfer and final return scores remain relatively stable, even for heavily overfitted source models trained on $10^7$ time steps. This indicates that PNNs can effectively reuse the initial low-fidelity knowledge without being as sensitive to the pretraining duration. This stability highlights a key advantage of PNNs, as they are able to leverage knowledge from long source training without catastrophic forgetting, making them particularly well-suited for continual learning scenarios.

\begin{figure}[p]
	\centering
	\begin{subfigure}{0.4\textwidth}
		\centering
		\includegraphics[trim={1.1cm 0.2cm 9.4cm 0.2cm}, clip, width=0.9\textwidth]{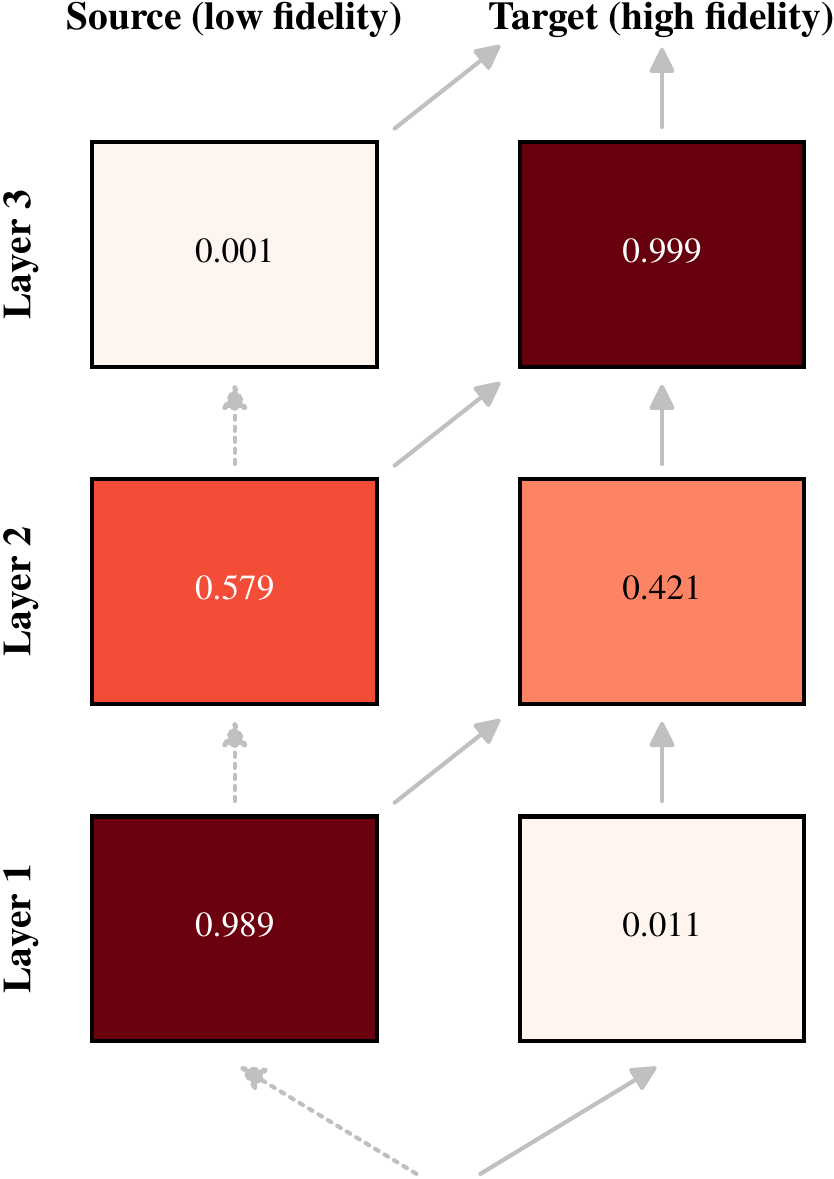}
		\caption{Time step $10^5$}
		\label{fig:pnn_aps_1e5}
	\end{subfigure}
	\hfill
	\begin{subfigure}{0.4\textwidth}
		\centering
		\includegraphics[trim={1.1cm 0.2cm 9.4cm 0.2cm}, clip, width=0.9\textwidth]{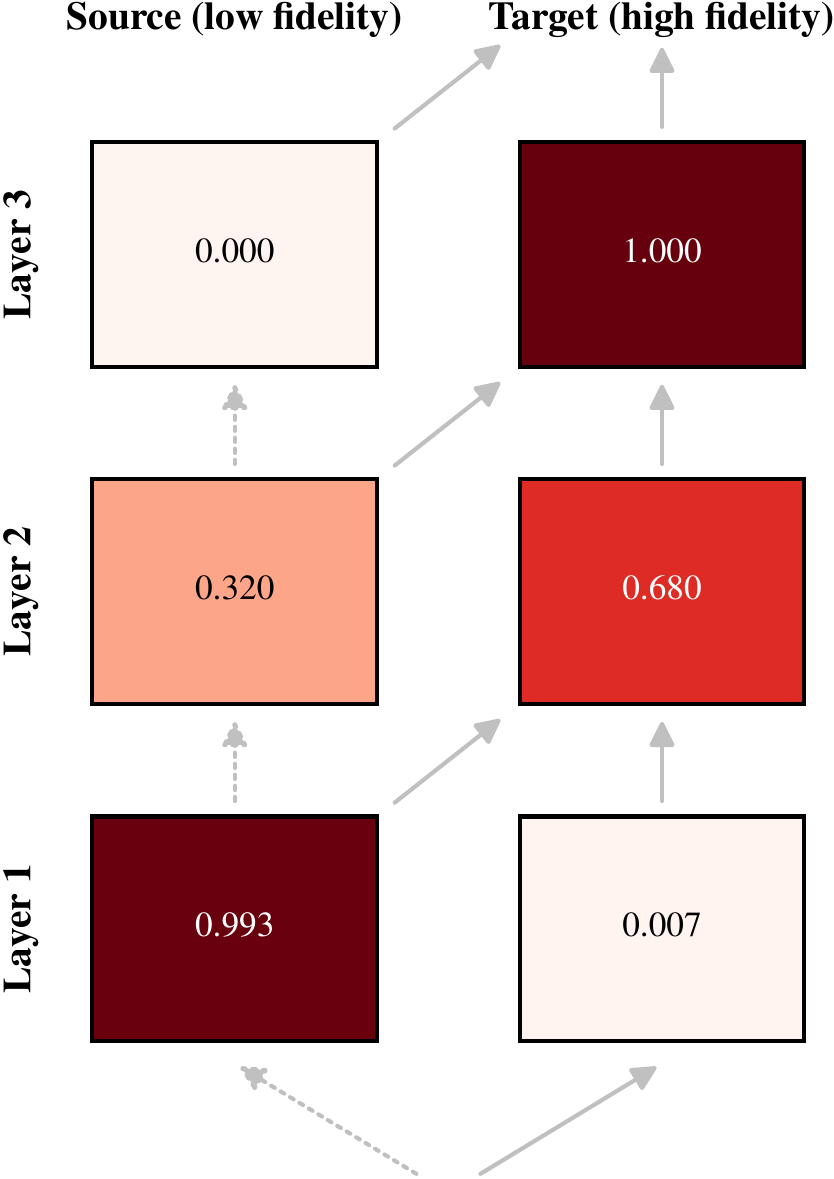}
		\caption{Time step $10^6$}
		\label{fig:pnn_aps_1e6}
	\end{subfigure}
	\par\bigskip
	\begin{subfigure}{0.4\textwidth}
		\centering
		\includegraphics[trim={1.1cm 0.2cm 9.4cm 0.2cm}, clip, width=0.9\textwidth]{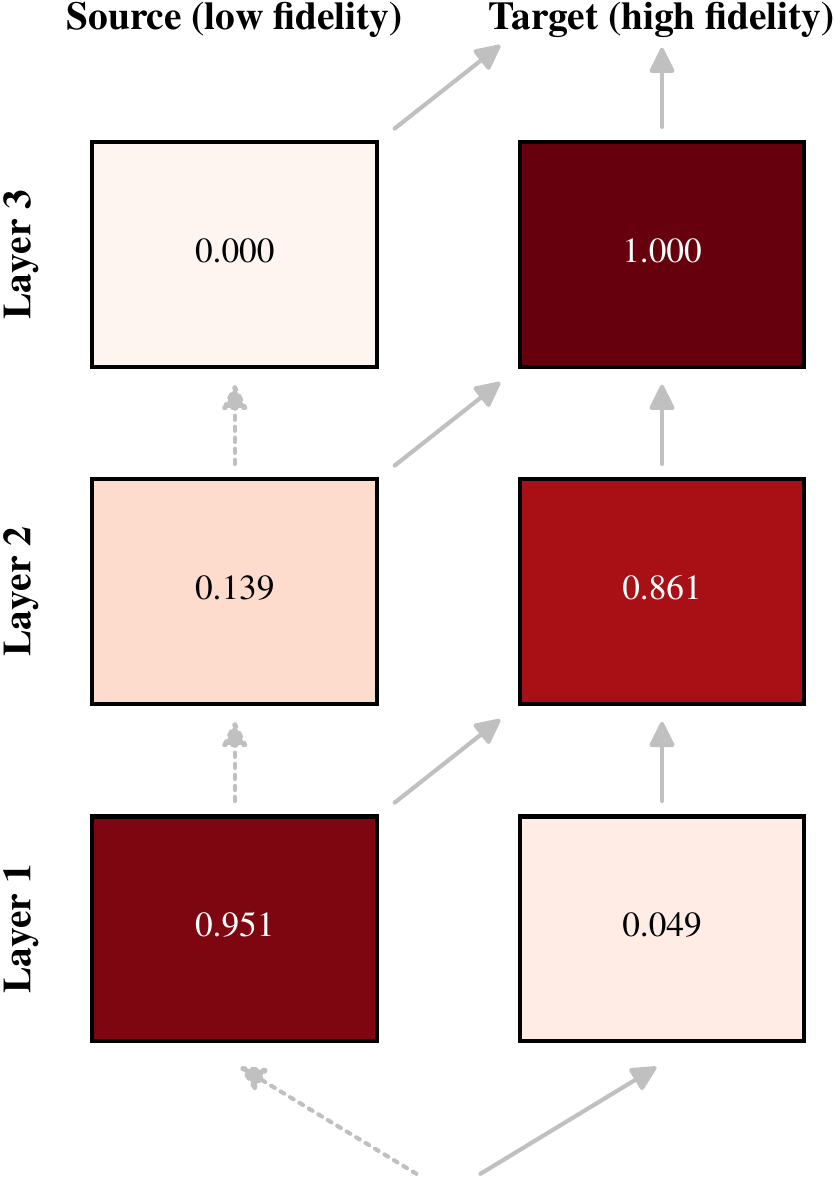}
		\caption{Time step $10^7$}
		\label{fig:pnn_aps_1e7}
	\end{subfigure}
	\hfill
	\begin{subfigure}{0.4\textwidth}
		\centering
		\includegraphics[trim={1.1cm 0.2cm 9.4cm 0.2cm}, clip, width=0.9\textwidth]{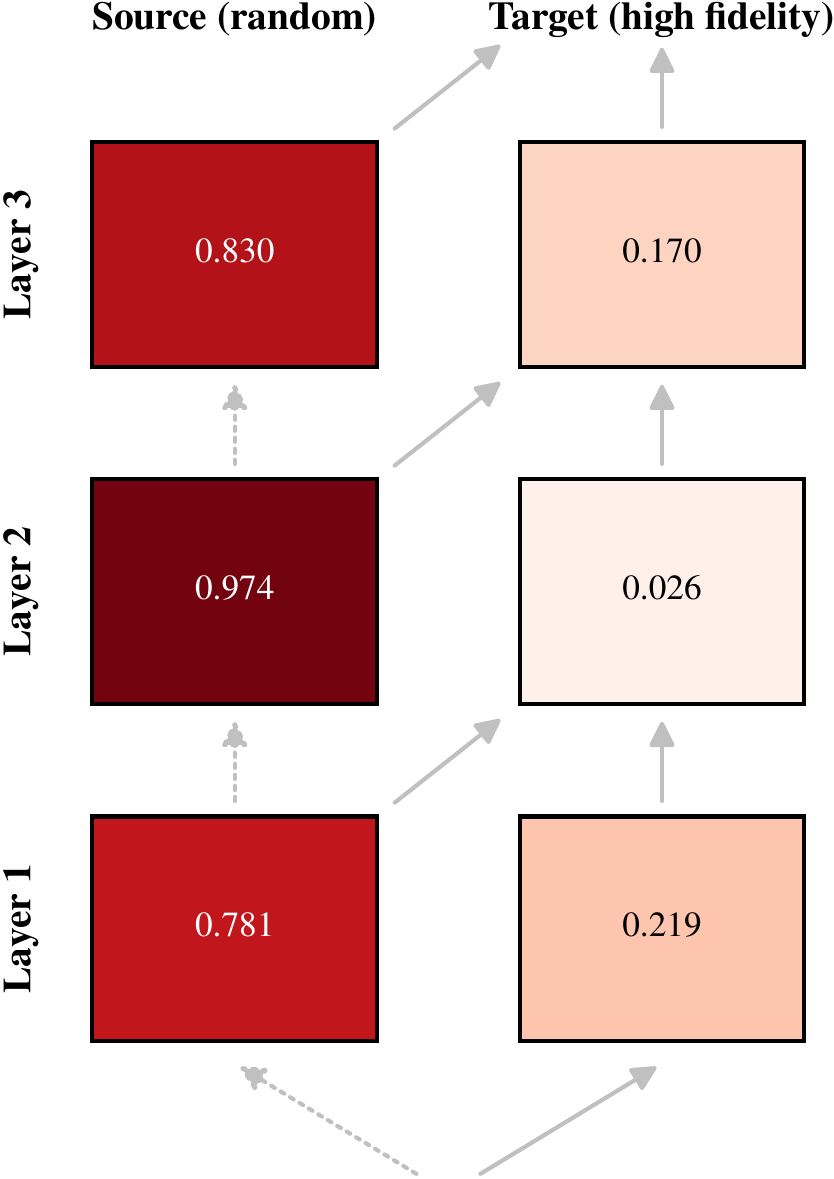}
		\caption{Random}
		\label{fig:pnn_aps_random}
	\end{subfigure}
	\par\bigskip
	\begin{subfigure}{0.6\textwidth}
		\centering
		\includegraphics[trim={1.8cm 1cm 1.8cm 0.6cm}, clip, width=0.9\textwidth]{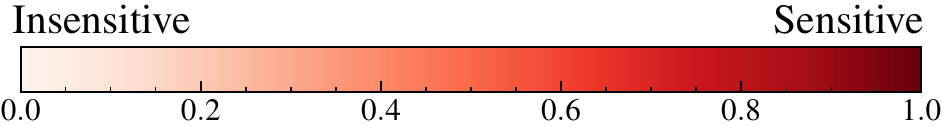}
	\end{subfigure}
	
	\caption{Average Perturbation Sensitivity (APS) maps for PNN with different low-fidelity pretraining durations. The first three subfigures correspond to models trained on the source environment for (\subref{fig:pnn_aps_1e5}) $10^5$, (\subref{fig:pnn_aps_1e6}) $10^6$, and (\subref{fig:pnn_aps_1e7}) $10^7$ time steps, respectively. The last subfigure (\subref{fig:pnn_aps_random}) shows the APS result for a model with a randomly initialized and frozen first column. The APS score reflects the relative importance of each layer of each column to the overall policy performance.}
	\label{fig:pnn_aps}
\end{figure}

To gain further insight into how the source column contributes to the target policy, the Average Perturbation Sensitivity (APS) analysis was conducted for different low-fidelity pretraining durations. The resulting APS maps are shown in Fig.~\ref{fig:pnn_aps}. Each subplot quantifies the relative importance of each layer in each column by measuring the degradation in performance when Gaussian noise is injected into individual activations (see Section~\ref{sec:transfer_aps} for more details.). 

The output of all PNN models (Figs.~\ref{fig:pnn_aps_1e5}--\ref{fig:pnn_aps_1e7}) heavily relies on the first layer of the source column, as indicated by their near-one APS scores and inactive first layer of the target column. This can be explained by the fact that the first layer is primarily responsible for extracting low-level features from the input, which remain relevant across both fidelity levels.

At earlier stages of training (Fig.~\ref{fig:pnn_aps_1e5}), the second layer of the source column also contributes significantly, even more than the second layer of the target column. This suggests that the PNN model builds upon intermediate representations learned in both low and high-fidelity environments. However, as training progresses to $10^6$ and $10^7$ time steps (Figs.~\ref{fig:pnn_aps_1e6} and~\ref{fig:pnn_aps_1e7}), the importance of the second layer in the source column steadily declines. This trend reflects increasing overfitting of the source model to low-fidelity dynamics, resulting in less transferable features. Concurrently, the second layer of the target column takes greater responsibility, indicating that the target model is progressively adapting to the specific demands of the high-fidelity environment.

The third layer of the source column remains inactive throughout all experiments, regardless of pretraining duration, suggesting that the high-level abstractions encoded by the source model are not beneficial for the target task. Instead, the final representation required for control in the high-fidelity environment is fully delegated to the last layer of the target column, which becomes solely responsible for producing the output. This observation is consistent with the expectation that high-level features in the source column are highly task-specific and prone to overfitting.

The APS result for the PNN with a randomly initialized and frozen first column (Fig.~\ref{fig:pnn_aps_random}) reveals a counter-intuitive outcome, in which the model exhibits strong sensitivity to the random column. A plausible explanation can be found in the structure of the PNN. While the hidden layers of the first column are frozen, the lateral connections feeding into the second column are fully learnable. The model effectively learns to work around the randomness by shaping its lateral pathways to accommodate and utilize the static, unstructured features. This behavior illustrates a key limitation. In the absence of a meaningful and structured source representation, the target column becomes largely inactive. It fails to develop useful features independently, as it is influenced by irrelevant signals from the frozen random column.

\begin{figure}[!tb]
	\centering
	\begin{subfigure}{0.495\textwidth}
		\centering
		\includegraphics[trim={0 0.35cm 0 0.2cm}, clip, width=\textwidth]{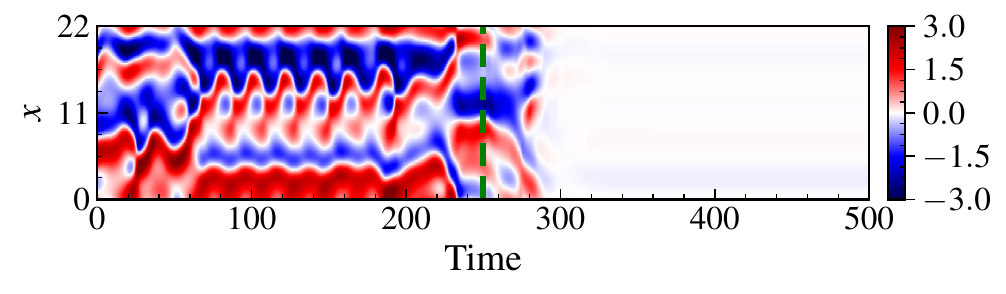}
		\caption{Full model}
		\label{fig:ablation_full}
	\end{subfigure}
	\hfill
	\begin{subfigure}{0.495\textwidth}
		\centering
		\includegraphics[trim={0 0.35cm 0 0.2cm}, clip, width=\textwidth]{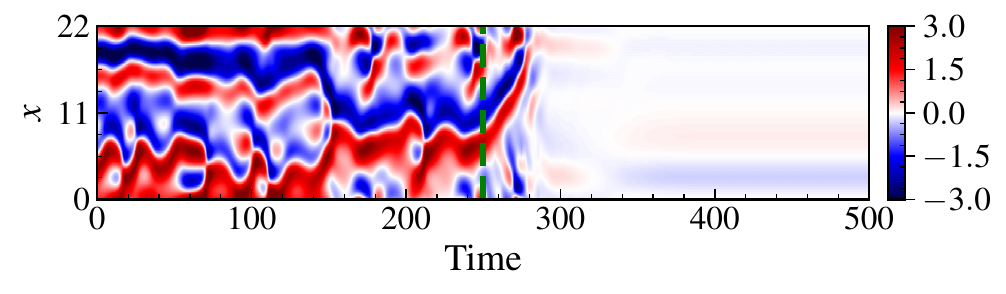}
		\caption{Layer 1 off}
		\label{fig:ablation_layer1}
	\end{subfigure}
	
	\bigskip
	
	\begin{subfigure}{0.495\textwidth}
		\centering
		\includegraphics[trim={0 0.35cm 0 0.2cm}, clip, width=\textwidth]{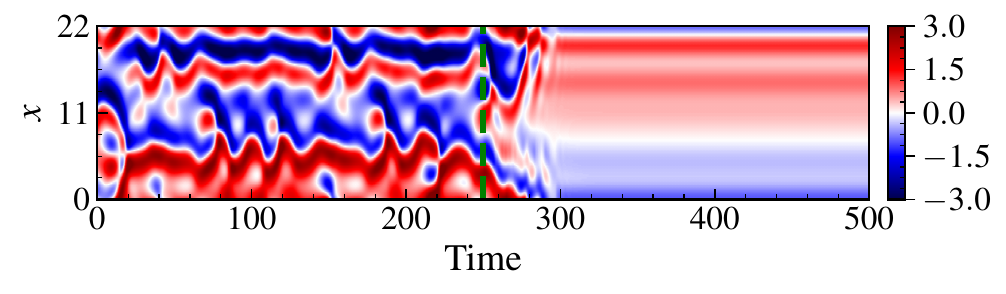}
		\caption{Layer 2 off}
		\label{fig:ablation_layer2}
	\end{subfigure}
	\hfill
	\begin{subfigure}{0.495\textwidth}
		\centering
		\includegraphics[trim={0 0.35cm 0 0.2cm}, clip, width=\textwidth]{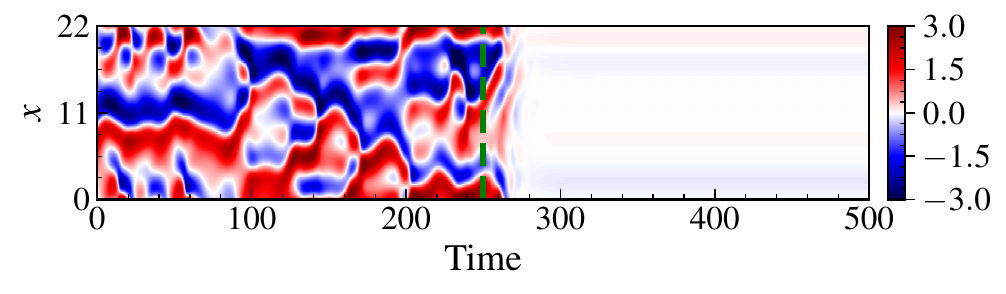}
		\caption{Layer 3 off}
		\label{fig:ablation_layer3}
	\end{subfigure}
	
	\caption{Ablation study of adapter gains in individual layers of the source column. The full model (\subref{fig:ablation_full}) is compared against variants where the adapter gain of each source layer is individually set to zero (\subref{fig:ablation_layer1} -- \subref{fig:ablation_layer3}).}
	\label{fig:ablation_layers}
\end{figure}

It is important to note that the APS score reflects the relative importance of each column at a given layer, rather than the absolute importance of layers themselves. 

To directly assess the functional role of each layer in the source column, an ablation study was performed in which the adapter gain of each layer was individually set to zero, making the corresponding layer of the source column inactive, while keeping the rest of the model fixed. The results in Fig.~\ref{fig:ablation_layers} show that disabling Layer 1 of the source column slightly affects the flow control, whereas removing Layer 2 leads to a substantial degradation in performance. Layer 3 has no observable impact. This indicates that the second layer of the source column encodes essential features that are not captured by the earlier or later layers, and is therefore critical for successful knowledge transfer.

\begin{figure}[!tb]
	\centering
	\includegraphics[trim={0 0.35cm 0 0.2cm}, clip, width=0.5\textwidth]{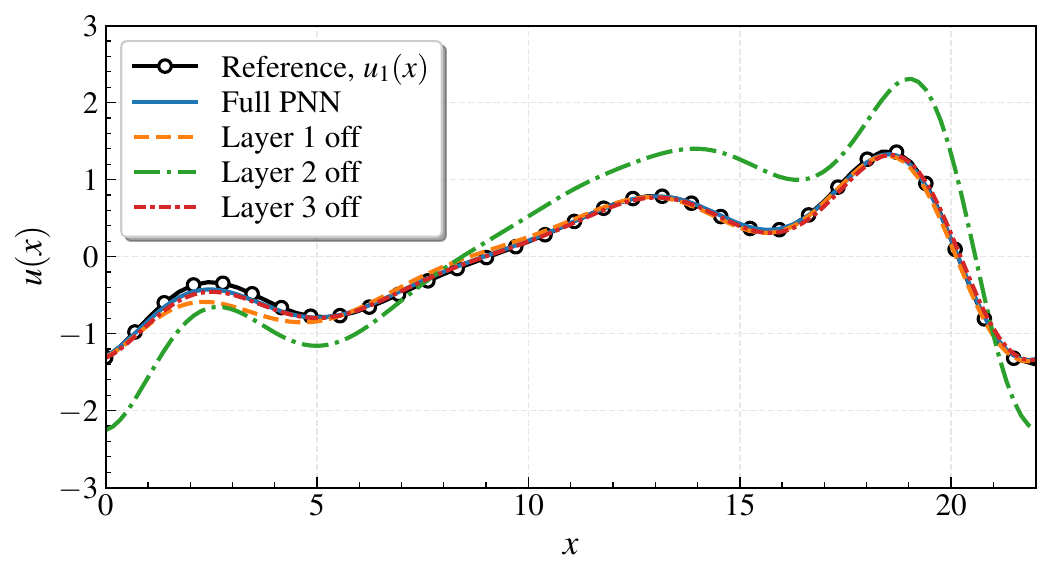}
	\caption{Final state \( u(x) \) for the full PNN model and ablation variants compared to the reference solution \( u_1(x) \). Turning off Layer 2 leads to a noticeable deviation, while Layers 1 and 3 have a negligible impact.}
	\label{fig:ablation_final_u}
\end{figure}

The final state of the controlled solution is also shown in Fig.~\ref{fig:ablation_final_u}. The largest deviation from the reference is observed when layer 2 of the source column is disabled, confirming its essential role in enabling the model to drive the flow toward the desired target. The solutions with layer 1 or 3 removed remain nearly identical to the full model.

\subsubsection{Knowledge transfer across different physical regimes}
\label{sec:transfer_physics}

In addition to variations in resolution, an important aspect of transfer learning in PDE-based systems is the ability to generalize across different physical regimes. In the KS equation, the hyperviscosity parameter \(\lambda\) directly influences the scale and intensity of instabilities in the flow. Even small changes in \(\lambda\) can significantly alter the underlying dynamics and control requirements.

\begin{figure}[!tb]
	\centering
	\includegraphics[trim={0 0.35cm 0 0.2cm}, clip, width=0.8\textwidth]{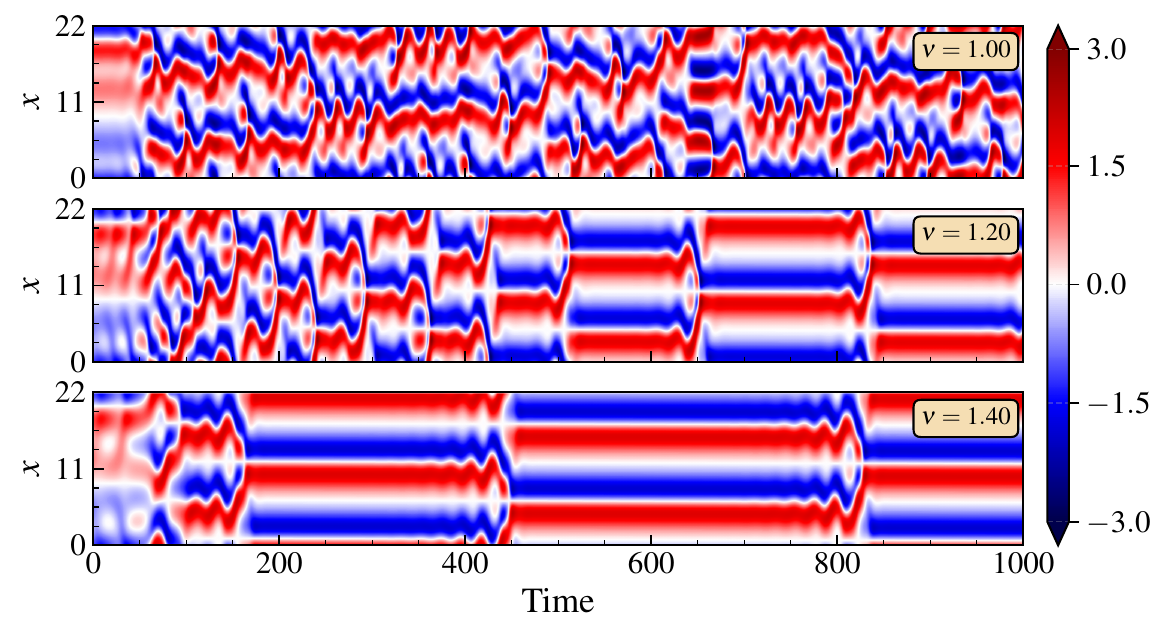}
	\caption{Spatiotemporal evolution of the solution of the KS equation for different hyperviscosity values, namely, \(\lambda = 1.00\), \(1.20\), and \(1.40\). Increasing \(\lambda\) suppresses chaotic behavior and promotes the formation of more stable structures.}
	\label{fig:ks_lambda}
\end{figure}

Fig.~\ref{fig:ks_lambda} illustrates the spatiotemporal evolution of the uncontrolled KS system for different values of the hyperviscosity parameter \(\lambda\). At \(\lambda = 1.00\), the flow is fully chaotic, exhibiting persistent, irregular fluctuations across space and time. Increasing \(\lambda\) enhances the system's diffusive behavior and reduces the intensity of chaos, leading to intermittent periods of smooth and laminar structures. At \(\lambda = 1.40\), the flow is dominated by long-lasting, quasi-stable patterns that undergo abrupt transitions to entirely new quasi-stable states.

It is observed that the hyperviscosity parameter has a significant impact on the underlying flow physics, fundamentally altering the system's dynamical regime from strongly chaotic to intermittently laminar as \(\lambda\) increases.

\begin{figure}[!tb]
	\centering
	\begin{subfigure}{0.495\textwidth}
		\centering
		\includegraphics[trim={0 0.25cm 0 0.2cm}, clip, width=\textwidth]{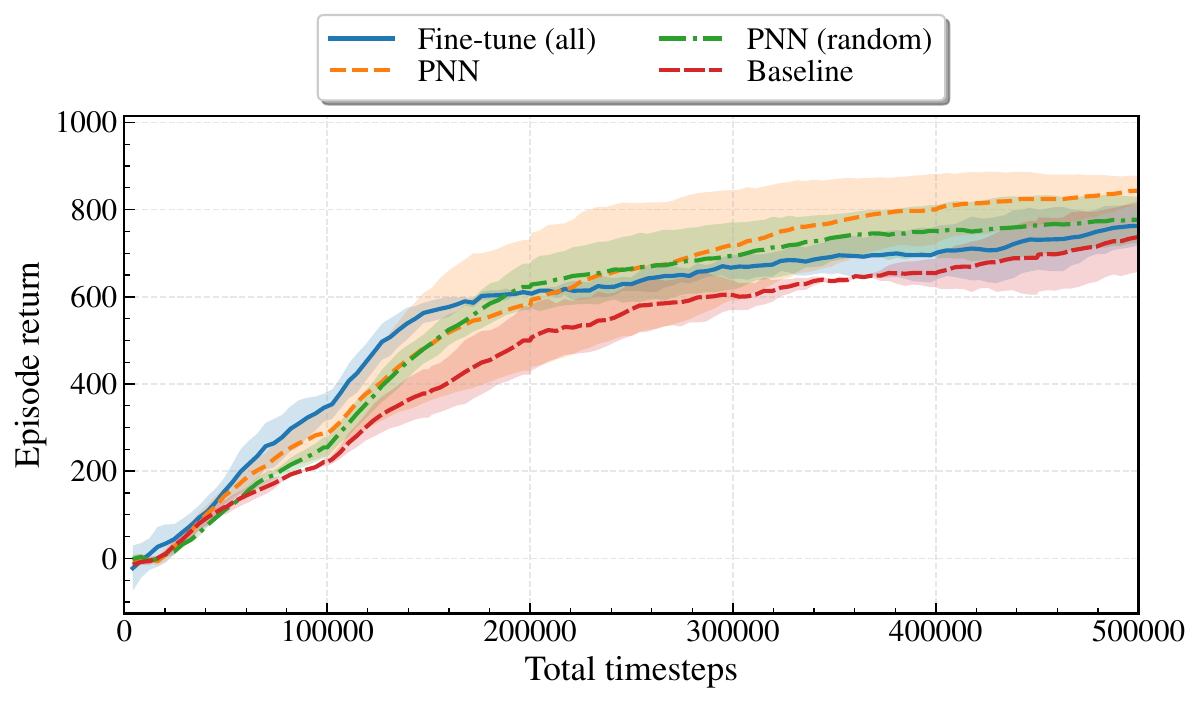}
		\caption{$\lambda=1.2$}
		\label{fig:pnn_lambda12}
	\end{subfigure}
	\hfill
	\begin{subfigure}{0.495\textwidth}
		\centering
		\includegraphics[trim={0 0.25cm 0 0.2cm}, clip, width=\textwidth]{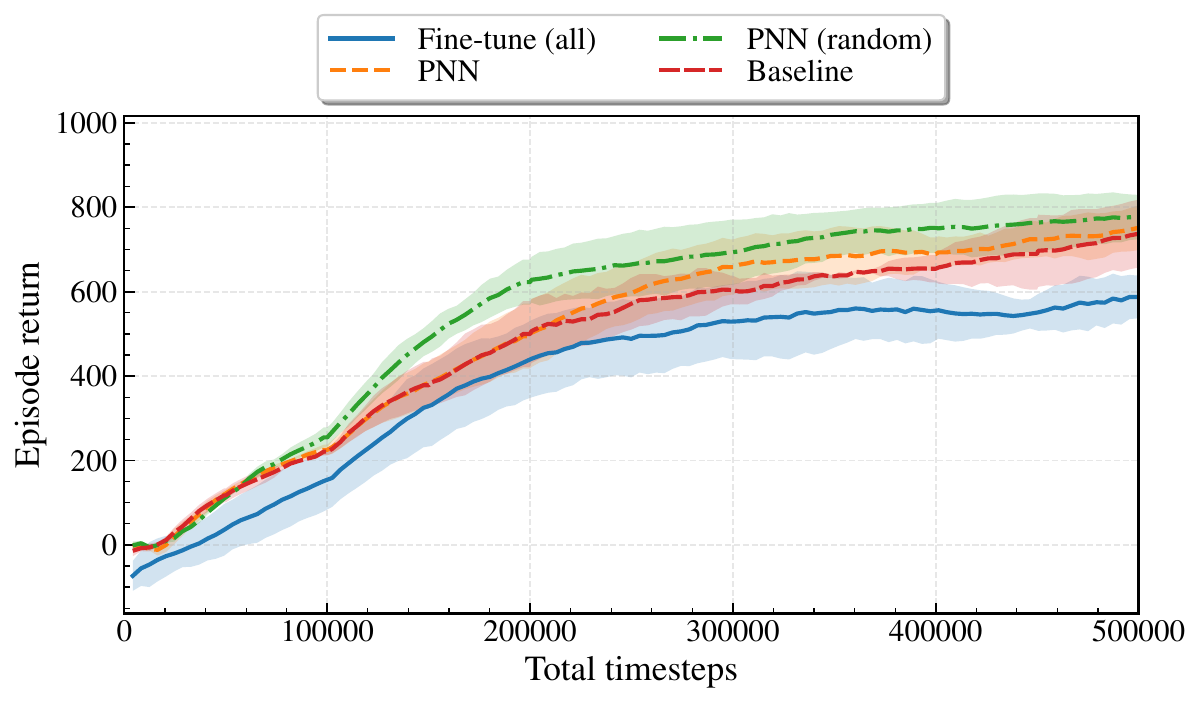}
		\caption{$\lambda=1.4$}
		\label{fig:pnn_lambda14}
	\end{subfigure}   
	\caption{Learning curves for transfer from source models trained on a coarse grid (\(N = 16\)) with hyperviscosity values of (\subref{fig:pnn_lambda12}) \(\lambda = 1.2\) and (\subref{fig:pnn_lambda14}) \(\lambda=1.4\) to a high-fidelity target environment (\(N = 128\), \(\lambda = 1.0\)).}
	\label{fig:pnn_lambda}
\end{figure}

To investigate whether policies trained under one set of physical conditions can be reused in another, a series of experiments was conducted where the source model was trained on a coarse grid (\(N = 16\)) with \(\lambda = 1.2\) and \(\lambda = 1.4\), then transferred and further trained on the standard high-fidelity configuration (\(N = 128\), \(\lambda = 1.0\)).

Fig.~\ref{fig:pnn_lambda} shows the learning curves for these transfers. In both cases, the standard PNN strategy consistently outperforms both the baseline and fine-tuning approaches. However, for \(\lambda = 1.4\), where the source physics deviates substantially from the target, the fine-tuning model fails to improve upon the baseline. This indicates that the transferred knowledge is no longer beneficial and may even hinder adaptation to the target dynamics.

In contrast, the PNN model maintains a performance gain over the baseline, demonstrating its robustness to discrepancies in physical regimes. Notably, the PNN with a randomly initialized and frozen source column outperforms the standard PNN in the \(\lambda = 1.4\) case. This behavior may be attributed to the flexibility of its lateral connections, which are not constrained by mismatched source features and can more freely adapt to the target task.

\subsubsection{Knowledge transfer under inconsistent objectives}
\label{sec:transfer_inconsistent_objectives}

In the previous experiments, PNNs demonstrated marginal improvements over fine-tuning. A plausible explanation is the strong similarity in the dynamics and objectives of the low and high-fidelity environments, which made the transfer relatively straightforward for both approaches. To more critically assess the benefits of PNNs, a more challenging setup was devised in which the source and target environments differ not only in resolution but also in control objective.

Specifically, the source model was trained on a coarse grid ($N=16$) to minimize deviations from the steady state solution $u_0$, while the target task, defined on a high-resolution grid ($N=128$), was designed to drive the flow toward a different reference state $u_1$. This deliberate mismatch in objectives is intended to simulate a case where the source and target environments are not only different in resolution but also in their control objectives, providing a more challenging transfer scenario.

\begin{figure}[!htbp]
	\centering
	\includegraphics[trim={0 0.25cm 0 0.2cm}, clip, width=0.75\textwidth]{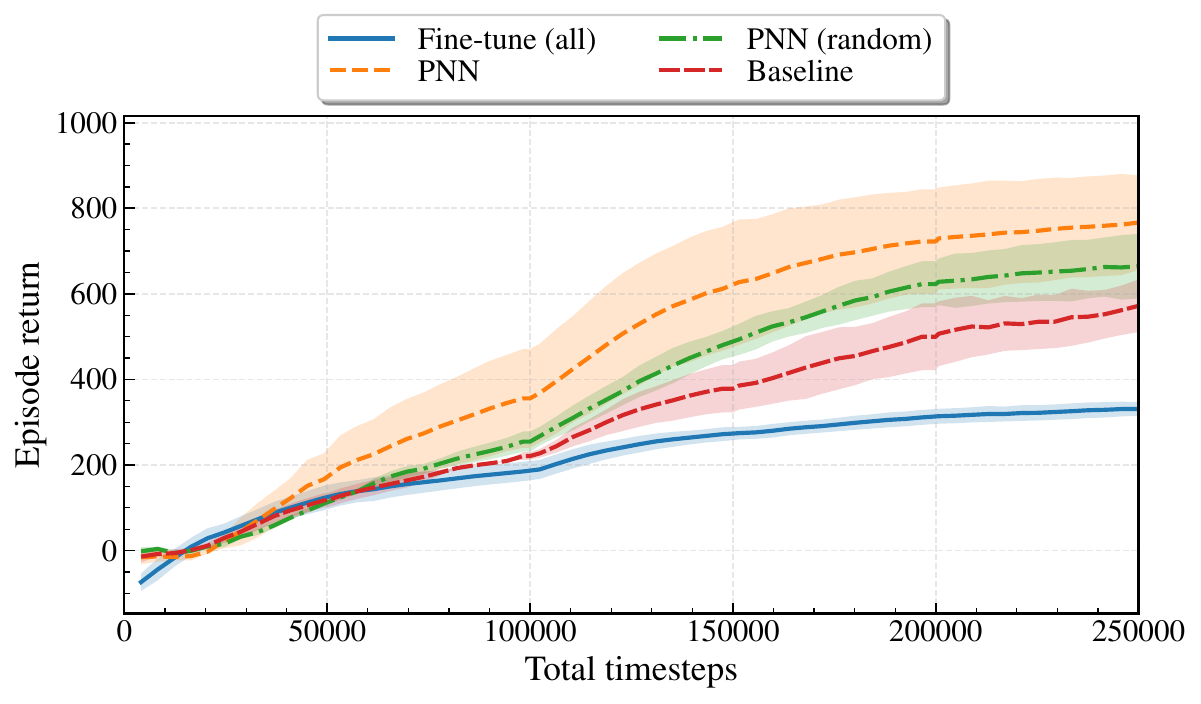}
	\caption{Learning curves under inconsistent objectives for PNN and fine-tuning strategies. The source model is trained on $N=16$ with a reward based on $u_0$ and transferred to the target model $N=128$ with reward based on $u_1$.}
	\label{fig:convergence_pnn_ref0}
\end{figure}

The convergence curves for this experiment are shown in Fig.~\ref{fig:convergence_pnn_ref0}. The results reveal a clear failure of the fine-tuning strategy, as the fine-tuned agent's learning is significantly slower than the baseline. This suggests that the knowledge embedded in the source model becomes misleading when directly adapted to the new task, causing ineffective learning or even negative transfer.

In contrast, the PNN approach demonstrates robust learning behavior. It steadily improves and reaches significantly higher return values, clearly outperforming both the fine-tuned and baseline models. This confirms that PNNs are more resilient to inconsistencies in reward structure, as their architecture allows new features to be learned in isolation while still leveraging transferable components from the source model. 

The PNN with a random first column is also shown as a reference, which is again outperformed by the standard PNN, reinforcing the importance of meaningful source knowledge in enabling successful transfer.

\begin{figure}[!htbp]
	\centering
	\includegraphics[trim={0 0.35cm 0 0.2cm}, clip, width=0.75\textwidth]{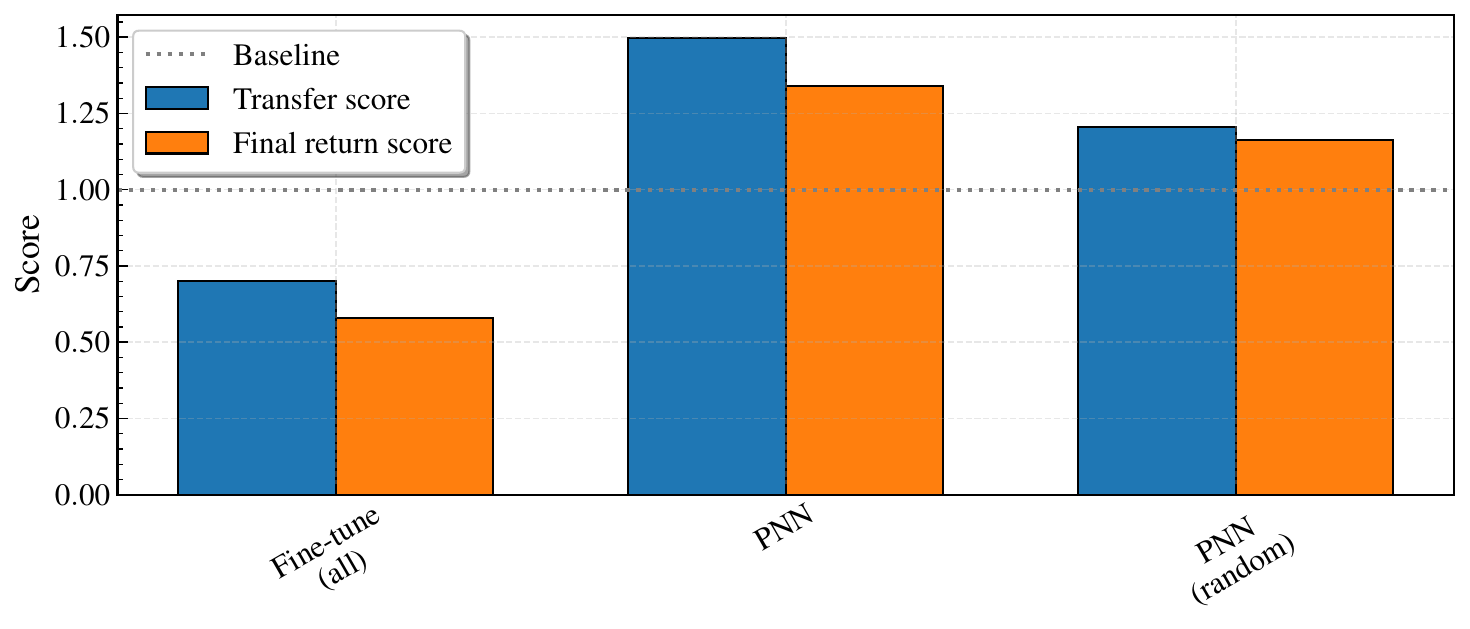}
	\caption{Transfer and final return scores under inconsistent objectives for different PNN strategies. The source model is trained on $N=16$ with a reward based on $u_0$ and transferred to the target model $N=128$ with a reward based on $u_1$.}
	\label{fig:convergence_pnn_ref0_scores}
\end{figure}

Fig.~\ref{fig:convergence_pnn_ref0_scores} presents the transfer and final return scores of the employed strategies. The fine-tuning strategy yields scores of lower than one, while the PNN strategy can achieve a 50\% improvement of the transfer score.

\section{Conclusion and future perspectives}
\label{sec:conclusion}

This work systematically investigated and benchmarked various transfer learning strategies to accelerate Deep Reinforcement Learning (DRL) for active flow control. The comprehensive analysis, spanning multifidelity environments with different levels of discretizations, varying physical regimes, and inconsistent control objectives, highlights the critical role of structured transfer learning frameworks in overcoming the computational bottlenecks and knowledge retention challenges inherent in DRL-based flow control.

The numerical experiments utilized the chaotic Kuramoto-Sivashinsky (KS) equation as a challenging yet computationally tractable test case. It was demonstrated that conventional fine-tuning strategies can accelerate convergence in the target environment. A controller trained on low-fidelity environments primarily provides control over large-scale flow structures, which is then refined to include smaller-scale details in higher-fidelity settings. However, the effectiveness of fine-tuning strategies is highly sensitive to the duration of pretraining (i.e., overfitting) and critically susceptible to catastrophic forgetting. Additionally, fine-tuning becomes particularly ineffective when there are significant discrepancies between the source and target domains, such as differing control objectives. In such challenging scenarios, fine-tuning leads to negative transfer, indicating that the transferred knowledge becomes misleading or irrelevant.

Progressive Neural Networks (PNNs) proved to be a superior and more robust knowledge transfer mechanism. This work marks the first attempt to employ PNNs in the context of DRL-based flow control. By explicitly preserving previously acquired knowledge in frozen columns and enabling lateral transfer through learnable adapter connections, PNNs consistently delivered stable and efficient knowledge transfer, which proved robust against overfitting. The quantitative assessments confirmed maintaining high performance even when the source and target environments differed substantially. The Average Perturbation Sensitivity (APS) analysis further illuminated how PNNs leverage low-level features from the source domain while adapting higher-level representations to the specific demands of the target task.

The findings highlight the potential of PNNs to enable robust, scalable, and computationally efficient DRL-based flow control. Their ability to generalize across diverse fidelity levels, physical conditions, and control objectives represents a significant step toward deploying DRL in complex, real-world fluid dynamics applications. This research motivates further exploration of more intricate PNN architectures and their application to more complex flow fields, such as turbulent flows, potentially accelerating the discovery of effective control strategies. Although this study focused on bi-fidelity transfer, the PNN framework is not restricted to two fidelity levels. Future work could examine multilevel knowledge transfer across a spectrum of fidelities (e.g., URANS, DES, LES, DNS).

\section*{Acknowledgements}
\label{sec:acknowledgements}

The research presented was carried out as a part of the ``Swedish Centre for Sustainable Hydropower - SVC''. SVC has been established by the Swedish Energy Agency, Energiforsk and Svenska kraftnät together with Luleå University of Technology, Uppsala University, KTH Royal Institute of Technology, Chalmers University of Technology, Karlstad University, Umeå University and Lund University, \href{https://svc.energiforsk.se/}{\texttt{svc.energiforsk.se}}.

The computations were enabled by resources provided by the National Academic Infrastructure for Supercomputing in Sweden (NAISS) at NSC and C3SE partially funded by the Swedish Research Council through grant agreement no. 2022-06725.

The author would like to especially express sincere appreciation to Professor Håkan Nilsson (Chalmers University of Technology) for his valuable contributions to funding acquisition and computational resource allocation, as well as for his continuous support and insightful discussions throughout the project.

\section*{Data Availability Statement}

All the codes and cases used in this study are available in the open-source GitHub repository \texttt{TL\_DRL}: \url{https://github.com/salehisaeed/TL_DRL}.

\section*{Declaration of Interests}
The author reports no conflict of interest.


\begin{thebibliography}{50}%
\makeatletter
\providecommand \@ifxundefined [1]{%
 \@ifx{#1\undefined}
}%
\providecommand \@ifnum [1]{%
 \ifnum #1\expandafter \@firstoftwo
 \else \expandafter \@secondoftwo
 \fi
}%
\providecommand \@ifx [1]{%
 \ifx #1\expandafter \@firstoftwo
 \else \expandafter \@secondoftwo
 \fi
}%
\providecommand \natexlab [1]{#1}%
\providecommand \enquote  [1]{``#1''}%
\providecommand \bibnamefont  [1]{#1}%
\providecommand \bibfnamefont [1]{#1}%
\providecommand \citenamefont [1]{#1}%
\providecommand \href@noop [0]{\@secondoftwo}%
\providecommand \href [0]{\begingroup \@sanitize@url \@href}%
\providecommand \@href[1]{\@@startlink{#1}\@@href}%
\providecommand \@@href[1]{\endgroup#1\@@endlink}%
\providecommand \@sanitize@url [0]{\catcode `\\12\catcode `\$12\catcode
  `\&12\catcode `\#12\catcode `\^12\catcode `\_12\catcode `\%12\relax}%
\providecommand \@@startlink[1]{}%
\providecommand \@@endlink[0]{}%
\providecommand \url  [0]{\begingroup\@sanitize@url \@url }%
\providecommand \@url [1]{\endgroup\@href {#1}{\urlprefix }}%
\providecommand \urlprefix  [0]{URL }%
\providecommand \Eprint [0]{\href }%
\providecommand \doibase [0]{http://dx.doi.org/}%
\providecommand \selectlanguage [0]{\@gobble}%
\providecommand \bibinfo  [0]{\@secondoftwo}%
\providecommand \bibfield  [0]{\@secondoftwo}%
\providecommand \translation [1]{[#1]}%
\providecommand \BibitemOpen [0]{}%
\providecommand \bibitemStop [0]{}%
\providecommand \bibitemNoStop [0]{.\EOS\space}%
\providecommand \EOS [0]{\spacefactor3000\relax}%
\providecommand \BibitemShut  [1]{\csname bibitem#1\endcsname}%
\let\auto@bib@innerbib\@empty
\bibitem [{\citenamefont {Mnih}\ \emph {et~al.}(2015)\citenamefont {Mnih},
  \citenamefont {Kavukcuoglu}, \citenamefont {Silver}, \citenamefont {Rusu},
  \citenamefont {Veness}, \citenamefont {Bellemare}, \citenamefont {Graves},
  \citenamefont {Riedmiller}, \citenamefont {Fidjeland}, \citenamefont
  {Ostrovski}, \citenamefont {Petersen}, \citenamefont {Beattie}, \citenamefont
  {Sadik}, \citenamefont {Antonoglou}, \citenamefont {King}, \citenamefont
  {Kumaran}, \citenamefont {Wierstra}, \citenamefont {Legg},\ and\
  \citenamefont {Hassabis}}]{Mnih2015}%
  \BibitemOpen
  \bibfield  {author} {\bibinfo {author} {\bibfnamefont {V.}~\bibnamefont
  {Mnih}}, \bibinfo {author} {\bibfnamefont {K.}~\bibnamefont {Kavukcuoglu}},
  \bibinfo {author} {\bibfnamefont {D.}~\bibnamefont {Silver}}, \bibinfo
  {author} {\bibfnamefont {A.~A.}\ \bibnamefont {Rusu}}, \bibinfo {author}
  {\bibfnamefont {J.}~\bibnamefont {Veness}}, \bibinfo {author} {\bibfnamefont
  {M.~G.}\ \bibnamefont {Bellemare}}, \bibinfo {author} {\bibfnamefont
  {A.}~\bibnamefont {Graves}}, \bibinfo {author} {\bibfnamefont
  {M.}~\bibnamefont {Riedmiller}}, \bibinfo {author} {\bibfnamefont {A.~K.}\
  \bibnamefont {Fidjeland}}, \bibinfo {author} {\bibfnamefont {G.}~\bibnamefont
  {Ostrovski}}, \bibinfo {author} {\bibfnamefont {S.}~\bibnamefont {Petersen}},
  \bibinfo {author} {\bibfnamefont {C.}~\bibnamefont {Beattie}}, \bibinfo
  {author} {\bibfnamefont {A.}~\bibnamefont {Sadik}}, \bibinfo {author}
  {\bibfnamefont {I.}~\bibnamefont {Antonoglou}}, \bibinfo {author}
  {\bibfnamefont {H.}~\bibnamefont {King}}, \bibinfo {author} {\bibfnamefont
  {D.}~\bibnamefont {Kumaran}}, \bibinfo {author} {\bibfnamefont
  {D.}~\bibnamefont {Wierstra}}, \bibinfo {author} {\bibfnamefont
  {S.}~\bibnamefont {Legg}}, \ and\ \bibinfo {author} {\bibfnamefont
  {D.}~\bibnamefont {Hassabis}},\ }\bibfield  {title} {\enquote {\bibinfo
  {title} {{Human-level control through deep reinforcement learning}},}\ }\href
  {\doibase 10.1038/nature14236} {\bibfield  {journal} {\bibinfo  {journal}
  {Nature}\ }\textbf {\bibinfo {volume} {518}},\ \bibinfo {pages} {529--533}
  (\bibinfo {year} {2015})}\BibitemShut {NoStop}%
\bibitem [{\citenamefont {Silver}\ \emph {et~al.}(2017)\citenamefont {Silver},
  \citenamefont {Schrittwieser}, \citenamefont {Simonyan}, \citenamefont
  {ioannis Antonoglou}, \citenamefont {Huang}, \citenamefont {Guez},
  \citenamefont {Hubert}, \citenamefont {Baker}, \citenamefont {Lai},
  \citenamefont {Bolton}, \citenamefont {Chen}, \citenamefont {Lillicrap},
  \citenamefont {Hui}, \citenamefont {Sifre}, \citenamefont {van~den
  Driessche}, \citenamefont {Graepel},\ and\ \citenamefont
  {Hassabis}}]{Silver2017}%
  \BibitemOpen
  \bibfield  {author} {\bibinfo {author} {\bibfnamefont {D.}~\bibnamefont
  {Silver}}, \bibinfo {author} {\bibfnamefont {J.}~\bibnamefont
  {Schrittwieser}}, \bibinfo {author} {\bibfnamefont {K.}~\bibnamefont
  {Simonyan}}, \bibinfo {author} {\bibnamefont {ioannis Antonoglou}}, \bibinfo
  {author} {\bibfnamefont {A.}~\bibnamefont {Huang}}, \bibinfo {author}
  {\bibfnamefont {A.}~\bibnamefont {Guez}}, \bibinfo {author} {\bibfnamefont
  {T.}~\bibnamefont {Hubert}}, \bibinfo {author} {\bibfnamefont
  {L.}~\bibnamefont {Baker}}, \bibinfo {author} {\bibfnamefont
  {M.}~\bibnamefont {Lai}}, \bibinfo {author} {\bibfnamefont {A.}~\bibnamefont
  {Bolton}}, \bibinfo {author} {\bibfnamefont {Y.}~\bibnamefont {Chen}},
  \bibinfo {author} {\bibfnamefont {T.}~\bibnamefont {Lillicrap}}, \bibinfo
  {author} {\bibfnamefont {F.}~\bibnamefont {Hui}}, \bibinfo {author}
  {\bibfnamefont {L.}~\bibnamefont {Sifre}}, \bibinfo {author} {\bibfnamefont
  {G.}~\bibnamefont {van~den Driessche}}, \bibinfo {author} {\bibfnamefont
  {T.}~\bibnamefont {Graepel}}, \ and\ \bibinfo {author} {\bibfnamefont
  {D.}~\bibnamefont {Hassabis}},\ }\bibfield  {title} {\enquote {\bibinfo
  {title} {{Mastering the game of Go without human knowledge}},}\ }\href
  {\doibase 10.1038/nature24270} {\bibfield  {journal} {\bibinfo  {journal}
  {Nature Publishing Group}\ }\textbf {\bibinfo {volume} {550}} (\bibinfo
  {year} {2017}),\ 10.1038/nature24270}\BibitemShut {NoStop}%
\bibitem [{\citenamefont {Reddy}\ \emph {et~al.}(2016)\citenamefont {Reddy},
  \citenamefont {Celani}, \citenamefont {Sejnowski},\ and\ \citenamefont
  {Vergassola}}]{Reddy2016}%
  \BibitemOpen
  \bibfield  {author} {\bibinfo {author} {\bibfnamefont {G.}~\bibnamefont
  {Reddy}}, \bibinfo {author} {\bibfnamefont {A.}~\bibnamefont {Celani}},
  \bibinfo {author} {\bibfnamefont {T.~J.}\ \bibnamefont {Sejnowski}}, \ and\
  \bibinfo {author} {\bibfnamefont {M.}~\bibnamefont {Vergassola}},\ }\bibfield
   {title} {\enquote {\bibinfo {title} {{Learning to soar in turbulent
  environments}},}\ }\href {\doibase 10.1073/pnas.1606075113} {\bibfield
  {journal} {\bibinfo  {journal} {Proceedings of the National Academy of
  Sciences of the United States of America}\ }\textbf {\bibinfo {volume}
  {113}},\ \bibinfo {pages} {E4877--E4884} (\bibinfo {year}
  {2016})}\BibitemShut {NoStop}%
\bibitem [{\citenamefont {Novati}\ \emph {et~al.}(2017)\citenamefont {Novati},
  \citenamefont {Verma}, \citenamefont {Alexeev}, \citenamefont {Rossinelli},
  \citenamefont {{Van Rees}},\ and\ \citenamefont {Koumoutsakos}}]{Novati2017}%
  \BibitemOpen
  \bibfield  {author} {\bibinfo {author} {\bibfnamefont {G.}~\bibnamefont
  {Novati}}, \bibinfo {author} {\bibfnamefont {S.}~\bibnamefont {Verma}},
  \bibinfo {author} {\bibfnamefont {D.}~\bibnamefont {Alexeev}}, \bibinfo
  {author} {\bibfnamefont {D.}~\bibnamefont {Rossinelli}}, \bibinfo {author}
  {\bibfnamefont {W.~M.}\ \bibnamefont {{Van Rees}}}, \ and\ \bibinfo {author}
  {\bibfnamefont {P.}~\bibnamefont {Koumoutsakos}},\ }\bibfield  {title}
  {\enquote {\bibinfo {title} {{Synchronisation through learning for two
  self-propelled swimmers}},}\ }\href {\doibase 10.1088/1748-3190/aa6311}
  {\bibfield  {journal} {\bibinfo  {journal} {Bioinspiration and Biomimetics}\
  }\textbf {\bibinfo {volume} {12}} (\bibinfo {year} {2017}),\
  10.1088/1748-3190/aa6311}\BibitemShut {NoStop}%
\bibitem [{\citenamefont {Verma}, \citenamefont {Novati},\ and\ \citenamefont
  {Koumoutsakos}(2018)}]{Verma2018}%
  \BibitemOpen
  \bibfield  {author} {\bibinfo {author} {\bibfnamefont {S.}~\bibnamefont
  {Verma}}, \bibinfo {author} {\bibfnamefont {G.}~\bibnamefont {Novati}}, \
  and\ \bibinfo {author} {\bibfnamefont {P.}~\bibnamefont {Koumoutsakos}},\
  }\bibfield  {title} {\enquote {\bibinfo {title} {{Efficient collective
  swimming by harnessing vortices through deep reinforcement learning}},}\
  }\href {\doibase 10.1073/pnas.1800923115} {\bibfield  {journal} {\bibinfo
  {journal} {Proceedings of the National Academy of Sciences of the United
  States of America}\ }\textbf {\bibinfo {volume} {115}},\ \bibinfo {pages}
  {5849--5854} (\bibinfo {year} {2018})}\BibitemShut {NoStop}%
\bibitem [{\citenamefont {Ma}\ \emph {et~al.}(2018)\citenamefont {Ma},
  \citenamefont {Tian}, \citenamefont {Pan}, \citenamefont {Ren},\ and\
  \citenamefont {Manocha}}]{Ma2018}%
  \BibitemOpen
  \bibfield  {author} {\bibinfo {author} {\bibfnamefont {P.}~\bibnamefont
  {Ma}}, \bibinfo {author} {\bibfnamefont {Y.}~\bibnamefont {Tian}}, \bibinfo
  {author} {\bibfnamefont {Z.}~\bibnamefont {Pan}}, \bibinfo {author}
  {\bibfnamefont {B.}~\bibnamefont {Ren}}, \ and\ \bibinfo {author}
  {\bibfnamefont {D.}~\bibnamefont {Manocha}},\ }\bibfield  {title} {\enquote
  {\bibinfo {title} {{Fluid directed rigid body control using deep
  reinforcement learning}},}\ }\href {\doibase 10.1145/3197517.3201334}
  {\bibfield  {journal} {\bibinfo  {journal} {ACM Transactions on Graphics}\
  }\textbf {\bibinfo {volume} {37}} (\bibinfo {year} {2018}),\
  10.1145/3197517.3201334}\BibitemShut {NoStop}%
\bibitem [{\citenamefont {Lee}\ \emph {et~al.}(2018)\citenamefont {Lee},
  \citenamefont {Balu}, \citenamefont {Stoecklein}, \citenamefont
  {Ganapathysubramanian},\ and\ \citenamefont {Sarkar}}]{YeowLee}%
  \BibitemOpen
  \bibfield  {author} {\bibinfo {author} {\bibfnamefont {X.~Y.}\ \bibnamefont
  {Lee}}, \bibinfo {author} {\bibfnamefont {A.}~\bibnamefont {Balu}}, \bibinfo
  {author} {\bibfnamefont {D.}~\bibnamefont {Stoecklein}}, \bibinfo {author}
  {\bibfnamefont {B.}~\bibnamefont {Ganapathysubramanian}}, \ and\ \bibinfo
  {author} {\bibfnamefont {S.}~\bibnamefont {Sarkar}},\ }\bibfield  {title}
  {\enquote {\bibinfo {title} {Flow shape design for microfluidic devices using
  deep reinforcement learning},}\ }\href@noop {} {\bibfield  {journal}
  {\bibinfo  {journal} {CoRR}\ }\textbf {\bibinfo {volume} {abs/1811.12444}}
  (\bibinfo {year} {2018})},\ \Eprint {http://arxiv.org/abs/1811.12444}
  {1811.12444} \BibitemShut {NoStop}%
\bibitem [{\citenamefont {Viquerat}\ \emph {et~al.}(2021)\citenamefont
  {Viquerat}, \citenamefont {Rabault}, \citenamefont {Kuhnle}, \citenamefont
  {Ghraieb}, \citenamefont {Larcher},\ and\ \citenamefont
  {Hachem}}]{Viquerat2021}%
  \BibitemOpen
  \bibfield  {author} {\bibinfo {author} {\bibfnamefont {J.}~\bibnamefont
  {Viquerat}}, \bibinfo {author} {\bibfnamefont {J.}~\bibnamefont {Rabault}},
  \bibinfo {author} {\bibfnamefont {A.}~\bibnamefont {Kuhnle}}, \bibinfo
  {author} {\bibfnamefont {H.}~\bibnamefont {Ghraieb}}, \bibinfo {author}
  {\bibfnamefont {A.}~\bibnamefont {Larcher}}, \ and\ \bibinfo {author}
  {\bibfnamefont {E.}~\bibnamefont {Hachem}},\ }\bibfield  {title} {\enquote
  {\bibinfo {title} {{Direct shape optimization through deep reinforcement
  learning}},}\ }\href {\doibase 10.1016/j.jcp.2020.110080} {\bibfield
  {journal} {\bibinfo  {journal} {Journal of Computational Physics}\ }\textbf
  {\bibinfo {volume} {428}} (\bibinfo {year} {2021}),\
  10.1016/j.jcp.2020.110080}\BibitemShut {NoStop}%
\bibitem [{\citenamefont {Rabault}\ \emph {et~al.}(2019)\citenamefont
  {Rabault}, \citenamefont {Kuchta}, \citenamefont {Jensen}, \citenamefont
  {R{\'{e}}glade},\ and\ \citenamefont {Cerardi}}]{Rabault2019a}%
  \BibitemOpen
  \bibfield  {author} {\bibinfo {author} {\bibfnamefont {J.}~\bibnamefont
  {Rabault}}, \bibinfo {author} {\bibfnamefont {M.}~\bibnamefont {Kuchta}},
  \bibinfo {author} {\bibfnamefont {A.}~\bibnamefont {Jensen}}, \bibinfo
  {author} {\bibfnamefont {U.}~\bibnamefont {R{\'{e}}glade}}, \ and\ \bibinfo
  {author} {\bibfnamefont {N.}~\bibnamefont {Cerardi}},\ }\bibfield  {title}
  {\enquote {\bibinfo {title} {{Artificial neural networks trained through deep
  reinforcement learning discover control strategies for active flow
  control}},}\ }\href {\doibase 10.1017/jfm.2019.62} {\bibfield  {journal}
  {\bibinfo  {journal} {Journal of Fluid Mechanics}\ }\textbf {\bibinfo
  {volume} {865}},\ \bibinfo {pages} {281--302} (\bibinfo {year}
  {2019})}\BibitemShut {NoStop}%
\bibitem [{\citenamefont {Ren}, \citenamefont {Rabault},\ and\ \citenamefont
  {Tang}(2021)}]{Ren2021}%
  \BibitemOpen
  \bibfield  {author} {\bibinfo {author} {\bibfnamefont {F.}~\bibnamefont
  {Ren}}, \bibinfo {author} {\bibfnamefont {J.}~\bibnamefont {Rabault}}, \ and\
  \bibinfo {author} {\bibfnamefont {H.}~\bibnamefont {Tang}},\ }\bibfield
  {title} {\enquote {\bibinfo {title} {{Applying deep reinforcement learning to
  active flow control in weakly turbulent conditions}},}\ }\href {\doibase
  10.1063/5.0037371} {\bibfield  {journal} {\bibinfo  {journal} {Physics of
  Fluids}\ }\textbf {\bibinfo {volume} {33}},\ \bibinfo {pages} {37121}
  (\bibinfo {year} {2021})}\BibitemShut {NoStop}%
\bibitem [{\citenamefont {Fan}\ \emph {et~al.}(2020)\citenamefont {Fan},
  \citenamefont {Yang}, \citenamefont {Wang}, \citenamefont {Triantafyllou},\
  and\ \citenamefont {Karniadakis}}]{Fan2020}%
  \BibitemOpen
  \bibfield  {author} {\bibinfo {author} {\bibfnamefont {D.}~\bibnamefont
  {Fan}}, \bibinfo {author} {\bibfnamefont {L.}~\bibnamefont {Yang}}, \bibinfo
  {author} {\bibfnamefont {Z.}~\bibnamefont {Wang}}, \bibinfo {author}
  {\bibfnamefont {M.~S.}\ \bibnamefont {Triantafyllou}}, \ and\ \bibinfo
  {author} {\bibfnamefont {G.~E.}\ \bibnamefont {Karniadakis}},\ }\bibfield
  {title} {\enquote {\bibinfo {title} {{Reinforcement learning for bluff body
  active flow control in experiments and simulations}},}\ }\href {\doibase
  10.1073/pnas.2004939117} {\bibfield  {journal} {\bibinfo  {journal}
  {Proceedings of the National Academy of Sciences of the United States of
  America}\ }\textbf {\bibinfo {volume} {117}},\ \bibinfo {pages}
  {26091--26098} (\bibinfo {year} {2020})}\BibitemShut {NoStop}%
\bibitem [{\citenamefont {Vinuesa}\ \emph {et~al.}(2022)\citenamefont
  {Vinuesa}, \citenamefont {Lehmkuhl}, \citenamefont {Lozano-Durán},\ and\
  \citenamefont {Rabault}}]{fluids7020062}%
  \BibitemOpen
  \bibfield  {author} {\bibinfo {author} {\bibfnamefont {R.}~\bibnamefont
  {Vinuesa}}, \bibinfo {author} {\bibfnamefont {O.}~\bibnamefont {Lehmkuhl}},
  \bibinfo {author} {\bibfnamefont {A.}~\bibnamefont {Lozano-Durán}}, \ and\
  \bibinfo {author} {\bibfnamefont {J.}~\bibnamefont {Rabault}},\ }\bibfield
  {title} {\enquote {\bibinfo {title} {Flow control in wings and discovery of
  novel approaches via deep reinforcement learning},}\ }\href {\doibase
  10.3390/fluids7020062} {\bibfield  {journal} {\bibinfo  {journal} {Fluids}\
  }\textbf {\bibinfo {volume} {7}} (\bibinfo {year} {2022}),\
  10.3390/fluids7020062}\BibitemShut {NoStop}%
\bibitem [{\citenamefont {Bucci}\ \emph {et~al.}(2019)\citenamefont {Bucci},
  \citenamefont {Semeraro}, \citenamefont {Allauzen}, \citenamefont
  {Wisniewski}, \citenamefont {Cordier},\ and\ \citenamefont
  {Mathelin}}]{Bucci2019}%
  \BibitemOpen
  \bibfield  {author} {\bibinfo {author} {\bibfnamefont {M.~A.}\ \bibnamefont
  {Bucci}}, \bibinfo {author} {\bibfnamefont {O.}~\bibnamefont {Semeraro}},
  \bibinfo {author} {\bibfnamefont {A.}~\bibnamefont {Allauzen}}, \bibinfo
  {author} {\bibfnamefont {G.}~\bibnamefont {Wisniewski}}, \bibinfo {author}
  {\bibfnamefont {L.}~\bibnamefont {Cordier}}, \ and\ \bibinfo {author}
  {\bibfnamefont {L.}~\bibnamefont {Mathelin}},\ }\bibfield  {title} {\enquote
  {\bibinfo {title} {{Control of chaotic systems by deep reinforcement
  learning}},}\ }\href {\doibase 10.1098/rspa.2019.0351} {\bibfield  {journal}
  {\bibinfo  {journal} {Proceedings of the Royal Society A: Mathematical,
  Physical and Engineering Sciences}\ }\textbf {\bibinfo {volume} {475}}
  (\bibinfo {year} {2019}),\ 10.1098/rspa.2019.0351}\BibitemShut {NoStop}%
\bibitem [{\citenamefont {Peitz}\ \emph {et~al.}(2023)\citenamefont {Peitz},
  \citenamefont {Stenner}, \citenamefont {Chidananda}, \citenamefont
  {Wallscheid}, \citenamefont {Brunton},\ and\ \citenamefont
  {Taira}}]{Peitz2023}%
  \BibitemOpen
  \bibfield  {author} {\bibinfo {author} {\bibfnamefont {S.}~\bibnamefont
  {Peitz}}, \bibinfo {author} {\bibfnamefont {J.}~\bibnamefont {Stenner}},
  \bibinfo {author} {\bibfnamefont {V.}~\bibnamefont {Chidananda}}, \bibinfo
  {author} {\bibfnamefont {O.}~\bibnamefont {Wallscheid}}, \bibinfo {author}
  {\bibfnamefont {S.~L.}\ \bibnamefont {Brunton}}, \ and\ \bibinfo {author}
  {\bibfnamefont {K.}~\bibnamefont {Taira}},\ }\bibfield  {title} {\enquote
  {\bibinfo {title} {{Distributed Control of Partial Differential Equations
  Using Convolutional Reinforcement Learning}},}\ }\href {\doibase
  10.1016/j.physd.2024.134096} {\bibfield  {journal} {\bibinfo  {journal}
  {Physica D: Nonlinear Phenomena}\ }\textbf {\bibinfo {volume} {461}},\
  \bibinfo {pages} {134096} (\bibinfo {year} {2023})}\BibitemShut {NoStop}%
\bibitem [{\citenamefont {Sonoda}\ \emph {et~al.}(2023)\citenamefont {Sonoda},
  \citenamefont {Liu}, \citenamefont {Itoh},\ and\ \citenamefont
  {Hasegawa}}]{Sonoda2023a}%
  \BibitemOpen
  \bibfield  {author} {\bibinfo {author} {\bibfnamefont {T.}~\bibnamefont
  {Sonoda}}, \bibinfo {author} {\bibfnamefont {Z.}~\bibnamefont {Liu}},
  \bibinfo {author} {\bibfnamefont {T.}~\bibnamefont {Itoh}}, \ and\ \bibinfo
  {author} {\bibfnamefont {Y.}~\bibnamefont {Hasegawa}},\ }\bibfield  {title}
  {\enquote {\bibinfo {title} {{Reinforcement learning of control strategies
  for reducing skin friction drag in a fully developed turbulent channel
  flow}},}\ }\href {\doibase 10.1017/jfm.2023.147} {\bibfield  {journal}
  {\bibinfo  {journal} {Journal of Fluid Mechanics}\ }\textbf {\bibinfo
  {volume} {960}} (\bibinfo {year} {2023}),\ 10.1017/jfm.2023.147}\BibitemShut
  {NoStop}%
\bibitem [{\citenamefont {Font}\ \emph {et~al.}(2025)\citenamefont {Font},
  \citenamefont {Alcántara-Ávila}, \citenamefont {Rabault}, \citenamefont
  {Vinuesa},\ and\ \citenamefont
  {Lehmkuhl}}]{fontDeepReinforcementLearning2025}%
  \BibitemOpen
  \bibfield  {author} {\bibinfo {author} {\bibfnamefont {B.}~\bibnamefont
  {Font}}, \bibinfo {author} {\bibfnamefont {F.}~\bibnamefont
  {Alcántara-Ávila}}, \bibinfo {author} {\bibfnamefont {J.}~\bibnamefont
  {Rabault}}, \bibinfo {author} {\bibfnamefont {R.}~\bibnamefont {Vinuesa}}, \
  and\ \bibinfo {author} {\bibfnamefont {O.}~\bibnamefont {Lehmkuhl}},\
  }\bibfield  {title} {\enquote {\bibinfo {title} {Deep reinforcement learning
  for active flow control in a turbulent separation bubble},}\ }\href {\doibase
  10.1038/s41467-025-56408-6} {\bibfield  {journal} {\bibinfo  {journal}
  {Nature Communications}\ }\textbf {\bibinfo {volume} {16}},\ \bibinfo {pages}
  {1422} (\bibinfo {year} {2025})}\BibitemShut {NoStop}%
\bibitem [{\citenamefont {Zhou}, \citenamefont {Zhang},\ and\ \citenamefont
  {Zhu}(2025)}]{zhouReinforcementlearningbasedControlTurbulent2025a}%
  \BibitemOpen
  \bibfield  {author} {\bibinfo {author} {\bibfnamefont {Z.}~\bibnamefont
  {Zhou}}, \bibinfo {author} {\bibfnamefont {M.}~\bibnamefont {Zhang}}, \ and\
  \bibinfo {author} {\bibfnamefont {X.}~\bibnamefont {Zhu}},\ }\bibfield
  {title} {\enquote {\bibinfo {title} {Reinforcement-learning-based control of
  turbulent channel flows at high {{Reynolds}} numbers},}\ }\href {\doibase
  10.1017/jfm.2025.27} {\bibfield  {journal} {\bibinfo  {journal} {Journal of
  Fluid Mechanics}\ }\textbf {\bibinfo {volume} {1006}},\ \bibinfo {pages}
  {A12} (\bibinfo {year} {2025})}\BibitemShut {NoStop}%
\bibitem [{\citenamefont {Rabault}\ and\ \citenamefont
  {Kuhnle}(2019)}]{Rabault2019}%
  \BibitemOpen
  \bibfield  {author} {\bibinfo {author} {\bibfnamefont {J.}~\bibnamefont
  {Rabault}}\ and\ \bibinfo {author} {\bibfnamefont {A.}~\bibnamefont
  {Kuhnle}},\ }\bibfield  {title} {\enquote {\bibinfo {title} {{Accelerating
  deep reinforcement learning strategies of flow control through a
  multi-environment approach}},}\ }\href {\doibase 10.1063/1.5116415}
  {\bibfield  {journal} {\bibinfo  {journal} {Physics of Fluids}\ }\textbf
  {\bibinfo {volume} {31}} (\bibinfo {year} {2019}),\
  10.1063/1.5116415}\BibitemShut {NoStop}%
\bibitem [{\citenamefont {Chatzimanolakis}, \citenamefont {Weber},\ and\
  \citenamefont
  {Koumoutsakos}(2024)}]{chatzimanolakisLearningTwoDimensions2024}%
  \BibitemOpen
  \bibfield  {author} {\bibinfo {author} {\bibfnamefont {M.}~\bibnamefont
  {Chatzimanolakis}}, \bibinfo {author} {\bibfnamefont {P.}~\bibnamefont
  {Weber}}, \ and\ \bibinfo {author} {\bibfnamefont {P.}~\bibnamefont
  {Koumoutsakos}},\ }\bibfield  {title} {\enquote {\bibinfo {title} {Learning
  in two dimensions and controlling in three},}\ }\href {\doibase
  10.1103/PhysRevFluids.9.043902} {\bibfield  {journal} {\bibinfo  {journal}
  {Physical Review Fluids}\ }\textbf {\bibinfo {volume} {9}},\ \bibinfo {pages}
  {043902} (\bibinfo {year} {2024})}\BibitemShut {NoStop}%
\bibitem [{\citenamefont {Suárez}\ \emph {et~al.}(2025)\citenamefont
  {Suárez}, \citenamefont {Alcántara-Ávila}, \citenamefont {Miró},
  \citenamefont {Rabault}, \citenamefont {Font}, \citenamefont {Lehmkuhl},\
  and\ \citenamefont {Vinuesa}}]{suarezActiveFlowControl2025}%
  \BibitemOpen
  \bibfield  {author} {\bibinfo {author} {\bibfnamefont {P.}~\bibnamefont
  {Suárez}}, \bibinfo {author} {\bibfnamefont {F.}~\bibnamefont
  {Alcántara-Ávila}}, \bibinfo {author} {\bibfnamefont {A.}~\bibnamefont
  {Miró}}, \bibinfo {author} {\bibfnamefont {J.}~\bibnamefont {Rabault}},
  \bibinfo {author} {\bibfnamefont {B.}~\bibnamefont {Font}}, \bibinfo {author}
  {\bibfnamefont {O.}~\bibnamefont {Lehmkuhl}}, \ and\ \bibinfo {author}
  {\bibfnamefont {R.}~\bibnamefont {Vinuesa}},\ }\bibfield  {title} {\enquote
  {\bibinfo {title} {Active flow control for drag reduction through multi-agent
  reinforcement learning on a turbulent cylinder at $\mathrm{Re}_{D}=3900$},}\
  }\href {\doibase 10.1007/s10494-025-00642-x} {\bibfield  {journal} {\bibinfo
  {journal} {Flow, Turbulence and Combustion}\ } (\bibinfo {year} {2025}),\
  10.1007/s10494-025-00642-x}\BibitemShut {NoStop}%
\bibitem [{\citenamefont {Guastoni}\ \emph {et~al.}(2023)\citenamefont
  {Guastoni}, \citenamefont {Rabault}, \citenamefont {Schlatter}, \citenamefont
  {Azizpour},\ and\ \citenamefont {Vinuesa}}]{Guastoni2023}%
  \BibitemOpen
  \bibfield  {author} {\bibinfo {author} {\bibfnamefont {L.}~\bibnamefont
  {Guastoni}}, \bibinfo {author} {\bibfnamefont {J.}~\bibnamefont {Rabault}},
  \bibinfo {author} {\bibfnamefont {P.}~\bibnamefont {Schlatter}}, \bibinfo
  {author} {\bibfnamefont {H.}~\bibnamefont {Azizpour}}, \ and\ \bibinfo
  {author} {\bibfnamefont {R.}~\bibnamefont {Vinuesa}},\ }\bibfield  {title}
  {\enquote {\bibinfo {title} {{Deep reinforcement learning for turbulent drag
  reduction in channel flows}},}\ }\href {\doibase
  10.1140/epje/s10189-023-00285-8} {\bibfield  {journal} {\bibinfo  {journal}
  {European Physical Journal E}\ }\textbf {\bibinfo {volume} {46}} (\bibinfo
  {year} {2023}),\ 10.1140/epje/s10189-023-00285-8}\BibitemShut {NoStop}%
\bibitem [{\citenamefont {{Giselle
  Fern{\'a}ndez-Godino}}(2023)}]{MF_review2023}%
  \BibitemOpen
  \bibfield  {author} {\bibinfo {author} {\bibfnamefont {M.}~\bibnamefont
  {{Giselle Fern{\'a}ndez-Godino}}},\ }\bibfield  {title} {\enquote {\bibinfo
  {title} {Review of multi-fidelity models},}\ }\href {\doibase
  10.3934/acse.2023015} {\bibfield  {journal} {\bibinfo  {journal} {Advances in
  Computational Science and Engineering}\ }\textbf {\bibinfo {volume} {1}},\
  \bibinfo {pages} {351--400} (\bibinfo {year} {2023})}\BibitemShut {NoStop}%
\bibitem [{\citenamefont {Taylor}\ and\ \citenamefont
  {Stone}(2009)}]{Taylor2009}%
  \BibitemOpen
  \bibfield  {author} {\bibinfo {author} {\bibfnamefont {M.~E.}\ \bibnamefont
  {Taylor}}\ and\ \bibinfo {author} {\bibfnamefont {P.}~\bibnamefont {Stone}},\
  }\bibfield  {title} {\enquote {\bibinfo {title} {{Transfer learning for
  reinforcement learning domains: A survey}},}\ }\href@noop {} {\bibfield
  {journal} {\bibinfo  {journal} {Journal of Machine Learning Research}\
  }\textbf {\bibinfo {volume} {10}},\ \bibinfo {pages} {1633--1685} (\bibinfo
  {year} {2009})}\BibitemShut {NoStop}%
\bibitem [{\citenamefont {Zhu}\ \emph {et~al.}(2023)\citenamefont {Zhu},
  \citenamefont {Lin}, \citenamefont {Jain},\ and\ \citenamefont
  {Zhou}}]{Zhu2023}%
  \BibitemOpen
  \bibfield  {author} {\bibinfo {author} {\bibfnamefont {Z.}~\bibnamefont
  {Zhu}}, \bibinfo {author} {\bibfnamefont {K.}~\bibnamefont {Lin}}, \bibinfo
  {author} {\bibfnamefont {A.~K.}\ \bibnamefont {Jain}}, \ and\ \bibinfo
  {author} {\bibfnamefont {J.}~\bibnamefont {Zhou}},\ }\bibfield  {title}
  {\enquote {\bibinfo {title} {{Transfer Learning in Deep Reinforcement
  Learning: A Survey}},}\ }\href {\doibase 10.1109/TPAMI.2023.3292075}
  {\bibfield  {journal} {\bibinfo  {journal} {IEEE Transactions on Pattern
  Analysis and Machine Intelligence}\ }\textbf {\bibinfo {volume} {45}},\
  \bibinfo {pages} {13344--13362} (\bibinfo {year} {2023})},\ \Eprint
  {http://arxiv.org/abs/2009.07888} {2009.07888} \BibitemShut {NoStop}%
\bibitem [{\citenamefont {Hinton}\ and\ \citenamefont
  {Salakhutdinov}(2006)}]{Hinton2006Reducing}%
  \BibitemOpen
  \bibfield  {author} {\bibinfo {author} {\bibfnamefont {G.~E.}\ \bibnamefont
  {Hinton}}\ and\ \bibinfo {author} {\bibfnamefont {R.~R.}\ \bibnamefont
  {Salakhutdinov}},\ }\bibfield  {title} {\enquote {\bibinfo {title} {Reducing
  the {{Dimensionality}} of {{Data}} with {{Neural Networks}}},}\ }\href
  {\doibase 10.1126/science.1127647} {\bibfield  {journal} {\bibinfo  {journal}
  {Science}\ }\textbf {\bibinfo {volume} {313}},\ \bibinfo {pages} {504--507}
  (\bibinfo {year} {2006})}\BibitemShut {NoStop}%
\bibitem [{\citenamefont {Bhola}\ \emph {et~al.}(2023)\citenamefont {Bhola},
  \citenamefont {Pawar}, \citenamefont {Balaprakash},\ and\ \citenamefont
  {Maulik}}]{bholaMultifidelityReinforcementLearning2023}%
  \BibitemOpen
  \bibfield  {author} {\bibinfo {author} {\bibfnamefont {S.}~\bibnamefont
  {Bhola}}, \bibinfo {author} {\bibfnamefont {S.}~\bibnamefont {Pawar}},
  \bibinfo {author} {\bibfnamefont {P.}~\bibnamefont {Balaprakash}}, \ and\
  \bibinfo {author} {\bibfnamefont {R.}~\bibnamefont {Maulik}},\ }\bibfield
  {title} {\enquote {\bibinfo {title} {Multi-fidelity reinforcement learning
  framework for shape optimization},}\ }\href {\doibase
  10.1016/j.jcp.2023.112018} {\bibfield  {journal} {\bibinfo  {journal}
  {Journal of Computational Physics}\ }\textbf {\bibinfo {volume} {482}},\
  \bibinfo {pages} {112018} (\bibinfo {year} {2023})}\BibitemShut {NoStop}%
\bibitem [{\citenamefont {Wang}\ \emph {et~al.}(2022)\citenamefont {Wang},
  \citenamefont {Hua}, \citenamefont {Aubry}, \citenamefont {Chen},
  \citenamefont {Wu},\ and\ \citenamefont
  {Cui}}]{wangAcceleratingImprovingDeep2022}%
  \BibitemOpen
  \bibfield  {author} {\bibinfo {author} {\bibfnamefont {Y.-Z.}\ \bibnamefont
  {Wang}}, \bibinfo {author} {\bibfnamefont {Y.}~\bibnamefont {Hua}}, \bibinfo
  {author} {\bibfnamefont {N.}~\bibnamefont {Aubry}}, \bibinfo {author}
  {\bibfnamefont {Z.-H.}\ \bibnamefont {Chen}}, \bibinfo {author}
  {\bibfnamefont {W.-T.}\ \bibnamefont {Wu}}, \ and\ \bibinfo {author}
  {\bibfnamefont {J.}~\bibnamefont {Cui}},\ }\bibfield  {title} {\enquote
  {\bibinfo {title} {Accelerating and improving deep reinforcement
  learning-based active flow control: {{Transfer}} training of policy
  network},}\ }\href {\doibase 10.1063/5.0099699} {\bibfield  {journal}
  {\bibinfo  {journal} {Physics of Fluids}\ }\textbf {\bibinfo {volume} {34}},\
  \bibinfo {pages} {073609} (\bibinfo {year} {2022})}\BibitemShut {NoStop}%
\bibitem [{\citenamefont {Wang}\ \emph {et~al.}(2023)\citenamefont {Wang},
  \citenamefont {Fan}, \citenamefont {Jiang}, \citenamefont {Triantafyllou},\
  and\ \citenamefont {Karniadakis}}]{wangDeepReinforcementTransfer2023}%
  \BibitemOpen
  \bibfield  {author} {\bibinfo {author} {\bibfnamefont {Z.}~\bibnamefont
  {Wang}}, \bibinfo {author} {\bibfnamefont {D.}~\bibnamefont {Fan}}, \bibinfo
  {author} {\bibfnamefont {X.}~\bibnamefont {Jiang}}, \bibinfo {author}
  {\bibfnamefont {M.~S.}\ \bibnamefont {Triantafyllou}}, \ and\ \bibinfo
  {author} {\bibfnamefont {G.~E.}\ \bibnamefont {Karniadakis}},\ }\bibfield
  {title} {\enquote {\bibinfo {title} {Deep reinforcement transfer learning of
  active control for bluff body flows at high {{Reynolds}} number},}\ }\href
  {\doibase 10.1017/jfm.2023.637} {\bibfield  {journal} {\bibinfo  {journal}
  {Journal of Fluid Mechanics}\ }\textbf {\bibinfo {volume} {973}},\ \bibinfo
  {pages} {A32} (\bibinfo {year} {2023})}\BibitemShut {NoStop}%
\bibitem [{\citenamefont {He}\ \emph {et~al.}(2023)\citenamefont {He},
  \citenamefont {Wang}, \citenamefont {Hua}, \citenamefont {Chen},
  \citenamefont {Li},\ and\ \citenamefont
  {Wu}}]{PolicyTransferReinforcement2023}%
  \BibitemOpen
  \bibfield  {author} {\bibinfo {author} {\bibfnamefont {X.-J.}\ \bibnamefont
  {He}}, \bibinfo {author} {\bibfnamefont {Y.-Z.}\ \bibnamefont {Wang}},
  \bibinfo {author} {\bibfnamefont {Y.}~\bibnamefont {Hua}}, \bibinfo {author}
  {\bibfnamefont {Z.-H.}\ \bibnamefont {Chen}}, \bibinfo {author}
  {\bibfnamefont {Y.-B.}\ \bibnamefont {Li}}, \ and\ \bibinfo {author}
  {\bibfnamefont {W.-T.}\ \bibnamefont {Wu}},\ }\bibfield  {title} {\enquote
  {\bibinfo {title} {Policy transfer of reinforcement learning-based flow
  control: From two- to three-dimensional environment},}\ }\href {\doibase
  10.1063/5.0147190} {\bibfield  {journal} {\bibinfo  {journal} {Physics of
  Fluids}\ }\textbf {\bibinfo {volume} {35}},\ \bibinfo {pages} {055116}
  (\bibinfo {year} {2023})}\BibitemShut {NoStop}%
\bibitem [{\citenamefont {Yan}\ \emph {et~al.}(2025)\citenamefont {Yan},
  \citenamefont {Wang}, \citenamefont {Hu}, \citenamefont {Chen},\ and\
  \citenamefont {Noack}}]{yanDeepReinforcementCrossdomain2025}%
  \BibitemOpen
  \bibfield  {author} {\bibinfo {author} {\bibfnamefont {L.}~\bibnamefont
  {Yan}}, \bibinfo {author} {\bibfnamefont {Q.}~\bibnamefont {Wang}}, \bibinfo
  {author} {\bibfnamefont {G.}~\bibnamefont {Hu}}, \bibinfo {author}
  {\bibfnamefont {W.}~\bibnamefont {Chen}}, \ and\ \bibinfo {author}
  {\bibfnamefont {B.~R.}\ \bibnamefont {Noack}},\ }\bibfield  {title} {\enquote
  {\bibinfo {title} {Deep reinforcement cross-domain transfer learning of
  active flow control for three-dimensional bluff body flow},}\ }\href
  {\doibase 10.1016/j.jcp.2025.113893} {\bibfield  {journal} {\bibinfo
  {journal} {Journal of Computational Physics}\ }\textbf {\bibinfo {volume}
  {529}},\ \bibinfo {pages} {113893} (\bibinfo {year} {2025})}\BibitemShut
  {NoStop}%
\bibitem [{\citenamefont {Campos}\ \emph {et~al.}(2021)\citenamefont {Campos},
  \citenamefont {Sprechmann}, \citenamefont {Hansen}, \citenamefont {Barreto},
  \citenamefont {Kapturowski}, \citenamefont {Vitvitskyi}, \citenamefont
  {Badia},\ and\ \citenamefont {Blundell}}]{Campos2021}%
  \BibitemOpen
  \bibfield  {author} {\bibinfo {author} {\bibfnamefont {V.}~\bibnamefont
  {Campos}}, \bibinfo {author} {\bibfnamefont {P.}~\bibnamefont {Sprechmann}},
  \bibinfo {author} {\bibfnamefont {S.}~\bibnamefont {Hansen}}, \bibinfo
  {author} {\bibfnamefont {A.}~\bibnamefont {Barreto}}, \bibinfo {author}
  {\bibfnamefont {S.}~\bibnamefont {Kapturowski}}, \bibinfo {author}
  {\bibfnamefont {A.}~\bibnamefont {Vitvitskyi}}, \bibinfo {author}
  {\bibfnamefont {A.~P.}\ \bibnamefont {Badia}}, \ and\ \bibinfo {author}
  {\bibfnamefont {C.}~\bibnamefont {Blundell}},\ }\href
  {http://arxiv.org/abs/2102.13515} {\enquote {\bibinfo {title} {{Beyond
  Fine-Tuning: Transferring Behavior in Reinforcement Learning}},}\ } (\bibinfo
  {year} {2021}),\ \Eprint {http://arxiv.org/abs/2102.13515} {2102.13515}
  \BibitemShut {NoStop}%
\bibitem [{\citenamefont {Rusu}\ \emph {et~al.}(2016)\citenamefont {Rusu},
  \citenamefont {Rabinowitz}, \citenamefont {Desjardins}, \citenamefont
  {Soyer}, \citenamefont {Kirkpatrick}, \citenamefont {Kavukcuoglu},
  \citenamefont {Pascanu},\ and\ \citenamefont {Hadsell}}]{PNN2016}%
  \BibitemOpen
  \bibfield  {author} {\bibinfo {author} {\bibfnamefont {A.~A.}\ \bibnamefont
  {Rusu}}, \bibinfo {author} {\bibfnamefont {N.~C.}\ \bibnamefont
  {Rabinowitz}}, \bibinfo {author} {\bibfnamefont {G.}~\bibnamefont
  {Desjardins}}, \bibinfo {author} {\bibfnamefont {H.}~\bibnamefont {Soyer}},
  \bibinfo {author} {\bibfnamefont {J.}~\bibnamefont {Kirkpatrick}}, \bibinfo
  {author} {\bibfnamefont {K.}~\bibnamefont {Kavukcuoglu}}, \bibinfo {author}
  {\bibfnamefont {R.}~\bibnamefont {Pascanu}}, \ and\ \bibinfo {author}
  {\bibfnamefont {R.}~\bibnamefont {Hadsell}},\ }\href
  {http://arxiv.org/abs/1606.04671} {\enquote {\bibinfo {title} {Progressive
  {{Neural Networks}}},}\ } (\bibinfo {year} {2016}),\ \bibinfo {note}
  {preprint available on arXiv.},\ \Eprint {http://arxiv.org/abs/1606.04671}
  {1606.04671} \BibitemShut {NoStop}%
\bibitem [{\citenamefont {Yosinski}\ \emph {et~al.}(2014)\citenamefont
  {Yosinski}, \citenamefont {Clune}, \citenamefont {Bengio},\ and\
  \citenamefont {Lipson}}]{NIPS2014_532a2f85}%
  \BibitemOpen
  \bibfield  {author} {\bibinfo {author} {\bibfnamefont {J.}~\bibnamefont
  {Yosinski}}, \bibinfo {author} {\bibfnamefont {J.}~\bibnamefont {Clune}},
  \bibinfo {author} {\bibfnamefont {Y.}~\bibnamefont {Bengio}}, \ and\ \bibinfo
  {author} {\bibfnamefont {H.}~\bibnamefont {Lipson}},\ }\bibfield  {title}
  {\enquote {\bibinfo {title} {How transferable are features in deep neural
  networks?}}\ }in\ \href@noop {} {\emph {\bibinfo {booktitle} {Advances in
  Neural Information Processing Systems}}},\ Vol.~\bibinfo {volume} {27},\
  \bibinfo {editor} {edited by\ \bibinfo {editor} {\bibfnamefont
  {Z.}~\bibnamefont {Ghahramani}}, \bibinfo {editor} {\bibfnamefont
  {M.}~\bibnamefont {Welling}}, \bibinfo {editor} {\bibfnamefont
  {C.}~\bibnamefont {Cortes}}, \bibinfo {editor} {\bibfnamefont
  {N.}~\bibnamefont {Lawrence}}, \ and\ \bibinfo {editor} {\bibfnamefont
  {K.}~\bibnamefont {Weinberger}}}\ (\bibinfo  {publisher} {Curran Associates,
  Inc.},\ \bibinfo {year} {2014})\BibitemShut {NoStop}%
\bibitem [{\citenamefont {Gideon}\ \emph {et~al.}(2017)\citenamefont {Gideon},
  \citenamefont {Khorram}, \citenamefont {Aldeneh}, \citenamefont
  {Dimitriadis},\ and\ \citenamefont {Provost}}]{gideon2017PNN}%
  \BibitemOpen
  \bibfield  {author} {\bibinfo {author} {\bibfnamefont {J.}~\bibnamefont
  {Gideon}}, \bibinfo {author} {\bibfnamefont {S.}~\bibnamefont {Khorram}},
  \bibinfo {author} {\bibfnamefont {Z.}~\bibnamefont {Aldeneh}}, \bibinfo
  {author} {\bibfnamefont {D.}~\bibnamefont {Dimitriadis}}, \ and\ \bibinfo
  {author} {\bibfnamefont {E.~M.}\ \bibnamefont {Provost}},\ }\href
  {https://arxiv.org/abs/1706.03256} {\enquote {\bibinfo {title} {Progressive
  neural networks for transfer learning in emotion recognition},}\ } (\bibinfo
  {year} {2017}),\ \Eprint {http://arxiv.org/abs/1706.03256} {arXiv:1706.03256
  [cs.LG]} \BibitemShut {NoStop}%
\bibitem [{\citenamefont {Erg{\"u}n}\ and\ \citenamefont
  {T{\"o}reyin}(2021)}]{ergun2021sparse}%
  \BibitemOpen
  \bibfield  {author} {\bibinfo {author} {\bibfnamefont {E.}~\bibnamefont
  {Erg{\"u}n}}\ and\ \bibinfo {author} {\bibfnamefont {B.~U.}\ \bibnamefont
  {T{\"o}reyin}},\ }\bibfield  {title} {\enquote {\bibinfo {title} {Sparse
  progressive neural networks for continual learning},}\ }in\ \href {\doibase
  10.1007/978-3-030-88113-9_58} {\emph {\bibinfo {booktitle} {International
  Conference on Computational Collective Intelligence}}}\ (\bibinfo
  {organization} {Springer},\ \bibinfo {year} {2021})\ pp.\ \bibinfo {pages}
  {715--725}\BibitemShut {NoStop}%
\bibitem [{\citenamefont {Meng}\ \emph {et~al.}(2024)\citenamefont {Meng},
  \citenamefont {Ju}, \citenamefont {Ai}, \citenamefont {Gomez}, \citenamefont
  {Nichols},\ and\ \citenamefont {Li}}]{PNN_Robotics}%
  \BibitemOpen
  \bibfield  {author} {\bibinfo {author} {\bibfnamefont {W.}~\bibnamefont
  {Meng}}, \bibinfo {author} {\bibfnamefont {H.}~\bibnamefont {Ju}}, \bibinfo
  {author} {\bibfnamefont {T.}~\bibnamefont {Ai}}, \bibinfo {author}
  {\bibfnamefont {R.}~\bibnamefont {Gomez}}, \bibinfo {author} {\bibfnamefont
  {E.}~\bibnamefont {Nichols}}, \ and\ \bibinfo {author} {\bibfnamefont
  {G.}~\bibnamefont {Li}},\ }\bibfield  {title} {\enquote {\bibinfo {title}
  {Transferring meta-policy from simulation to reality via progressive neural
  network},}\ }\href {\doibase 10.1109/LRA.2024.3370034} {\bibfield  {journal}
  {\bibinfo  {journal} {IEEE Robotics and Automation Letters}\ }\textbf
  {\bibinfo {volume} {9}},\ \bibinfo {pages} {3696--3703} (\bibinfo {year}
  {2024})}\BibitemShut {NoStop}%
\bibitem [{\citenamefont {Moriya}\ \emph {et~al.}(2018)\citenamefont {Moriya},
  \citenamefont {Masumura}, \citenamefont {Asami}, \citenamefont {Shinohara},
  \citenamefont {Delcroix}, \citenamefont {Yamaguchi},\ and\ \citenamefont
  {Aono}}]{PNN_Acoustic}%
  \BibitemOpen
  \bibfield  {author} {\bibinfo {author} {\bibfnamefont {T.}~\bibnamefont
  {Moriya}}, \bibinfo {author} {\bibfnamefont {R.}~\bibnamefont {Masumura}},
  \bibinfo {author} {\bibfnamefont {T.}~\bibnamefont {Asami}}, \bibinfo
  {author} {\bibfnamefont {Y.}~\bibnamefont {Shinohara}}, \bibinfo {author}
  {\bibfnamefont {M.}~\bibnamefont {Delcroix}}, \bibinfo {author}
  {\bibfnamefont {Y.}~\bibnamefont {Yamaguchi}}, \ and\ \bibinfo {author}
  {\bibfnamefont {Y.}~\bibnamefont {Aono}},\ }\bibfield  {title} {\enquote
  {\bibinfo {title} {Progressive neural network-based knowledge transfer in
  acoustic models},}\ }in\ \href {\doibase 10.23919/APSIPA.2018.8659556} {\emph
  {\bibinfo {booktitle} {2018 Asia-Pacific Signal and Information Processing
  Association Annual Summit and Conference (APSIPA ASC)}}}\ (\bibinfo {year}
  {2018})\ pp.\ \bibinfo {pages} {998--1002}\BibitemShut {NoStop}%
\bibitem [{\citenamefont {Weiss}, \citenamefont {Khoshgoftaar},\ and\
  \citenamefont {Wang}(2016)}]{weiss2016survey}%
  \BibitemOpen
  \bibfield  {author} {\bibinfo {author} {\bibfnamefont {K.}~\bibnamefont
  {Weiss}}, \bibinfo {author} {\bibfnamefont {T.~M.}\ \bibnamefont
  {Khoshgoftaar}}, \ and\ \bibinfo {author} {\bibfnamefont {D.}~\bibnamefont
  {Wang}},\ }\bibfield  {title} {\enquote {\bibinfo {title} {A survey of
  transfer learning},}\ }\href {\doibase 10.1186/s40537-016-0043-6} {\bibfield
  {journal} {\bibinfo  {journal} {Journal of Big data}\ }\textbf {\bibinfo
  {volume} {3}},\ \bibinfo {pages} {9} (\bibinfo {year} {2016})}\BibitemShut
  {NoStop}%
\bibitem [{\citenamefont {Sutton}\ and\ \citenamefont
  {Barto}(2018)}]{sutton2018reinforcement}%
  \BibitemOpen
  \bibfield  {author} {\bibinfo {author} {\bibfnamefont {R.~S.}\ \bibnamefont
  {Sutton}}\ and\ \bibinfo {author} {\bibfnamefont {A.~G.}\ \bibnamefont
  {Barto}},\ }\href@noop {} {\emph {\bibinfo {title} {Reinforcement learning:
  An introduction}}}\ (\bibinfo  {publisher} {MIT press},\ \bibinfo {year}
  {2018})\BibitemShut {NoStop}%
\bibitem [{\citenamefont {Haarnoja}\ \emph {et~al.}(2018)\citenamefont
  {Haarnoja}, \citenamefont {Zhou}, \citenamefont {Abbeel},\ and\ \citenamefont
  {Levine}}]{SAC}%
  \BibitemOpen
  \bibfield  {author} {\bibinfo {author} {\bibfnamefont {T.}~\bibnamefont
  {Haarnoja}}, \bibinfo {author} {\bibfnamefont {A.}~\bibnamefont {Zhou}},
  \bibinfo {author} {\bibfnamefont {P.}~\bibnamefont {Abbeel}}, \ and\ \bibinfo
  {author} {\bibfnamefont {S.}~\bibnamefont {Levine}},\ }\href
  {https://arxiv.org/abs/1801.01290} {\enquote {\bibinfo {title} {Soft
  actor-critic: Off-policy maximum entropy deep reinforcement learning with a
  stochastic actor},}\ } (\bibinfo {year} {2018}),\ \Eprint
  {http://arxiv.org/abs/1801.01290} {arXiv:1801.01290 [cs.LG]} \BibitemShut
  {NoStop}%
\bibitem [{\citenamefont
  {Sivashinsky}(1977)}]{sivashinskyNonlinearAnalysisHydrodynamic1977}%
  \BibitemOpen
  \bibfield  {author} {\bibinfo {author} {\bibfnamefont {G.~I.}\ \bibnamefont
  {Sivashinsky}},\ }\bibfield  {title} {\enquote {\bibinfo {title} {Nonlinear
  analysis of hydrodynamic instability in laminar flames---{{I}}.
  {{Derivation}} of basic equations},}\ }\href {\doibase
  10.1016/0094-5765(77)90096-0} {\bibfield  {journal} {\bibinfo  {journal}
  {Acta Astronautica}\ }\textbf {\bibinfo {volume} {4}},\ \bibinfo {pages}
  {1177--1206} (\bibinfo {year} {1977})}\BibitemShut {NoStop}%
\bibitem [{\citenamefont {Kuramoto}\ and\ \citenamefont
  {Tsuzuki}(1976)}]{kuramotoPersistentPropagationConcentration1976}%
  \BibitemOpen
  \bibfield  {author} {\bibinfo {author} {\bibfnamefont {Y.}~\bibnamefont
  {Kuramoto}}\ and\ \bibinfo {author} {\bibfnamefont {T.}~\bibnamefont
  {Tsuzuki}},\ }\bibfield  {title} {\enquote {\bibinfo {title} {Persistent
  {{Propagation}} of {{Concentration Waves}} in {{Dissipative Media Far}} from
  {{Thermal Equilibrium}}},}\ }\href {\doibase 10.1143/PTP.55.356} {\bibfield
  {journal} {\bibinfo  {journal} {Progress of Theoretical Physics}\ }\textbf
  {\bibinfo {volume} {55}},\ \bibinfo {pages} {356--369} (\bibinfo {year}
  {1976})}\BibitemShut {NoStop}%
\bibitem [{\citenamefont {Cvitanovi{\'c}}, \citenamefont {Davidchack},\ and\
  \citenamefont {Siminos}(2010)}]{cvitanovicStateSpaceGeometry2010a}%
  \BibitemOpen
  \bibfield  {author} {\bibinfo {author} {\bibfnamefont {P.}~\bibnamefont
  {Cvitanovi{\'c}}}, \bibinfo {author} {\bibfnamefont {R.~L.}\ \bibnamefont
  {Davidchack}}, \ and\ \bibinfo {author} {\bibfnamefont {E.}~\bibnamefont
  {Siminos}},\ }\bibfield  {title} {\enquote {\bibinfo {title} {On the {{State
  Space Geometry}} of the {{Kuramoto}}--{{Sivashinsky Flow}} in a {{Periodic
  Domain}}},}\ }\href {\doibase 10.1137/070705623} {\bibfield  {journal}
  {\bibinfo  {journal} {SIAM Journal on Applied Dynamical Systems}\ }\textbf
  {\bibinfo {volume} {9}},\ \bibinfo {pages} {1--33} (\bibinfo {year}
  {2010})}\BibitemShut {NoStop}%
\bibitem [{\citenamefont {Zeng}\ and\ \citenamefont {Graham}(2021)}]{Zeng2021}%
  \BibitemOpen
  \bibfield  {author} {\bibinfo {author} {\bibfnamefont {K.}~\bibnamefont
  {Zeng}}\ and\ \bibinfo {author} {\bibfnamefont {M.~D.}\ \bibnamefont
  {Graham}},\ }\bibfield  {title} {\enquote {\bibinfo {title} {Symmetry
  reduction for deep reinforcement learning active control of chaotic
  spatiotemporal dynamics},}\ }\href {\doibase 10.1103/PhysRevE.104.014210}
  {\bibfield  {journal} {\bibinfo  {journal} {Physical Review E}\ }\textbf
  {\bibinfo {volume} {104}} (\bibinfo {year} {2021}),\
  10.1103/PhysRevE.104.014210},\ \Eprint {http://arxiv.org/abs/2104.05437}
  {arXiv:2104.05437} \BibitemShut {NoStop}%
\bibitem [{\citenamefont {Zeng}, \citenamefont {Linot},\ and\ \citenamefont
  {Graham}(2022)}]{zengDatadrivenControlSpatiotemporal2022}%
  \BibitemOpen
  \bibfield  {author} {\bibinfo {author} {\bibfnamefont {K.}~\bibnamefont
  {Zeng}}, \bibinfo {author} {\bibfnamefont {A.~J.}\ \bibnamefont {Linot}}, \
  and\ \bibinfo {author} {\bibfnamefont {M.~D.}\ \bibnamefont {Graham}},\
  }\bibfield  {title} {\enquote {\bibinfo {title} {Data-driven control of
  spatiotemporal chaos with reduced-order neural {{ODE-based}} models and
  reinforcement learning},}\ }\href {\doibase 10.1098/rspa.2022.0297}
  {\bibfield  {journal} {\bibinfo  {journal} {Proceedings of the Royal Society
  A: Mathematical, Physical and Engineering Sciences}\ }\textbf {\bibinfo
  {volume} {478}},\ \bibinfo {pages} {20220297} (\bibinfo {year}
  {2022})}\BibitemShut {NoStop}%
\bibitem [{\citenamefont {Paris}, \citenamefont {Beneddine},\ and\
  \citenamefont {Dandois}(2023)}]{Paris2023}%
  \BibitemOpen
  \bibfield  {author} {\bibinfo {author} {\bibfnamefont {R.}~\bibnamefont
  {Paris}}, \bibinfo {author} {\bibfnamefont {S.}~\bibnamefont {Beneddine}}, \
  and\ \bibinfo {author} {\bibfnamefont {J.}~\bibnamefont {Dandois}},\
  }\bibfield  {title} {\enquote {\bibinfo {title}
  {{Reinforcement-learning-based actuator selection method for active flow
  control}},}\ }\href {\doibase 10.1017/jfm.2022.1043} {\bibfield  {journal}
  {\bibinfo  {journal} {Journal of Fluid Mechanics}\ }\textbf {\bibinfo
  {volume} {955}},\ \bibinfo {pages} {A8} (\bibinfo {year} {2023})}\BibitemShut
  {NoStop}%
\bibitem [{\citenamefont {Werner}\ and\ \citenamefont
  {Peitz}(2024)}]{wernerNumericalEvidenceSample2024}%
  \BibitemOpen
  \bibfield  {author} {\bibinfo {author} {\bibfnamefont {S.}~\bibnamefont
  {Werner}}\ and\ \bibinfo {author} {\bibfnamefont {S.}~\bibnamefont {Peitz}},\
  }\bibfield  {title} {\enquote {\bibinfo {title} {Numerical {{Evidence}} for
  {{Sample Efficiency}} of {{Model-Based Over Model-Free Reinforcement Learning
  Control}} of {{Partial Differential Equations}}},}\ }in\ \href {\doibase
  10.23919/ECC64448.2024.10590945} {\emph {\bibinfo {booktitle} {2024
  {{European Control Conference}} ({{ECC}})}}}\ (\bibinfo {year} {2024})\ pp.\
  \bibinfo {pages} {2965--2971}\BibitemShut {NoStop}%
\bibitem [{\citenamefont {Peitz}\ \emph {et~al.}(2024)\citenamefont {Peitz},
  \citenamefont {Stenner}, \citenamefont {Chidananda}, \citenamefont
  {Wallscheid}, \citenamefont {Brunton},\ and\ \citenamefont
  {Taira}}]{peitzDistributedControlPartial2024}%
  \BibitemOpen
  \bibfield  {author} {\bibinfo {author} {\bibfnamefont {S.}~\bibnamefont
  {Peitz}}, \bibinfo {author} {\bibfnamefont {J.}~\bibnamefont {Stenner}},
  \bibinfo {author} {\bibfnamefont {V.}~\bibnamefont {Chidananda}}, \bibinfo
  {author} {\bibfnamefont {O.}~\bibnamefont {Wallscheid}}, \bibinfo {author}
  {\bibfnamefont {S.~L.}\ \bibnamefont {Brunton}}, \ and\ \bibinfo {author}
  {\bibfnamefont {K.}~\bibnamefont {Taira}},\ }\bibfield  {title} {\enquote
  {\bibinfo {title} {Distributed control of partial differential equations
  using convolutional reinforcement learning},}\ }\href {\doibase
  10.1016/j.physd.2024.134096} {\bibfield  {journal} {\bibinfo  {journal}
  {Physica D: Nonlinear Phenomena}\ }\textbf {\bibinfo {volume} {461}},\
  \bibinfo {pages} {134096} (\bibinfo {year} {2024})},\ \Eprint
  {http://arxiv.org/abs/2301.10737} {arXiv:2301.10737} \BibitemShut {NoStop}%
\bibitem [{\citenamefont {Whitaker}(2015)}]{JswhitPyksData}%
  \BibitemOpen
  \bibfield  {author} {\bibinfo {author} {\bibfnamefont {J.}~\bibnamefont
  {Whitaker}},\ }\href@noop {} {\enquote {\bibinfo {title} {Jswhit/pyks: Data
  assimilation for the {1-D {Kuramoto-Sivashinsky}} equation},}\ }\bibinfo
  {howpublished} {https://github.com/jswhit/pyks/} (\bibinfo {year}
  {2015})\BibitemShut {NoStop}%
\bibitem [{\citenamefont {Raffin}\ \emph {et~al.}(2021)\citenamefont {Raffin},
  \citenamefont {Hill}, \citenamefont {Gleave}, \citenamefont {Kanervisto},
  \citenamefont {Ernestus},\ and\ \citenamefont {Dormann}}]{stable-baselines3}%
  \BibitemOpen
  \bibfield  {author} {\bibinfo {author} {\bibfnamefont {A.}~\bibnamefont
  {Raffin}}, \bibinfo {author} {\bibfnamefont {A.}~\bibnamefont {Hill}},
  \bibinfo {author} {\bibfnamefont {A.}~\bibnamefont {Gleave}}, \bibinfo
  {author} {\bibfnamefont {A.}~\bibnamefont {Kanervisto}}, \bibinfo {author}
  {\bibfnamefont {M.}~\bibnamefont {Ernestus}}, \ and\ \bibinfo {author}
  {\bibfnamefont {N.}~\bibnamefont {Dormann}},\ }\bibfield  {title} {\enquote
  {\bibinfo {title} {Stable-baselines3: Reliable reinforcement learning
  implementations},}\ }\href {http://jmlr.org/papers/v22/20-1364.html}
  {\bibfield  {journal} {\bibinfo  {journal} {Journal of Machine Learning
  Research}\ }\textbf {\bibinfo {volume} {22}},\ \bibinfo {pages} {1--8}
  (\bibinfo {year} {2021})}\BibitemShut {NoStop}%
\end{thebibliography}
\end{document}